%% file: main.tex
\documentclass[twoside]{article}

\usepackage[arxiv]{aistats2023}

\usepackage[round]{natbib}
\renewcommand{\bibname}{References}

\usepackage[utf8]{inputenc}

\usepackage{xcolor}
\usepackage{algorithm}
\usepackage{algorithmic}
\usepackage{booktabs}
\usepackage{hyperref}
\usepackage[capitalise]{cleveref}
\hypersetup{
    colorlinks,
    linkcolor={blue!50!black},
    citecolor={blue!50!black},
    urlcolor={blue!80!black}
}
\usepackage{amssymb,amsfonts,amsthm,amsmath,soul,cancel,enumitem}
\usepackage{nicefrac}
\usepackage{subfigure}
\usepackage{minitoc}
\usepackage{caption}
\usepackage{tikz} %
\usetikzlibrary{calc, fadings, decorations, shapes, positioning, arrows}

\mtcsettitle{minitoc}{}
\mtcsetrules{*}{off}

\input{def.set}
\newcommand{\gtsv}[1]{\bar\psi_{#1}}

\newcommand{\divLB}[1][h]{\cD_{\chi^2,\mrm{lb}}({#1})}
\newcommand{\popuLoss}[1][h]{\cD({#1})}
\newcommand{\gtRKHS}[1][\beta]{\bar\cH_{{#1}}}
\newcommand{\Dh}{\cD^{(n)}_1}
\newcommand{\Dcov}{\cD^{(n)}_2}
\newcommand{\Dkiv}{\cD^{(n)}_3}
\DeclareMathOperator{\Ran}{Ran}

\newcommand{\eg}{\emph{e.g.}}

\newcommand{\zwLQ}[1]{}

\def\conf{0}
\if\conf1
\newcommand{\vshrink}[1]{\vspace{#1}}
\else
\newcommand{\vshrink}[1]{}
\fi
\def\confAppendixHack{0}

\def\draft{0}
\if\draft1
\newcommand{\SkipNOTE}[1]{{\color{gray!80} [#1]}}
\usepackage[color=lightgray,textsize=tiny]{todonotes}
\else 
\newcommand{\SkipNOTE}[1]{}
\usepackage[disable]{todonotes}
\fi

\if\confAppendixHack1
\usepackage{bibunits}
\fi

\begin{document}

\def\parshrink{0.5em}

\runningauthor{Ziyu Wang, Yucen Luo, Yueru Li, Jun Zhu, Bernhard Sch\"olkopf}

\twocolumn[

\aistatstitle{%
Spectral Representation Learning for Conditional Moment Models}

\aistatsauthor{ Ziyu Wang \And Yucen Luo \And Yueru Li} 

\aistatsaddress{ Tsinghua University \And  Max Planck Institute for Intelligent Systems \And Tsinghua University } 

\aistatsauthor{Jun Zhu \And Bernhard Sch\"olkopf }

\aistatsaddress{ Tsinghua University \And  Max Planck Institute for Intelligent Systems} 
]

\begin{abstract}
Many problems in causal inference and economics can be formulated in the framework of conditional moment models, which characterize the target function through a collection of conditional moment restrictions. 
For nonparametric conditional moment models, efficient estimation often relies on preimposed conditions on various measures of ill-posedness of the hypothesis space, which 
are hard to validate when flexible models are used. %
In this work, we %
address this issue by proposing a procedure that automatically learns representations with controlled measures of ill-posedness. 
Our method approximates a linear representation defined by the spectral decomposition of a conditional expectation operator, which can be used for kernelized estimators and is known to facilitate minimax optimal estimation in certain settings. %
We show this representation can be efficiently estimated from data, and establish $L_2$ consistency for the resulting estimator. %
We evaluate the proposed method on proximal causal inference tasks, exhibiting promising performance on high-dimensional, semi-synthetic data.
\end{abstract}

\doparttoc %
\faketableofcontents %

\if\confAppendixHack1
\begin{bibunit}[apalike]
\fi

\input{fig-cg.tex}

\input{sec-intro.tex}

\input{sec-setup.tex}
\input{sec-estimability.tex} %
\input{sec-svd.tex}

\input{sec-related-work.tex}
\input{sec-exp.tex}

\input{sec-conclusion.tex}

\if\confAppendixHack1
\putbib[main]
\end{bibunit}
\newcommand{\prodHypoSpace}{1}
\else
\bibliographystyle{apalike}
\bibliography{main}
\fi

\appendix

\if\confAppendixHack1
\begin{bibunit}[apalike]
\fi

\onecolumn
\aistatstitle{
Spectral Representation Learning for Conditional Moment Models:\\
Supplementary Materials
}

\part{} %
\parttoc %
\clearpage
\input{app-additional-discussion.tex}
\input{app-proof-sec3.tex}
\input{app-proof.tex}

\input{app-proof-extra-repr.tex}

\input{app-cmm-results.tex}

\input{app-exp.tex}

\if\confAppendixHack1
\renewcommand\bibname{}
\section{Supplementary References}
\putbib[main]
\end{bibunit}
\fi

\end{document}

%% file: fig-cg.tex
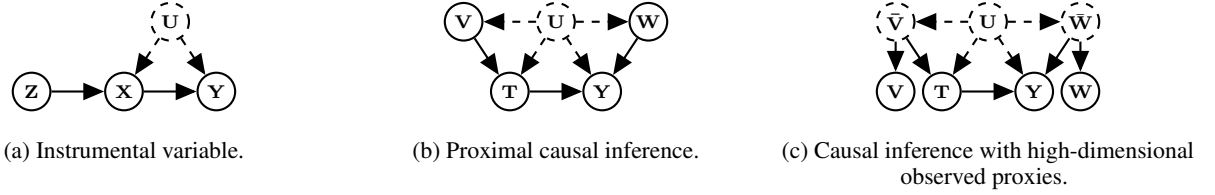
\begin{figure*}
    \begin{minipage}[t]{.33\textwidth}
    \centering
  \begin{tikzpicture}[node distance=4mm and 7mm, thick, main/.style = {draw, circle, minimum size=0.5cm, inner sep=1pt}, >={triangle 45}]  
    \node[main] (z) {$\scriptstyle \bZ$}; 
    \node[main] (x) [right =of z] {$\scriptstyle \bX$}; 
    \node[main] (y) [right =of x] {$\scriptstyle \bY$}; 
    \node[main, dashed] (u) [above =of  $(x.north)!0.5!(y.north)$] {$\scriptstyle \bU$}; 
    \draw[->] (z) -- (x); 
    \draw[->, dashed] (u) -- (x); 
    \draw[->, dashed] (u) -- (y); 
    \draw[->] (x) -- (y); 
  \end{tikzpicture}
\\\vspace{0.75em}
  \footnotesize{(a) Instrumental variable.}
  \end{minipage}
\hfill
\begin{minipage}[t]{.33\textwidth}
\centering
    \begin{tikzpicture}[node distance=4mm and 7mm, thick, main/.style = {draw, circle, minimum size=0.5cm, inner sep=1pt}, >={triangle 45}]  
    \node[main] (v) {$\scriptstyle \bV$}; 
    \node[main, dashed] (u) [right =of v] {$\scriptstyle \bU$}; 
    \node[main] (w) [right =of u] {$\scriptstyle \bW$}; 
    \node[main] (t) [below =of $(v.south)!0.5!(u.south)$] {$\scriptstyle \bT$}; 
    \node[main] (y) [right =of t] {$\scriptstyle \bY$}; 
    \draw[->] (v) -- (t);
    \draw[->, dashed] (u) -- (t); 
    \draw[->, dashed] (u) -- (y); 
    \draw[->] (t) -- (y); 
    \draw[->, dashed] (u) -- (w);
    \draw[->, dashed] (u) -- (v);
    \draw[->] (w) -- (y);
  \end{tikzpicture}%
\\\vspace{0.75em}
  \footnotesize{(b) Proximal causal inference.}
  \end{minipage}
\hfill
\begin{minipage}[t]{.33\textwidth}
\centering
    \begin{tikzpicture}[node distance=4mm and 7mm, thick, main/.style = {draw, circle, minimum size=0.5cm, inner sep=1pt}, >={triangle 45}]  
    \node[main, dashed] (v) {$\scriptstyle \mathbf{\bar \bV}$}; 
    \node[main, dashed] (u) [right =of v] {$\scriptstyle \bU$}; 
    \node[main, dashed] (w) [right =of u] {$\scriptstyle \bar\bW$}; 
    \node[main] (t) [below =of $(v.south)!0.5!(u.south)$] {$\scriptstyle \bT$}; 
    \node[main] (y) [right =of t] {$\scriptstyle \bY$}; 
    \node[main] (vv) [below =of v] {$\scriptstyle \bV$}; 
    \node[main] (ww) [below =of w] {$\scriptstyle \bW$}; 
    \draw[->] (v) -- (t);
    \draw[->, dashed] (u) -- (t); 
    \draw[->, dashed] (u) -- (y); 
    \draw[->] (t) -- (y); 
    \draw[->, dashed] (u) -- (w);
    \draw[->, dashed] (u) -- (v);
    \draw[->] (w) -- (y);
    \draw[->] (v) -- (vv);
    \draw[->] (w) -- (ww);
  \end{tikzpicture}
\\\vspace{0.75em}
  \footnotesize{
    (c) Causal inference with high-dimensional observed proxies.
  }
  \end{minipage}
\caption{Sample causal graphs for various conditional moment models.}\label{fig:cg}
\end{figure*}

%% file: sec-intro.tex
\vshrink{-0.5em}
\section{INTRODUCTION}
\vshrink{-0.5em}

\todo{FIXME: compiling in draft mode}
Many problems in causal inference and statistics can be formulated in the framework of \emph{conditional moment models} (CMMs), where the goal is to
\begin{equation}\label{eq:cmm-defn}
\text{find $f_0$ s.t.}~~
\EE(f_0(\bx)-\by\mid\bz)=0~\text{a.s.}~[P(dz)].
\end{equation}
Examples include instrumental variable (IV) regression \citep{wright1928tariff}, where $\bx,\by,\bz$ denote the treatment, outcome and {\em instruments} (\cref{fig:cg} a); proximal causal inference \citep{miao_identifying_2018,tchetgen_introduction_2020}), where $\bx,\bz$ are obtained by concatenating the treatment variable and two {\em proxy variables} of the unobserved confounder (\cref{fig:cg} b); and various 
problems in reinforcement learning where similar instruments or proxies can be found \citep[e.g.,][]{chen2021instrumental,li2021causal,shi_minimax_2022}. 

For nonparametric models, 
estimation of \eqref{eq:cmm-defn} usually relies on minimizing estimable surrogates for an averaged violation of the conditional moment restrictions, 
\begin{equation}\label{eq:popu-obj}
f\mapsto \EE(\EE(f(\bx)-\by\mid\bz)^2).
\end{equation}
It is relatively easy to establish finite-sample guarantees for \eqref{eq:popu-obj}, %
using statistical learning theory techniques. However, nonparametric CMMs are 
typically {\em ill-posed} \citep{kress1989linear}, meaning that approximate minimization of \eqref{eq:popu-obj}
cannot guarantee approximate recovery of $f_0$ in more intuitive %
criteria, %
such as the $L_2$ norm, i.e., the root mean squared error \emph{over $\bx$}. Such distinctions between identifiability and estimability are %
ubiquitous in nonlinear causal problems \citep{horowitz_ill-posed_2014,maclaren_what_2020}. 

A standard approach for establishing uniform $L_2$ convergence is to impose assumptions on 
certain {\em moduli of continuity} \citep{chen_estimation_2012,knapik_general_2017}, which directly quantifies the maximum possible $L_2$ error for functions that approximately optimize \eqref{eq:popu-obj} or its surrogate and live in the {\em given hypothesis space}. 
Unfortunately, such assumptions are rather opaque: 
as we explain in \cref{sec:setup}, they require the
complexity of the hypothesis space to be related to that 
of the {\em conditional expectation operator} which defines \eqref{eq:cmm-defn}, which is usually unknown {\em a priori}. These assumptions are thus difficult to justify, especially with flexible machine learning models using neural networks, and %
hard, if not impossible, to verify empirically.
Therefore, it is highly desirable to develop methods that automatically learn hypothesis spaces with well-controlled moduli of continuity.

This work is motivated to fill this crucial gap %
by learning linear representations -- {\em kernels} that are based on the spectral decomposition of the conditional expectation operator. %
It is long known that estimation based on the exact decomposition would be minimax optimal in a certain sense, as we discuss in \cref{sec:setup}; 
but no guarantee has been established for the use of approximate decompositions that {\em can %
be learned from finite observations}. 
We prove that 
such decompositions can still define hypothesis spaces with well-controlled moduli of continuity, 
and present a kernel learning algorithm for an estimator of \eqref{eq:cmm-defn} that enables $L_2$-consistent estimation,  
with rates adaptive to certain latent variables of the problem. 
When instantiated on the problem of proximal causal inference, our method demonstrates competitive performance on high-dimensional semi-synthetic datasets. %

Our analysis reveals a quantitative benefit of the kernelized formulation of IV/CMM estimation: as shown in \cref{sec:est-klearn}, by learning the inner product structure in addition to the basis of the linear representation space, we can avoid the catastrophic sharp transition behavior, in which the modulus of continuity, and thus the $L_2$ error, 
become unbounded once the spectral decomposition error increases above an {\em unobservable} threshold. %
When working with learned kernels, the measure degrades gracefully.
This is interesting because although kernelized estimators generalize the classical sieve methods, %
in past works on estimation with fixed-form representations, the theoretical benefits of kernelized methods %
have been less transparent: %
for instance, both approaches attain the same minimax rate \citep{chen_rate_2007,wang2021quasibayesian}. 

Spectral decomposition is an important problem on its own, for which various procedures have been proposed~\citep{pfau_spectral_2020,klus_eigendecompositions_2020,gemp2020eigengame,Neuralef}. 
We improve the understanding of this problem by showing that {\em faster rates of convegence} can be achieved through a contrastive learning procedure, which maximizes a variational formulation of an $f$-divergence. %
This is different from existing methods based on NN models, which not only lack the faster rate guarantees but also require more computationally expensive bi-level optimization. 

The rest of this paper is organized as follows: \cref{sec:setup} reviews background knowledge and sets up a running example. In \cref{sec:estimability} we 
establish abstract guarantees for kernels %
derived from approximate spectral decompositions. %
In \cref{sec:estimation} we construct an algorithm that fulfills the abstract criterion, and establish $L_2$ consistency for the resulted estimator of \eqref{eq:cmm-defn}. We review related work in \cref{sec:related-work} and evaluate our method experimentally in \cref{sec:exp}. Finally, we discuss the conclusion and future work in \cref{sec:conclusion}.

%% file: sec-setup.tex
\section{NOTATION AND SETUP}\label{sec:setup}

\vshrink{-\parshrink}
\paragraph{Notations} We use $P_z,P_x,P_{zx}$ to denote marginal data distributions of the corresponding covariate(s), and $P_z\otimes P_x$ denotes the product measure.  
For functions $f,g$ of any covariate(s), $\<f,g\>_2$ denotes the $L_2$ inner product w.r.t.~the respective data distribution. For finite-dimensional vectors, $\<\cdot,\cdot\>_2$ denotes the Euclidean inner product. $\|\cdot\|_2$ denotes the induced norm. %
$\Ran(\cdot)$ denotes the range of an operator. $\asymp$ denotes equivalence up to multiplicative constants.

\vshrink{-\parshrink}
\paragraph{Conditional moment models} We first consider estimation problems in \eqref{eq:cmm-defn}, which defines a continuum of moment restrictions indexed by $z$. To 
derive an estimation objective, we observe 
that an averaged violation of the moment restrictions, Eq.~\eqref{eq:popu-obj}, can be estimated from data:
\begin{align*}
&\phantom{=} \EE(\EE(f(\bx)-\by\mid\bz)^2) 
\\ &= \sup_{g\in L_2(P_z)} \EE(2\EE(f(\bx)-\by\cancel{\mid\bz}) g(\bz)) -\EE(g^2(\bz))
\\ &\approx \sup_{g\in \cI} \frac{1}{n} \sum_{i=1}^n 2\left(f\left(x_i\right)-y_i\right)g(z_i) - g^2(z_i).
\end{align*}
In the above, 
$(x_i,y_i,z_i)$ are i.i.d.~data samples, and 
$\cI\subset L_2(P_z)$ denotes a smaller space in which the optimization can be implemented; 
the equality follows by a basic property of the inner product space $L_2(P_z)$, and the tower property. %
As the left-hand side equals zero if and only if $f$ satisfies \eqref{eq:cmm-defn}, an estimator $\hat f_n$
can be derived:
\begin{align*}
\hat f_n := \arg\min_{f\in\cH} \sup_{g\in\cI} &\:\frac{1}{n}\sum_{i=1}^n\big(2(f(x_i)-y_i)g(z_i) - g^2(z_i)\big) \\ & - \nu_n \|g\|_\cI^2 + \lambda_n \|f\|_\cH^2,\numberthis\label{eq:minimax-estimator}
\end{align*}
where $\cH$ is the hypothesis space. This type of estimator has been proposed in \citet{lewis2018adversarial,bennett2019deep,liao_provably_2020,muandet_dual_2020}, and is also related to the formulation of \citet{muandet2020kernel,zhang_maximum_2020}; see \cref{app:disc-cmm-estimators} for details. 
\citet{dikkala_minimax_2020} established fast-rate convergence 
for the unregularized population objective \eqref{eq:popu-obj};  %
their results require $\cH,\cI$ to have small (local Rademacher) complexity and approximation overhead. 
To match the classical minimax rates \citep{chen_rate_2007}, however, additional assumption on $\cH$ must be imposed \citep{%
wang2021quasibayesian}. 

While \eqref{eq:minimax-estimator} requires two specific choices, 
the choice of $\cI$ is easier because the requirements imposed by \citet{dikkala_minimax_2020} 
are the same as those in standard learning theory. %
Moreover, when $\cH$ is %
a reproducing kernel Hilbert space (RKHS), there exists an ``optimal RKHS'' $\cI$ that generalizes the notion of optimal instrument \citep{chamberlain_asymptotic_1987}, and can be efficiently learned from data \citep{wang2022fast}. %

In contrast, the choice of $\cH$ is more difficult, especially\footnote{
A controlled modulus of continuity can also lead to improved bounds for the population loss \citep{wang2021quasibayesian}.
} if one wishes to establish $L_2$ convergence, 
because it is necessary to control quantities of the form 
\begin{align*}
\omega_n^2 &:= \sup\{\|f-f_0\|_2^2: f\in\cH, \\ 
&\hspace{6em} \EE(\EE(f(\bx)-\by\mid\bz)^2) \le \delta_n^2\}, \numberthis\label{eq:moc-orig}
\end{align*}
which translates the learning theoretic guarantees for the population loss to the $L_2$ error of interest. %
Such {\bf moduli of continuity} are properties of $\cH$, and rely on the interaction between the complexity of $\cH$ and that of the conditional expectation operator 
$$
E: L_2(P_x) \to L_2(P_z), ~f\mapsto \EE(f(\bx)\mid\bz=\cdot),
$$
in a certain sense. In order to illustrate the interaction, suppose $E$ has 
spectral decomposition 
$ %
E = \sum_{i=1}^\infty \bar s_i \bar\phi_i\otimes \bar \psi_i,
$
where $\{\bar\phi_i\}\subset L_2(P_z),\{\gtsv{i}\}\subset L_2(P_x)$ are the singular vectors. Then $\cH = \mrm{span}\{\gtsv{i_1},\ldots,\gtsv{i_J}\}$ has $
\omega_n^2/\delta_n^2 = \max\{\bar s_{i_1},\ldots,\bar s_{i_J}\}$.
More generally, controlling the modulus of continuity requires $\cH$ to have a stronger regularization effect in the subspace spanned by the trailing singular vectors of $E$ while less information is known about $f_0$ in those directions. 
The requirement is nontrivial for models that are not defined in accordance to $E$, especially for flexible models such as deep neural networks. %

\vshrink{-\parshrink}
\paragraph{CMM as an inverse problem} \eqref{eq:cmm-defn} is related to the following linear inverse problem \citep{hall_nonparametric_2005}: 
\begin{equation}\label{eq:cmm-ip}\tag{\ref{eq:cmm-defn}'}
\text{find $f_0$ s.t.}~~
E f_0 = \hat g_n, 
\end{equation}
where $\hat g_n$ is a ``noisy observation'' for $g_0 := \EE(\by\mid\bz=\cdot)$. %
We can assume access to such $\hat g_n$ because 
relevant information about $g_0$ can be estimated from data. 
If the operator $E$ were known, \eqref{eq:cmm-ip} reduces to a standard linear inverse problem, for which a common approach 
is to find an inner product space $\cH$ that well approximates $f_0$, and to solve the regularized problem: %
\begin{align}
\hat f_n &:= \arg\min_{f\in\cH} \|E f - \hat g_n\|^2_2 + \lambda_n \|f\|_\cH^2 \label{eq:regularized-population-obj}\tag{\ref{eq:minimax-estimator}'} \\ 
&\phantom{:}=
\arg\min_{f\in\cH} \|E f - \hat g_n\|^2_2 + \lambda_n \|T^{-1/2} f\|_2^2,\nonumber
\end{align}
where $T: L_2(P(dx))\to L_2(P(dx))$ is a self-adjoint %
operator that characterizes the inner product $\<\cdot,\cdot\>_\cH$. \eqref{eq:regularized-population-obj} is closely related to \eqref{eq:minimax-estimator}, which also has its first term estimating $\EE(\EE(f(\bx)-\by\mid\bz)^2)=\|Ef-g_0\|_2^2$. 

The operator $T$ fully characterizes our prior knowledge about $f_0$. In machine learning, such knowledge is often characterized by a {\bf kernel} $k_x$, which defines an RKHS \citep{SchSmo02}. %
The two approaches are often equivalent: for any $k_x$, setting $T$ to its $P_x$-integral operator recovers the inner product structure of the corresponding RKHS; conversely, for any self-adjoint, trace-class operator $T$, there exists an RKHS that defines the same inner product as $\cH$ \citep{steinwart_mercers_2012}. %
Thus, the specification of a data-dependent $T$ can be equivalently viewed as a kernel learning process. %

As shown in our discussion about \eqref{eq:cmm-defn} earlier, 
the problem \eqref{eq:cmm-ip} is also typically ill-posed. %
Following the convention in the nonparametric literature \citep[e.g.,][]{darolles2011nonparametric,chen_estimation_2012}, 
we first assume the variables {\em $\bz$ and $\bx$ have no overlap}, deferring the general case to \cref{sec:algo-impl}. Then, 
under mild regularity conditions, the operator $E$ is compact, and 
it is recognized that efficient estimation
can be achieved if $\cH$ is defined in accordance to its SVD \citep{carrasco2007linear}: 
let $\{\gtsv{j}\}$ denote the sorted right singular vectors of $E$, and 
$\beta> 0$ be the largest number s.t.~the {\em source condition} holds with exponent $\beta$:
\begin{equation}\label{eq:general-source-cond}
f_0 \in \Ran\,(E^\top E)^{\beta/2}. 
\end{equation}
Then, rate-optimal estimation %
can be achieved by defining $\cH$ with 
$
\bar T_\alpha = (E^\top E)^\alpha = \sum_{j=1}^\infty \bar s_j^{2\alpha}\gtsv{j}\gtsv{j}^\top, 
$
for any $\alpha>0$, or by defining $\cH := \mrm{span}(\{\gtsv{j}\}_{j=1}^{J_n})$, for some slowly growing $J_n$.  
The requirement of using the exact $\{\gtsv{j}\}$ can be relaxed into a {\em link condition} \citep{chen_rate_2007}. However, as will be explained in App.~\ref{app:disc-cmm-estimators} in more detail, 
all existing approaches require {\em a priori} knowledge of %
inner product spaces of the form %
\begin{equation}\label{eq:ideal-rkhs}
\gtRKHS[\alpha] := \Ran\,(E^\top E)^{\alpha/2}, ~
\|f\|_{\gtRKHS[\alpha]} := \|(E^\top E)^{-\frac{\alpha}{2}} f\|_2.
\end{equation}
This is a strong assumption, especially in high dimensions, because 
as we demonstrate below, it requires knowledge of certain lower-dimensional latent variables that explains all correlation structures between $\bx$ and $\bz$.

We now set up the running example in this work, %
which also provides intuition on the inductive bias of $\gtRKHS[\alpha]$:
\begin{example}[compositional model]\label{ex:main}
Let $\bar\bx,\bar\bz$ be correlated random variables, and $\bx_{\perp},\bz_{\perp}$ be s.t.~$\bx_{\perp}\indep(\bz_{\perp},\bar\bx,\bar\bz), \bz_{\perp}\indep(\bx_{\perp},\bar\bx,\bar\bz)$. Suppose there exist functions s.t.~$\bx = f_{dec,x}(\bar\bx, \bx_{\perp}), \bz = f_{dec,z}(\bar\bz, \bz_{\perp}), \bar\bx = f_{fea,x}(\bx), \bar\bz = f_{fea,z}(\bz)$. 

We can show that when $\alpha>0$, any 
$
f\in \gtRKHS[\alpha] 
$ must have the form of $f = \bar f\circ f_{enc,x}$, meaning that $f$ only depends on $\bx$ through $\bar\bx$; see App.~\ref{app:disc-gtRKHS}. %
Intuitively, $(\bar\bx,\bar\bz)$ are {\em informative latent variables} that explain all (nonlinear) correlations between $(\bx,\bz)$.
For all $f,f'$ on suitable domains, we have\vshrink{-0.5em}
\begin{enumerate}[leftmargin=*,label=(\roman*)]
    \item \cref{eq:cmm-defn} is equivalent to $\EE(f(\bx)-\by\mid \bar\bz)=0$. 
    \item\label{it:pc} In proxy control problems, $\EE(f(\bu)\mid \bx) = \EE(f(\bu)\mid\bar\bx)$ and 
    $\EE(f(\bu)\mid \bz) = \EE(f(\bu)\mid\bar\bz)$. 
    \item\label{it:falsifiable} $\EE(f(\bar\bx)f'(\bx_\perp)\mid\bz)=0$ whenver $\EE(f'(\bx_\perp))=0$.
\end{enumerate}
\vshrink{-0.5em}
\ref{it:pc} shows that in proxy control problems, $\bar\bx,\bar\bz$ summarize all observable information about the latent confounder $\bu$; thus, %
they may have much lower dimensions than $\bx,\bz$. As an example, in the scenario of \cref{fig:cg}~(c) 
we can have $\dim\bar\bx\le \dim\bar\bv+\dim\bt.$ 
For general problems, 
\ref{it:falsifiable} shows that, under \eqref{eq:cmm-defn}, any hypothesis $f_0(x)$ that is {\em falsifiable} \citep{corfield2009falsificationism} from data must depend on $\bx$ only through $\bar\bx$; additional components of the form $f'(\bar\bx)f''(\bx_{\perp})$ will be indistinguishable from noise. Thus, it is intuitive that $\gtRKHS[\alpha]$ may encode desirable inductive bias for estimation. We provide additional discussion in App.~\ref{app:disc-gtRKHS}.  
\end{example}

To provide further intuition on the regularity of $\gtRKHS[\alpha]$, we
extend the example by combining it with the classic torus example in literature \citep[e.g.,][]{chen_rate_2007}:
\begin{example}[compositional torus model]\label{ex:main-torus} In the setting of Ex.~\ref{ex:main}, 
let $\bar\bz$ be uniformly distributed on the torus $\mb{T}^{d_l}$, and $\bar\bx\in\mb{T}^{d_l}$ be s.t.$$\EE(f(\bar\bx)\mid\bar\bz=\bar z)=\int_0^1\ldots\int_0^1\prod_{i=1}^{d_l} k(\bar x_i-\bar z_i)f(\bar x_i)d\bar x_i,$$ where $k$ is a 1-periodic even function. In words, $\bar\bx$ is obtained by adding to $\bar\bz$ an independent noise. Its marginal distribution is also uniform.

Let the $j$-th Fourier coefficient of $k$ be $c_j \asymp (1+j)^{-d_l p}$, for some constant $p>0$. Then we have $\bar s_j \asymp j^{-p}$, and 
$$
\gtRKHS[\alpha] = \{\bar f\circ f_{enc,x}, \bar f\in W^{\alpha pd_l, 2}_{per}\}, 
$$ %
where $ W^{\alpha pd_l, 2}_{per} $ denotes a periodic Sobolev space; when $\alpha pd_l\in\mb{N}$, it is equivalent to the manifold Mat\'ern RKHS \citep{borovitskiy2020matern}. 
\end{example}

We note that %
a similar connection exists for (the manifold analogy of) Gaussian noise, 
in which case the resulted space for $\bar f$ equals the manifold Gaussian RKHS \citep{de_vito_reproducing_2019}. 

%% file: sec-estimability.tex
\vshrink{-0.5em}
\section{BOUNDED ILL-POSEDNESS WITH LEARNED REPRESENTATIONS}\label{sec:estimability}
\vshrink{-0.5em}

As we reviewed in Section 2, knowledge of the exact SVD of $E$ would facilitate sample-efficient estimation of \eqref{eq:cmm-defn}. In reality, however, %
it is only possible to approximately recover the SVD: 
we can only approximately optimize certain variational objectives of it, as %
we can only observe $E$ indirectly through the finite samples $\{(z_i,x_i)\}$. 
In this section, we show that 
approximate optima of certain variational objective can still enable
efficient estimation of \eqref{eq:cmm-defn}, and that thanks to the kernelized formulation of conditional moment models, the estimation error can have a robust dependency on the approximation quality for the SVD. 

\vshrink{-0.3em}
\subsection{The Basis Learning View and a Catastrophic Transition of Estimation Error}
\vshrink{-0.3em}

Before introducing our main results, 
let us first consider a \emph{na\"ive approach} that simply uses the learned singular vectors to define the basis of the hypothesis space. The singular vector estimates $\{\psi_i\}_{i=1}^{J_n}$ can be parameterized with DNNs, and trained with the ``variance explained'' formulation of SVD, so that for some $\epsilon_n\to 0$, it has
\begin{equation}\label{eq:svd-approx-opt}
\sum_{i=1}^{J_n} \|E\psi_i\|_2^2 
 > \sum_{i=1}^{J_n} \|E\gtsv{i}\|_2^2 - \epsilon_n^2,
\end{equation}
Following this, we could define the learned hypothesis space as $\cH := \mrm{span}\{\psi_1,\ldots,\psi_{J_n}\}$, and use it to instantiate the classical sieve estimator \citep{chen_rate_2007}. %

An important limitation of this approach is that it does not learn an \textbf{inner product structure}. %
While this did not affect convergence in the classical setting where we had access to the exact singular vectors, 
it becomes problematic now, as the learned $\{\psi_i\}$ can only be approximations in the weak sense of \eqref{eq:svd-approx-opt}. 
To see this, 
consider the $L_2$ error bound 
\begin{align*}
\|\hat f_n - f_0\|_2 &\le \left(\sup_{f\in\cH}\frac{\|f-f_0\|_2}{\|E(f-f_0)\|_2}\right)\|E(\hat f_n-f_0)\|_2 \\ 
&= \|E\psi_{J_n}\|_2^{-1} \|E(\hat f_n-f_0)\|_2,\numberthis\label{eq:naive-L2-bound}
\end{align*}
where $\|E\psi_{J_n}\|_2^{-1}$ can be viewed as an \textbf{ill-posedness measure}. 
It is easy to show that 
\begin{equation}\label{eq:psiJ}
\|E\psi_{J_n}\|_2^{-1} < (\max\{0, \|E\bar\psi_{J_n}\|_2^2 - \epsilon_n^2\})^{-1/2}.
\end{equation}
Observe that when $\epsilon_n^2 \le c\|E\bar\psi_{J_n}\|_2^2$ for some $c<1$, both \eqref{eq:psiJ} and \eqref{eq:naive-L2-bound} can match the order of the respective bounds for the exact basis $\{\bar\psi_j\}$. In this regime, these bounds can be order-optimal, as we discuss in \cref{ex:main-hypo}. 

Unfortunately, the bound \eqref{eq:psiJ} blows up as $\epsilon_n^2\to \|E\gtsv{J_n}\|_2^2$, {\em yet it continues to be tight.} To see this, consider  
$\{\psi_j\} := \{\bar\psi_1,\ldots,\bar\psi_{J_n-1},\bar\psi_{J_n+M}\}$, and let $M\to\infty$. %
Thus, as soon as $\epsilon_n^2$ reaches $\|E\bar\psi_{J_n}\|_2^2$, we lose any guarantee for $\|E\psi_{J_n}\|_2^{-1}$ and subsequently the $L_2$ error. 
(We can easily construct a sequence of 
approximate maxima for \eqref{eq:popu-obj}, %
for which $\|f-f_0\|_2$ can get arbitrarily large.)  
This is in stark contrast to the previous regime, %
and demonstrates 
a catastrophic transition behavior for this na\"ive %
approach. It is especially concerning because without resorting to extremely conservative choices of $J_n$, we cannot expect to estimate $\epsilon_n^2$ or $\|E\bar\psi_{J_n}\|_2$ reliably from data; the best we may hope for is to guarantee that $\epsilon_n^2$ is of an optimal order of magnitude, which, due to the transition behavior, cannot provide any guarantees on the estimation error. %

\vshrink{-0.3em}
\subsection{Bounded Ill-posedness with Learned Kernels}\label{sec:est-klearn}
\vshrink{-0.3em}

The above pathology can be remedied,  
if we additionally learn to approximate a carefully chosen \textbf{inner product structure}. Concretely, suppose $\cH$ can be defined as 
\begin{equation}\label{eq:H-defn}
\cH := \Ran\,\tilde T^{\alpha/2},~
\|f\|_{\cH} := \|\tilde T^{-\alpha/2} f\|_2,
\end{equation}
where the trace-class operator $\tilde T$ satisfies 
\begin{equation}\label{eq:hs-norm} %
\tilde T := \tilde E^\top \tilde E, ~~\text{where}~~
\|\tilde E - E\|_{\mathrm{HS}} \le \epsilon_n,
\end{equation}
\eqref{eq:hs-norm} is an intuitive error criterion for spectral decomposition; it also implies \eqref{eq:svd-approx-opt} holds for the basis of $\cH$ (see App.~\ref{app:reproof-loss-equivalence}). 
As we discuss in App.~\ref{app:proof-lem-approx-hs}, it can be relaxed %
to provide further flexibility. %

Let us provide some intuition for the learned inner product structure $\<\cdot,\cdot\>_\cH$. 
From \eqref{eq:H-defn} and \eqref{eq:hs-norm} we can see it approximates that of \eqref{eq:ideal-rkhs}, which   
can be viewed as weighing the $L_2$ orthonormal basis $\{\bar\psi_j\}$ using the singular values $\bar s_j^\alpha$. Thus, we are imposing stronger regularization in directions where %
\eqref{eq:cmm-ip} or \eqref{eq:cmm-defn} provides less information about $f_0$. 

We first present an approximation bound. In particular, it applies to $f_0$ satisfying \eqref{eq:general-source-cond}.
\begin{lemma}[Proof in App.~\ref{app:proof-lem-approx-hs}]\label{lem:approx-hs}
Let $\cH$ be defined as above. Then, for any $\beta\ge\alpha$ and $\bar f\in\gtRKHS[\beta]$, there exists $\tilde f\in\cH$ s.t.
\begin{align}\label{eq:lem-approx-hs-claim}
\!\!\!\!\|\tilde f\|_\cH \le c_\alpha\|f_0\|_{\gtRKHS[\beta]},\,
\|\bar f - \tilde f\|_2 \le c_\alpha\epsilon_n^{\min\{\alpha, 1\}}\|\bar f\|_{\gtRKHS[\beta]},\!
\end{align}
where $c_\alpha$ is a universal constant determined by $\alpha$. 
\end{lemma}

We now bound the modulus of continuity:
\begin{theorem}[Proof in App.~\ref{app:proof-thm}]\label{thm:bounded-ill-posedness}
Let $\cH,f_0$ be defined as above, $\tilde f\in\cH$ be the approximation to $f_0$ in Lem.~\ref{lem:approx-hs}, and $\{\delta_n\},\{\lambda_n\}\subset\RR_+$ be arbitrary sequences of numbers. Then a ``regularized moduli of continuity'' satisfies 
\begin{align*}
\omega_n'^2 &:= \sup\{\|f-f_0\|_2^2: f\in\cH,
\\ & \hspace{5.8em}
\|E(f-f_0)\|_2^2 + \lambda_n \|f\|_\cH^2 \le \delta_n^2\} \numberthis \label{eq:moc-reg} 
\\ 
 &\phantom{:}= \cO\big(
\|\tilde f - f_0\|_2^2 + 
    (\lambda_n^{-\frac{1}{\alpha+1}} + \lambda_n^{-1}\epsilon_n^{2\alpha})  \cdot\\  
&\hspace{4.3em}
(\delta_n^2 + \lambda_n\|f_0\|_{\gtRKHS[\beta]}^2 + \|E(\tilde f - f_0)\|_2^2)
\big). \numberthis\label{eq:thm-bip-claim}
\end{align*}
\end{theorem}

Let us compare the result with \eqref{eq:psiJ}, which also describes a measure of ill-posedness. The above bound is less clean, as the approximation errors are harder to separate. However, %
if we ignore it for the moment and restrict to $\lambda_n\|f_0\|_{\gtRKHS[\beta]}^2 =\cO(\delta_n^2)$, the measure of ill-posedness reduces to 
\begin{equation}\label{eq:simplified-measure}
\frac{\omega_n'^2}{\delta_n^2} = \cO(\lambda_n^{-1/(\alpha+1)} + \lambda_n^{-1}\epsilon_n^{2\alpha}),
\end{equation}
which can be compared with \eqref{eq:psiJ}. We can then instantiate \eqref{eq:simplified-measure} and \eqref{eq:psiJ} in the context of our running example:
\begin{example}[Ex.~\ref{ex:main-torus}, continued]\label{ex:main-hypo}
Suppose $f_0$ satisfies \eqref{eq:general-source-cond} with $\beta={b}/{2p}$, $\alpha=\beta$, where $b>0$ is any constant; and %
$\|E(\hat f_n - f_0)\|_2$ can be order-optimal. In such cases, the various quantities in the kernelized and sieve estimators would satisfy \citep{chen_rate_2007,wang2021quasibayesian}
\begin{equation}\label{eq:optimal-hps}
J_n \asymp n\lambda_n \asymp n \|E(\hat f_n-f_0)\|_2^2\asymp  n^{1/(b+2p+1)}.
\end{equation}
Thus, if $\epsilon_n\le C n^{-{p}/{(b+2p+1)}} \asymp \bar s_{J_n}$, and the constant $C$ is sufficiently small, 
both \eqref{eq:psiJ} and \eqref{eq:simplified-measure} would have the same order, and %
lead to the minimax $L_2$ rate 
$
\|\hat f_n - f_0\|_2^2 \le \cO(n^{-b/(b+2p+1)}).
$
However, when $C$ reaches $n^{p/(b+2p+1)}s_{J_n}=\cO(1)$, 
the bound \eqref{eq:psiJ} will blow up, whereas \eqref{eq:simplified-measure} continues to be order-optimal. 
\end{example}

The above discussion is hypothetical, as it relies on $\epsilon_n^2$ and $\|E(\hat f_n-f_0)\|_2$ being optimized at a certain rate. 
As we establish in App.~\ref{app:additional-results}, 
The requirement on $\epsilon_n^2$ can often be met. However, technicalities in constructing $\cI$ will prevent us to match the exact order of \eqref{eq:optimal-hps}, and the rates we established for CMM will be slightly worse. Nonetheless, the discussion provides valuable intuition that {\em kernel learning for CMM is a robust process}, as the $L_2$ error degrades gracefully as $\epsilon_n^2$ deteriorates. 

%% file: sec-svd.tex
\vshrink{-0.5em}
\section{SPECTRAL REPRESENTATION LEARNING FOR CMM}\label{sec:estimation}
\vshrink{-0.5em}

We now derive a spectral decomposition algorithm, which learns a linear representation parameterized with DNNs, and fulfills the condition \eqref{eq:hs-norm} with high precision. While a number of approaches exist for spectral decomposition, their statistical guarantees remain largely unclear. 
In this section, we first establish that a contrastive learning algorithm, first proposed in \citet{haochen_provable_2021} for self-supervised learning, allows us to establish fast-rate convergence for the spectral decomposition problem, which in turn leads to fast-rate bounds for \eqref{eq:hs-norm}. 
Using the learned representation and its guarantees,  
we then establish $L_2$ error bounds for the CMM estimator \eqref{eq:minimax-estimator}. 
Finally, we discuss the implementation of the complete algorithm. %

\vshrink{-0.4em}
\subsection{Fast-Rate Spectral Representation Learning}\label{sec:srl}
\vshrink{-0.4em}

\citet{haochen_provable_2021} employs the population objective 
$$
\popuLoss[h]  := 2\EE_{P_{zx}} h(\bz,\bx) - \EE_{P_z\otimes P_x} h^2(\bz,\bx) - 1
$$
which 
was motivated as minimizing \eqref{eq:hs-norm} on finite neighbor graphs. We first establish similar results in our setting, and provides more insights for the objective:

\begin{lemma}[Proof in \cref{app:reproof-loss-equivalence}]\label{lem:loss-equivalence}

Let $J\in\mb{N}, \epsilon_n>0$ be arbitrary. 
Let $\Phi:\cZ\to\RR^J, \Psi:\cX\to\RR^J$ be arbitrary vector-valued functions, s.t.~$h(z,x)=\Phi(z)^\top\Psi(x)$ is $L_2(P_z\otimes P_x)$-integrable. 
Define $(\tilde E f)(\cdot) := \int f(x) h(\cdot,x) P(dx)$. 
Then we have 
\begin{align*}
\|\tilde E - E\|_{\mathrm{HS}} \le \epsilon
&\Leftrightarrow \popuLoss[h] \ge \popuLoss[h_0] - \epsilon_n^2
\\
&\Leftrightarrow \|h-h_0\|_{L_2(P_z\otimes P_x)} \le \epsilon_n 
,
 \numberthis\label{eq:loss-equiv-1}
\end{align*}
where $h_0 := \frac{dP_{zx}}{d(P_z\otimes P_x)}$ is the Radon-Nikodym derivative. And we have 
\begin{align*}
\sup_{h\in L_2(P_z\otimes P_x)} \popuLoss[h] &= \popuLoss[h_0] 
= \cD_{\chi^2}(P_{zx}\Vert P_z\otimes P_x) \\ 
&= \mrm{Tr}(E^\top E) - 1. \numberthis\label{eq:chisq-trace}
\end{align*}
\end{lemma}

\eqref{eq:loss-equiv-1} above is a slight generalization of \citet[Lemma 3.2]{haochen_provable_2021}. It 
shows that approximate optimization of $\popuLoss[\cdot]$ leads to an RKHS $\cH$ that fulfills the condition \eqref{eq:hs-norm}; its reproducing kernel equals (see \cref{app:popu-kernel-deriv})
\begin{equation}\label{eq:popu-kx}
\!\!k_{x,p}(x,x') = \hat\Psi_n^\top(x) \Sigma_x^{-\frac{1}{2}}(\Sigma_x^{\frac{1}{2}}\Sigma_z\Sigma_x^{\frac{1}{2}})^\alpha\Sigma_x^{-\frac{1}{2}} \hat\Psi_n(x'),
\end{equation}
where $\hat\Phi_n^\top(z)\hat\Psi_n(x)=:\hat h_n(z,x)$ defines the approximate optimum,  
$\Sigma_x = \EE_{\bx}\hat\Psi_n(\bx)\hat\Psi_n^\top(\bx)$ denotes the uncentered covariance, %
and $\Sigma_z$ is defined similarly using $\hat\Phi_n(\bz)$. In practice, we use its empirical approximation, 
\begin{equation}\label{eq:kernel-used}\tag{\ref{eq:popu-kx}'}
\!\!k_{x}(x,x') = \hat\Psi_n^\top(x) \hat\Sigma_x^{-\frac{1}{2}}(\hat\Sigma_x^{\frac{1}{2}}\hat\Sigma_z\hat\Sigma_x^{\frac{1}{2}})^\alpha\hat\Sigma_x^{-\frac{1}{2}} \hat\Psi_n(x'),
\end{equation}
where $(\hat\Sigma_x,\hat\Sigma_z)$ are plug-in estimates for $(\Sigma_x,\Sigma_z)$. {\em For theoretical analyses}, we assume $(\hat\Sigma_x,\hat\Sigma_z)$ are estimated using $\cO(n)$ additional i.i.d.~samples. 

\eqref{eq:chisq-trace} further connects spectral decomposition and kernel learning to a $\chi^2$ divergence. %
In particular, it shows the operator $E^\top E$ is trace-class if and only if the $\chi^2$-divergence is finite. %
This is the sufficient and necessary condition for the %
space $\gtRKHS[1]$ to have an RKHS structure \citep{steinwart_mercers_2012}, as we have also 
noted in \cref{sec:setup}. 
More generally, we always have %
$\cD_{\chi^2}(P_{zx}\Vert P_z\otimes P_x)+1=\mrm{Tr}(E^\top E) = \mrm{Var}\:\bar k_x(\bx,\bx)$, where $\bar k_x$ is the kernel for $\gtRKHS[1]$; the variance term on the right-hand side often appears in %
the analysis of %
kernel methods, and can be viewed as a regularity measure. 
When $\alpha=1$, 
$(\gtRKHS[1], \bar k_x)$ is the approximation target of $(\cH, k_x)$; thus, \eqref{eq:chisq-trace} provides an intuitive condition for the well-definedness of the kernel learning process.

\cref{lem:loss-equivalence} concerns the approximate optimum of the population loss $\popuLoss[\cdot]$. %
In practice, we can only optimize %
its finite-sample counterpart.  
The following proposition bounds the estimation error incurred in the process:

\newcommand{\prodHypoSpace}{\cF_h}

\begin{proposition}[Proof in App.~\ref{app:proof-est}]\label{prop:dr-est}
Let $B,C_1>0, q\in (0,1]$ be constants, $h^*\in L_2(P_z\otimes P_x)$ be an arbitrary function. 
Suppose the hypothesis space $\prodHypoSpace$ is star-shaped around $h^*$, and 
$\sup_{h\in\prodHypoSpace}\|h\|_\infty\le B$, $\log N(\prodHypoSpace,\|\cdot\|_\infty,\epsilon_n) \le C_1\epsilon_n^{-2q}$. Let $\{(\tilde z_i, \tilde x_i)\}_{i=1}^n$ be $n$ i.i.d.~training samples from $P_{zx}$, and $
\hat h_n := \arg\max_{h\in\prodHypoSpace} \cD_n(h)$, %
where
\begin{equation}\label{eq:empirical-risk}\!\!
\cD_n(h) := 
\frac{2}{n}\sum_{i=1}^n h(\tilde z_i, \tilde x_i) - \frac{1}{n(n-1)}\sum_{i\ne j} h(\tilde z_i, \tilde x_j)^2
\end{equation}
is the empirical risk. %
Then %
there exists some $C_q>0$ s.t.~for all $\zeta>0$, we have, w.p.~$\ge 1 - \zeta$, 
\begin{align*}
\popuLoss[h_0] \:-&\: \popuLoss[\hat h_n] \le C_q \Big( \\
&\!\!\!\popuLoss[h_0] -\popuLoss[h^*] + B^4 n^{-\frac{1}{1+q}} + 
B^4\frac{\log \zeta^{-1}}{n}
\Big).
\end{align*}
\end{proposition}

\begin{remark}
\cref{prop:dr-est} provides fast-rate guarantees for the estimation (i.e.,~generalization) error for $\popuLoss$.\footnote{
\citet{haochen_provable_2021} established a slow rate result in their setting, which saturates at $\Omega(n^{-1/2})$ and is less interesting in light of the discussion below. Their main results were developed under structural assumptions on their neighbor graph.
}
When $\cF_{h}$ admits the factorized structure in \cref{lem:approx-hs}, the result 
is equivalent to fast-rate guarantees for spectral decomposition as measured in  $\|\cdot\|_{\mrm{HS}}$. 
The $o(n^{-1/2})$ convergence rate is interesting, as it 
contrasts existing approaches %
based on explicit {\em normalization constraints}: e.g., constraining the ``eigenfunction estimates'' $\Psi$ to be orthogonal \citep{pfau_spectral_2020,Neuralef}, or at least having a covariance matrix with a bounded spectral norm. It would be %
more challenging to establish similar guarantees for those methods, because their constraints can only be enforced at a precision of $\cO(\sqrt{J_n/n})$, which is the error of covariance estimation. 

The evaluation of our kernel \eqref{eq:popu-kx} also involves covariance estimation. 
However, at least when $\alpha\in\{1,2\}$, its plug-in estimate \eqref{eq:kernel-used} may enjoy the same guarantees (Prop.~\ref{prop:kern-est}). From the perspectives of spectral decomposition and general representation learning, \cref{prop:dr-est} can also be interpreted as a guarantee for the learned $\Psi$ to approximate the subspace spanned by the leading singular vectors, as the Hilbert-Schmidt norm guarantee implies a similar guarantee for the ``variance explained'' objective \eqref{eq:svd-approx-opt} (App.~\ref{app:reproof-loss-equivalence}). 
\end{remark}

To obtain full guarantees for $\epsilon_n^2$, 
it remains to bound the approximation error $\popuLoss[h_0] - \popuLoss[h^*]$. 
In \cref{app:additional-results} we 
provide a generic approximation result, and instantiate it on our running example. As we will see, when using DNN models for $\cF_{h}$, the combined error rate $\epsilon_n^2$ can still be $o(n^{-1/2})$, and is adaptive to the dimensionality of the informative latent variables introduced in Example~\ref{ex:main}. %

\vshrink{-0.4em}
\subsection{Results for Conditional Moment Models}\label{sec:cmm}
\vshrink{-0.4em}

We then turn to consistency guarantees for the estimator \eqref{eq:minimax-estimator}, in which $\cH,\cI$ are instantiated using the learned representations: 
we define $\cH$ using the kernel in \eqref{eq:kernel-used}, in which $\hat h_n$ %
is defined as in \cref{prop:dr-est}. For simplicity, we define $\cI$ using a symmetric choice of kernel:
\begin{equation}\label{eq:kz-used}
k_z(z,z') := \hat\Phi_n^\top(z) \hat\Sigma_z^{-\frac{1}{2}}(
     \hat\Sigma_z^{\frac{1}{2}}\hat\Sigma_x\hat\Sigma_z^{\frac{1}{2}})^{\alpha'}\hat\Sigma_z^{-\frac{1}{2}} \hat\Phi_n(z').
\end{equation}
When $\alpha'=\alpha+1$, 
$k_z$ will approximate %
the ``optimal instrument'' kernel for $\cH$. For technical reasons, however, our analysis assumes $\alpha'=1$. See App.~\ref{app:disc-cmm-results} for details.

We now present the $L_2$ convergence result. For brevity, below we state the result in the setting of our running example; a more general result can be found in Appendix~\ref{app:proof-cmm-thm}. 
\begin{theorem}[Proof in App.~\ref{app:proof-cmm-thm}]\label{thm:cmm}
Let $\bx,\bz,f_{enc,x},f_{enc,z}$ be defined as in Ex.~\ref{ex:main}-\ref{ex:main-torus}. 
Suppose $p > 2$, there exists some $f_0$ that satisfies \eqref{eq:general-source-cond} with $\beta=b/2p$, and $f_{enc,x}, f_{enc,z}$ are $\beta_d$ H\"older regular \citep[see e.g.,][]{gine2021mathematical} with $2\beta_d \ge (p-2)\max\{\dim\bz,\dim\bx\}$. Let $\alpha\in\{1,2\}\le\beta$, and 
\begin{itemize}[leftmargin=*]
\item $\hat h_n(z,x) = \hat\Phi_n^\top(z)\hat\Psi_n(x)\in\cF_h$ be the empirical risk minimizer for \eqref{eq:emp-obj-expr} ($\alpha=1$), or the solution to the constrained optimization problem \eqref{eq:constrained-opt-problem} in appendix ($\alpha=2$), using a DNN model for $\cF_h$ (see appendix),
\item $(k_x,\cH,k_z,\cI)$ be defined by 
\eqref{eq:kernel-used}, \eqref{eq:kz-used} using $\alpha'=1$, $\alpha\in\{1,2\}$ and $\hat h_n$, with $(\hat\Sigma_x,\hat\Sigma_z)$ defined using another $n$ samples $\Dcov$, 
\item $\hat f_n$ be defined by \eqref{eq:minimax-estimator} using $\cH,\cI$, %
a third set of $n$ i.i.d.~samples $\Dkiv:=\{(z_i,x_i,y_i)\}$, and $(J_n,\nu_n,\lambda_n)$ be chosen as in \cref{app:proof-cmm-thm}. 
\end{itemize}
Then we have, w.p.~$\ge 1-n^{-10}$, 
$$
\|\hat f_n - f_0\|_2 = \tilde\cO\big(n^{-\frac{2\alpha(p-2)}{(\alpha+1)(4p+1)}}(1+\|f_0\|_{\gtRKHS[\beta]}^{\frac{2\alpha}{\alpha+1}})\big),
$$
where $\tilde\cO$ hides all sub-polynomial factors. 
\end{theorem}

\begin{remark}\label{rmk:thm}
\cref{thm:cmm} provides an $L_2$ rate of convergence without any preimposed condition on measures of ill-posedness. %
Under its conditions, \citet{chen_rate_2007} establishes the minimax lower rate of $\Omega(n^{-\frac{\alpha p}{((1+\alpha)2p+1)}})$, which can be realized assuming knowledge of \eqref{eq:ideal-rkhs}. %
Note that both rates approach the order of $n^{-\alpha/2(\alpha+1)}$ as $p\to\infty$. Also note that 
under %
\eqref{eq:general-source-cond}, both rates only depend on $p$, and adapt to the typically lower dimensionality of the informative latent variables (Ex.~\ref{ex:main-torus}). This is in contrast to what can possibly be established for estimators based on fixed-form kernels; see \cref{app:disc-cmm-results} for more discussion. 

To achieve similar $L_2$ rates, the only alternatives would be to employ Landweber or ridge regularization \emph{in the $L_2$ space} \citep{carrasco2007linear}, which we discuss in more detail in 
App.~\ref{app:disc-cmm-estimators}. Briefly, the latter strategy cannot adapt to $\beta > 1$, whereas the %
application of the former appears difficult in high dimensions, as it requires representing $\cH$ using a predetermined orthonormal basis. %
Hyperparameter selection for these approaches will also be less trivial and cannot be based on the comparison of \eqref{eq:popu-obj}.\todo{for submission, write down the calculations}
In aggregate, our kernel learning approach appears the most viable in achieving similar guarantees, if one wishes to benefit from the use of DNN models.
\end{remark}

\vshrink{-0.4em}
\subsection{Algorithm Implementation}\label{sec:algo-impl}
\vshrink{-0.4em}

\paragraph{The algorithm}
We instantiate the estimator \eqref{eq:minimax-estimator} using the learned kernels, %
and estimates the learned representation $(\hat\Psi_n,\hat\Phi_n)$ and $\hat f_n$ on separate partitions of the training data. 
Our theoretical discussions have assumed the variables $\bx,\bz$ do not overlap, and employed the kernels of \eqref{eq:kernel-used}, \eqref{eq:kz-used}. In practice, we use the same kernels, but impose fewer restrictions: we consider arbitrary values of $\alpha$, use $\alpha'=\alpha+1$, and reuse samples when estimating the covariance terms in $(k_x,k_z)$; the process is 
summarized in \cref{alg:klearn}. When $\bx$ and $\bz$ overlap, we modify the algorithm to learn product kernels, as we will describe shortly. 

\vshrink{-0.5em}
\begin{algorithm}[th]
\caption{Kernel learning by spectral decomposition}\label{alg:klearn}
\begin{algorithmic}
\REQUIRE $\{(\tilde z_i,\tilde x_i)\}_{i=1}^n$, hyperparameters $J_n, \alpha$, model $\cF_h = \{h = \Phi_h(z)^\top \Psi_h(x): \Phi_h\in \cF_{\Phi}, \Psi_h\in\cF_{\Psi}\}$
\ENSURE learned kernels $k_x,k_z$
\STATE Let $\hat h_n = \hat\Phi_n(z)^\top\hat\Psi_n(x)$ minimizes $\cD_n$ in \eqref{eq:emp-obj-expr}
\STATE $\hat\Sigma_x \gets \frac{1}{n}\sum_{i=1}^n \hat\Psi_n(\tilde x_i)\hat\Psi_n^\top(\tilde x_i)$
\STATE $\hat\Sigma_z \gets \frac{1}{n}\sum_{i=1}^n \hat\Phi_n(\tilde z_i)\hat\Phi_n^\top(\tilde z_i)$
\RETURN $k_x(x,x') := \hat\Psi_n'(x)\hat\Sigma_x^{\frac{\alpha-1}{2}}\hat\Sigma_z^\alpha\hat\Sigma_x^{\frac{\alpha-1}{2}}\hat\Psi_n(x')$, \\ 
$k_z(z,z') := \hat\Phi_n'(z)\hat\Sigma_z^{\frac{\alpha}{2}}\hat\Sigma_x^{\alpha+1}\hat\Sigma_z^{\frac{\alpha}{2}}\hat\Phi_n(z')$
\end{algorithmic}
\end{algorithm}
\vshrink{-0.5em}

The hyperparameters $\alpha, J_n$ and the regularization hyperparameters $\lambda_n, \nu_n$ in \eqref{eq:minimax-estimator} can be determined using the established process in kernelized IV estimation \citep{singh_kernel_2020,muandet_dual_2020,dikkala_minimax_2020}. The hyperparameters used in learning $\hat h_n$, such as the DNN architecture and optimization hyperparameters, are determined by evaluating the learning objective $\cD_n$ on a held-out dataset. As $\hat h_n$ has been trained on a subset of training data, we simply use the other subset for validation. 

\vshrink{-0.5em}
\paragraph{Extensions for overlap variables}
In many applications, the variables $\bx,\bz$ may have overlap, meaning that there exists $(\bx',\bz',\bw)$ s.t.~$\bx=(\bx',\bw), \bz=(\bz',\bw)$. For example, in proximal causal inference, $\bw$ will contain the treatment variable. In such cases, we can construct product kernels for \eqref{eq:minimax-estimator}: let $k_{x'}, k_{z}$ be the kernels returned by \cref{alg:klearn}, on the dataset $\{(\tilde z_i, \tilde x'_i)\}$. Then we can use $k_z$ to define $\cI$, and the following kernel for $\cH$: $$
k_x'((x_1', w_1), (x_2', w_2)) := k_{x'}(x_1', x_2')k_w(w_1,w_2),
$$
where $k_w$ can be a fixed-form kernel, which is appropriate for $\bw$ with moderate dimensions. We implement this algorithm in our experiments. 
When $\bw$ is high-dimensional, we can learn product kernels by feeding the above $k_{x'}$ to the extended algorithm in \citet[App.~G]{wang2022fast}.

The product kernel constructions are related to the product basis models which have appeared in literature \citep[e.g.,][]{kato_quasi-bayesian_2013}. Intuitively, more flexibility can be  
expected in specifying parts of the hypothesis space that are related to $\bw$, as they are in some sense less affected by the ill-posedness issue: 
we have $\EE(g(\bw)\mid\bz)=g(\bw)$ for all $g\in L_2(P_w)$. %

%% file: sec-related-work.tex
\vshrink{-0.5em}
\section{RELATED WORK}\label{sec:related-work}
\vshrink{-0.5em}

A series of works have developed estimators with a form similar to \eqref{eq:minimax-estimator}, 
as we briefly discussed in Section 2. 
App.~\ref{app:disc-cmm-estimators} reviews additional literature and possible strategies for achieving $L_2$ consistency. %
To our knowledge, %
only \citet{liao_provably_2020} provided an alternative result that allows the use of adaptive ML models (e.g., DNNs) and does not rely on preimposed assumptions on measures of ill-posedness. They have relied on $L_2$ regularization, and cannot adapt to $\beta>1$, in contrast to our result (\cref{rmk:thm}, \cref{app:disc-cmm-estimators}). \todo{in the next submission, add simulations and point to them}
Note that 
a separate line of work studies the use of ML for linear IV problems with fixed dimensionality, which is a fundamentally different setting %
(\citealp{singh2020machine}, p.~6; \citealp{foster2019orthogonal}, p.~8). %

Our analysis for spectral representation learning (Sec.~\ref{sec:srl}) is connected to recent theoretical works on multi-view learning \citep{arora_theoretical_2019,lee_predicting_2020,tosh_contrastive_2021}, which views $\bz,\bx$ as conditionally independent views of a latent variable. The connection between spectral decomposition and contrastive representation learning is also being recognized \citep{haochen_provable_2021,balestriero_contrastive_2022}. 
App.~\ref{app:disc-repr-learning-lit} discusses this literature in more detail. Briefly, our results are unique in 
relating representation learning to divergence maximization and kernel learning, and in 
establishing fast-rate convergence in a concrete high-dimensional setting (App.~\ref{app:additional-results}); in particular, there had been a lack of understanding of approximation guarantees. 
Note that 
our results for conditional moment models (Sec.~\ref{sec:estimability}, \ref{sec:cmm}) are independently developed, and constitute a separate contribution. 

\citet{michaeli_nonparametric_2016} studied the estimation of SVD of $E$, using kernel density estimates; also related are \citet{klus_eigendecompositions_2020,kostic2022learning} who studied the connected but different problem of eigendecomposition (of similar non-self-adjoint operators) using RKHS methods. 
As we discuss in App.~\ref{app:disc-cmm-results}, using fixed-form kernels or similar nonparametric models for estimation is often inapplicable for high-dimensional data, due to a provable lack of adaptivity. However, there are also scenarios that specifically require the decomposition of similar conditional expectation operators, restricted to prescribed RKHSes, such as in the estimation of the conditional mean embedding operator \citep{park_measure-theoretic_2021,talwai_sobolev_2022}. 

Concurrent to our work, \citet{johnson2022contrastive,deng2022neural} also studied connections between contrastive learning, spectral decomposition, and kernel machines, and the use of kernels similar to our $\gtRKHS[1]$ in a generic self-supervised learning setup. However, neither work provided statistical guarantees for the kernel learning process or the use of learned kernels in downstream tasks. 

%% file: sec-exp.tex
\vshrink{-0.6em}
\section{EXPERIMENTS}\label{sec:exp}
\vshrink{-0.6em}

We evaluate our method on synthetic and semi-synthetic datasets. The synthetic experiment evaluates the convergence of the representation learning objective and the $L_2$ error for conditional moment models; 
for space reasons, we defer it to Appendix~\ref{app:exp-synth}. Below 
we describe the semi-synthetic experiment for proximal causal inference.

Our setup is adapted from \citet{mastouri_proximal_2021} %
who fitted generative models on three real-world datasets, and used simulation from the model for evaluation. 
This design allows us to compare the estimated causal effect to the ``ground truth'' in the fitted generative model, %
even though both may be different from the (unobservable) causal effect in the real world.
To emulate a high-dimensional setup, we modify the proxy variables in their datasets by concatenating them with $D_{ex}$ independent random variables, and transforming them with feedforward DNNs. 

As baselines, we instantiate the estimator \eqref{eq:minimax-estimator} with DNN (\texttt{AGMM-NN}) and fixed-form kernel (\texttt{Kernel-RBF}) models. The baselines have established guarantees (for the population objective \eqref{eq:popu-obj}), and competitive empirical performance \citep{dikkala_minimax_2020}; the kernel baseline is also similar to those employed in \citet{mastouri_proximal_2021}. %

\begin{figure}[ht]
    \centering
    \includegraphics[width=0.675\linewidth,clip,trim={0.1cm 0.35cm 0cm 0.2cm}]{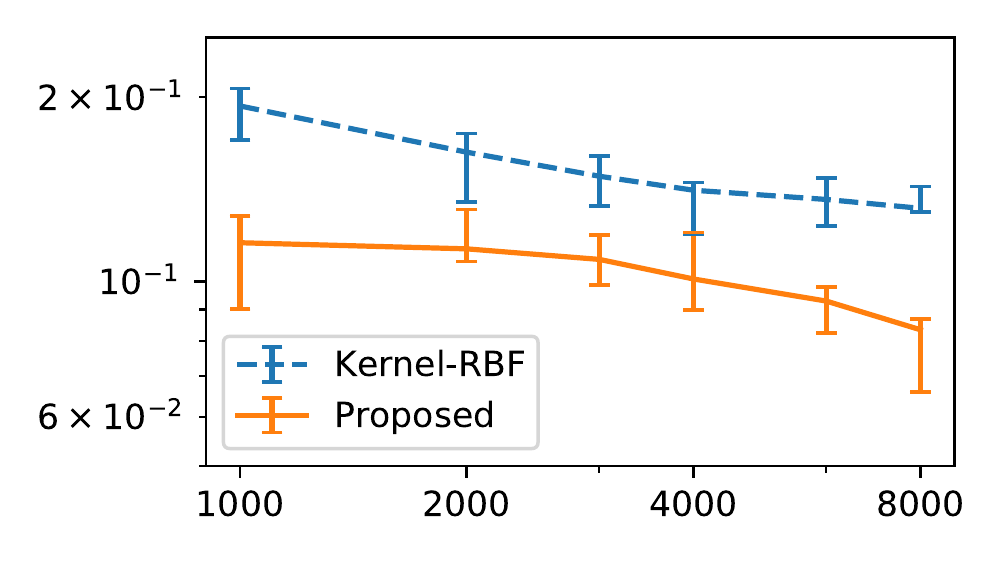}
\vshrink{-0.5em}
    \caption{Semi-synthetic experiment: test MAE for the abortion and crime data, for $D_{ex}=256$ and varying $N$. We omit \texttt{AGMM-NN} which is uncompetitive in this setting. Error bar indicates the $25\%$ and $75\%$ percentiles.
    }
    \label{fig:proxy-varying-N}
 \vshrink{-1em}
\end{figure}

We report a representative subset of results in \cref{tab:proxy_experiment} and \cref{fig:proxy-varying-N}. As we can see, our method has strong performance, especially in high dimensions. Due to space limits, we defer the full experiment setup and results to \cref{app:exp-pc}.

\begin{table}[h]
    \centering\small
    \begin{tabular}{cccc}
    \toprule
       $D_{ex}$   & AGMM-NN & Kernel-RBF & Proposed \\ \midrule
         \multicolumn{4}{l}{Abortion and Crime} \\ \midrule
         $4$ 
         & $0.12$ \scriptsize $[0.11, 0.13]$
         & $0.14$ \scriptsize $[0.11, 0.15]$ 
         & $\mathbf{0.11}$ \scriptsize $[0.09, 0.12]$  \\ 
         $32$ 
         & $0.30$ \scriptsize $[0.29, 0.33]$
         & $0.17$ \scriptsize $[0.15, 0.18]$
         & $\mathbf{0.12}$ \scriptsize $[0.10, 0.14]$
         \\ 
         $256$
         & $0.33$ \scriptsize $[0.32, 0.34]$
         & $0.16$ \scriptsize $[0.15, 0.20]$
         & $\mathbf{0.11}$ \scriptsize $[0.10, 0.12]$  \\ 
         \midrule 
         \multicolumn{4}{l}{Grade Retention in Reading} \\ \midrule
         $4$ 
         & $0.05$ \scriptsize $[0.05, 0.07]$ %
         & $0.07$ {\scriptsize $[0.06, 0.07]$}
         & $\mathbf{0.03}$ {\scriptsize $[0.03, 0.03]$}
         \\ 
          $32$ 
         & $0.11$ \scriptsize $[0.10, 0.12]$ %
         & $0.07$ {\scriptsize $[0.07, 0.07]$}
         & $\mathbf{0.03}$ \scriptsize $[0.03, 0.03]$
         \\
          $256$ 
         & $0.10$ \scriptsize $[0.10, 0.11]$ %
         & $0.07$ \scriptsize $[0.07, 0.07]$
         & $\mathbf{0.03}$ \scriptsize $[0.03, 0.04]$
         \\
         \midrule
         \multicolumn{4}{l}{Grade Retention in Math} \\ \midrule
         $4$ 
         & $\mathbf{0.05}$ \scriptsize $[0.03, 0.07]$ %
         & $0.09$ \scriptsize $[0.07, 0.14]$
         & $\mathbf{0.05}$ \scriptsize $[0.03, 0.10]$
         \\ 
         $32$ 
         & $0.09$ \scriptsize $[0.07, 0.10]$ %
         & $0.10$ \scriptsize $[0.07, 0.17]$
         & $\mathbf{0.06}$ \scriptsize $[0.03, 0.12]$
         \\
         $256$ 
         & $0.08$ \scriptsize $[0.08, 0.09]$ %
         & $0.09$ \scriptsize $[0.07, 0.16]$
         & $\mathbf{0.05}$ \scriptsize $[0.03, 0.11]$
         \\
         \bottomrule
    \end{tabular}
    \vshrink{-0.25em}
    \caption{Semi-synthetic experiment: mean average error (MAE) for the average treatment effect (ATE) estimates, across all datasets, for $N=2000$. 
    We report $50\%,25\%$ and $75\%$ percentiles across $10$ independent replications. {\bf Boldface} indicates the best result(s).
    }
    \vshrink{-0.5em}
    \label{tab:proxy_experiment}
\end{table}

%% file: sec-conclusion.tex
\vshrink{-0.6em}
\section{DISCUSSION}\label{sec:conclusion}
\vshrink{-0.6em}

Ill-posedness is a fundamental issue in nonparametric CMMs. This work demonstrates that representations with well-controlled measures of ill-posedness can be {\em robustly} learned from data, 
by studying a kernelized CMM estimator instantiated with %
learned kernels. The estimator 
enjoys consistency guarantees under intuitive assumptions, and exhibits promising empirical performance. 
Our results are also interesting from a few other aspects, as we discussed. 

Our learning target \eqref{eq:ideal-rkhs} has been a ``gold standard'' in a large proportion of the nonparametric IV/CMM literature. Still, it will not suit all applications. \cref{app:disc-gtRKHS} discusses its inductive bias in detail; 
below we briefly outline its limitations, and possible directions for future work:
\begin{itemize}[leftmargin=*,topsep=2pt,itemsep=1pt]
    \item The learning target \eqref{eq:ideal-rkhs} does not account for overlap variables. While we have derived extensions in Sec.~\ref{sec:algo-impl} and evaluated an extended algorithm, it remains future work to study the high-dimensional extension as mentioned, from both empirical and theoretical perspectives.
    \item While \eqref{eq:ideal-rkhs} adapts to certain latent structures which can be particularly desirable in some scenarios (Ex.~\ref{ex:main}), it does not benefit from supervision signals from $\by$. %
    In such cases, our results, in particular, %
    Thm.~\ref{thm:bounded-ill-posedness}, could still be useful to analyze a two-phase kernel learning algorithm, where the first phase takes in the supervision signals similar to \citet{xu_learning_2020}, 
    and the second stage modifies its output using information about $E$, to control the moduli of continuity. %
    \item While assumptions like \eqref{eq:general-source-cond} are necessary for efficient estimation, and inference under identifiability, they can be undesirable in non-identifiable scenarios if one wishes to construct confidence regions. %
    These two types of scenarios have been studied separately in the literature, and our work makes no exception. 
\end{itemize}
More broadly, different causal estimation problems necessarily require different treatments, but the issue of ill-posedness is a recurring theme. And so is the need to develop representation learning algorithms that address this issue. 
Our work can be viewed as evidence suggesting the RKHS perspective could be particularly helpful in such developments. 

%% file: app-additional-discussion.tex
\newcommand{\Ereg}{\mathrm{E}_r}

\section{Additional Notations and Conventions}\label{rmk:addi-conventions}
We introduce additional notations which will be used in the appendix: $\asymp, \lesssim, \gtrsim$ denote (in)equalities up to constants. $\prec,\succ$ denote the L\"owner order between operators. $a\wedge b := \min\{a, b\}, a\vee b := \max\{a,b\}$. 
Throughout the appendix we will use $c,c',c_1,c_2,\ldots$ to refer to universal constants, and $C,C',C_1,C_2,\ldots$ to refer to constants whose value may depend on constants introduced in the assumptions (\eg, $b,p,\alpha,\beta$), but are otherwise independent of the sample size or the data generating process. 
The exact value of such constants may change from line to line.

We will occasionally abuse notations and refer to operators from a finite-dimensional Euclidean space to some $L_2$ space as ``infinitely wide matrices''; e.g., given functions $\psi_1,\ldots,\psi_m\in L_2(P_x)$, we may use the notation $\Psi = (\psi_1; \ldots; \psi_m)$ to refer to the operator $$
\Psi: L_2(P_x)\to\RR^m, h\mapsto \begin{bmatrix}
    \<\psi_1, h\>_2 \\ 
    \ldots \\
    \<\psi_m, h\>_2
\end{bmatrix}.
$$
We also note that while in the text we do not distinguish between square-integrable functions, such as $f: \cX\to \RR$, and their respective $L_2$ equivalence classes, such as $[f]_\sim \in L_2(P_x)$, as is done in past works for readability \citep[e.g.,][]{chen_rate_2007}, we will make the distinction in appendix {\em when necessary}. In such places, we will use the operator 
$$
\Ereg: L_\infty(P_x) \to L_\infty(P_z)
$$
to refer to a version of conditional expectation $\EE(f(\bx)\mid\bz=\cdot) =: \Ereg f$, defined with an arbitrary version of the regular conditional probability which is fixed throughout the work. Under basic assumptions which hold for open subsets of $\RR^d$, there exists such an $\Ereg$, which is a bounded operator; see, e.g., \citet[Remark B.1]{wang2022fast}. 

\section{Additional Discussion: Background and Related Work}\label{app:bg}

\subsection{Additional Background: Conditional Moment Models}\label{app:disc-cmm-estimators}

\paragraph{Examples} To illustrate the assumptions in various conditional moment models, we provide sample %
causal graphs in \cref{fig:cg}. We also recall the core assumptions in proximal causal inference in the following example: 
\begin{example}[proxy control, \citealp{miao_identifying_2018}]\label{ex:pc}
Let the random variables $\bv,\bw,\bt,\bu,\by$ be s.t.
\begin{enumerate}[leftmargin=*,topsep=0pt,label=(\roman*)]
\item $\by = f_s(\bt,\bu)$ for some function $f_s$;
\item $\bw\indep (\bt,\bv)\mid \bu$;
\item The conditional expectation operators $E_{ut|z}: f\mapsto \EE(f(\bu,\bt)\mid \bt,\bv), E_{ut|x}: f\mapsto \EE(f(\bu,\bt)\mid\bt,\bw)$ are injective. 
\end{enumerate}
In the above, $\bt, \by, \bu$ denote the treatment, outcome, and (unobserved) confounder. $\bv,\bw$ denote two observed proxies for $\bu$. %
Condition (ii) is implied by the causal graph \cref{fig:cg} (b); \cref{fig:cg} (c) describes a particularly interesting special case, where 
the observable information about $\bu$ can be summarized into lower-dimensional variables $\bar\bv, \bar\bw$. %
(iii) are {\em completeness assumptions} which require $(\bv,\bw)$ to be sufficiently informative; our statement is adapted from \citet{deaner_proxy_2021}. 

By \citet{miao_identifying_2018,deaner_proxy_2021}, for any $f_0$ satisfying 
$
\EE(\by - f_0(\bt,\bw)\mid \bt,\bv) = 0,
$
we have 
$
\EE(\by\mid \mrm{do}(\bt=t)) = \int P_w(dw)f_0(t, w).
$
Thus, the treatment effect can be recovered by solving the problem \eqref{eq:cmm-defn} 
with $\bx\gets(\bt,\bw), \bz\gets(\bt,\bv)$. 
\end{example}

\paragraph{Population objectives} Estimators of the form \eqref{eq:minimax-estimator} have been proposed and studied in various recent works, as we have cited in the text. Here we provide a more complete review of different estimators and their population objectives. 
Classical nonparametric methods \citep{hall_nonparametric_2005,newey2003instrumental}, as well as some recent works in the ML literature \citep{hartford2017deep,xu_learning_2020,kato2021learning} are motivated as approximately optimizing \eqref{eq:popu-obj}. In connection with \eqref{eq:minimax-estimator}, 
\citet{zhang_maximum_2020} motivated an estimator by optimizing a kernelized statistic from \citet{muandet2020kernel}. The estimator has the form of 
\begin{equation}\label{eq:mmr-iv}
\hat f_n =\arg\min_{f\in\cH} \max_{g\in\cI, \|g\|_\cI=1} \frac{1}{n}\sum_{i=1}^n (f(x_i)-y_i)g(z_i) + \lambda_n\|f\|_\cH^2.
\end{equation}
In both \eqref{eq:minimax-estimator} and \eqref{eq:mmr-iv} the (unregularized population version of the) inner loop reduces to \eqref{eq:popu-obj}, if we replace $\cI$ with $L_2(P_z)$. Assuming $\cI$ well approximates the functions $\{E(f-f_0): f\in\cH\}$, we can also use the inner loop of \eqref{eq:mmr-iv} to upper bound \eqref{eq:popu-obj}, as is done in \citet[pp.~51-52]{kallus2021causal}. 
\citet{singh_kernel_2020} studied a similar estimator to \eqref{eq:minimax-estimator} for kernel hypothesis spaces, by generalizing the classical two-stage least square estimator. As noted in \citet[App.~C.1.4]{wang2021scalable}, the estimator is similar to \eqref{eq:minimax-estimator} and \eqref{eq:mmr-iv} up to the choices of regularizers. %
\citet{kremer2022functional} constructed an estimator based on the framework of empirical likelihood; while having a different populational objective, their nonasymptotic guarantees are still established for \eqref{eq:popu-obj}. 

\paragraph{Convergence results, measures of ill-posedness} Past works on NPIV estimation often impose variants of the {\em link and reverse link conditions} \citep[Assumptions~2, 5]{chen_rate_2007} which, for mildly ill-posed inverse problems,\footnote{
Mildly ill-posed inverse problems refer to settings where the singular values of $E$ have a polynomial decay. 
We restrict to such settings, as has been done in previous works \citep[e.g.,][]{knapik_bayesian_2011,szabo2015frequentist}, because subpolynomial factors are less relevant here, where the minimax rate has a polynomial decay. 
} require the existence of $\gamma,C>0$ s.t.~
\begin{equation}\label{eq:link}
C^{-1} \|T^{\gamma} f\|_2 \le
\|E f\|_2 \le C \|T^{\gamma} f\|_2 ~~
\forall f\in \mrm{Ran}(T^{1/2}). 
\end{equation}
The space $\mrm{Ran}(T^{1/2})$ has the role of the hypothesis space. If it is chosen as in \eqref{eq:ideal-rkhs}, \eqref{eq:link} will hold with $\gamma=1/2\alpha$. The link and reverse link conditions require the eigenfunctions of $T$ and $E^\top E$ be related in a strong sense: setting $f = \psi_i$, the $i$-th eigenfunction of $T$, in the above, we find 
\begin{equation}\label{eq:link-corr}\tag{\ref{eq:link}'}
\|E \psi_i\|_2 \asymp e_i^{2\gamma}(T) \overset{(i)}{\asymp} \bar s_i = \|E \bar\psi_i\|_2.
\end{equation}
where $e_i(\cdot)$ denotes the $i$-th eigenvalue of the operator, and (i) holds for our \eqref{eq:ideal-rkhs}, and more generally any $T$ based on which order-optimal $L_2$ rates can be derived. \eqref{eq:link-corr} is not strictly equivalent to requiring $\{\psi_i\}$ to approximate $\{\bar\psi_i\}$ in the sense of \eqref{eq:svd-approx-opt}; in particular, the leading ``singular vector estimates'' can incur a $\cO(1)$ suboptimality. However, if we consider the leading eigenfunctions (say, $\{\psi_i: i<J_n\}$) to be close to optimal, we can see that \eqref{eq:link-corr} imposes very strong conditions on the higher-order $\psi_{J_n}$, and cannot be implied by \eqref{eq:svd-approx-opt} or \eqref{eq:hs-norm} which can possibly be satisfied by learned singular vectors. 

As discussed in \citet[Section 2.2]{chen_rate_2007}, our source condition \eqref{eq:general-source-cond} implies their approximability condition (Asm.~1) if we additionally require $\|f_0\|_{\gtRKHS[\beta]} \le C$ to be uniformly bounded. Their latter condition is more common in literature, although there is no fundamental difference and our consistency results can be adapted to such settings with little effort. 

\paragraph{Alternative strategies for $L_2$ consistency} In the nonparametric literature, an alternative strategy for achieving $L_2$ consistency is to directly impose regularizations in the $L_2$ space \citep{carrasco2007linear}. Most relevant are regularization strategies that do not require knowledge of the SVD of $E$, which includes
Tikhonov regularization:
$$
\hat f_n := 
\argmin_{f\in L_2(P_x)} \|\hat E_n f - \hat g_n\|_2^2 + \alpha_n \|f\|_2^2 = \underbrace{(\hat E_n^\top \hat E_n + \alpha_n I)^{-1}}_{g_{\alpha_n}(\hat E_n^\top \hat E_n)}\hat E_n \hat g_n, 
$$
and the Landweber regularization which uses a different family of $\{g_{(\cdot)}\}$ and corresponds to doing gradient descent \emph{in the $L_2(P_x)$ space} with a fixed number of iterations (and step-size) determined by $\alpha_n$. 
The Landweber regularization is difficult to apply in the combination of adaptive machine learning models, because to ensure a gradient descent-based implementation matches its definition, we need to parameterize $f\in\cH$ using its coefficients on a predetermined orthonormal system in $L_2(P_x)$.\footnote{
Otherwise, the result after a fixed number of iterations will have a different form involving terms related to the parameterization map that can no longer be eliminated.
} The use of such a fixed basis, which must be truncated for a practical implementation, will prevent the adaptation to the informative latent structures as in Ex.~\ref{ex:main}. 
Tikhonov regularization does not have this issue, but the resulting rate will saturate at $\beta=1$ \citep{carrasco2007linear}. This is in contrast to our result, which adapts to $\beta>1$ (and provides a faster, $o(n^{-1/4})$ rate in such cases). 

Closely related is the analysis of a Tikhonov-regularized estimator, where the regularizer is defined using RKHS norms. While the resulted estimator is similar in form to \eqref{eq:minimax-estimator}, the analysis relies on a very different assumption \citep[e.g.,][Assumption 16]{mastouri_proximal_2021}, which is a source condition different from our \eqref{eq:general-source-cond}: 
\begin{equation}\label{eq:strong-src-cond}
f_0 \in \Ran (S^\top S)^{\beta/2}, ~~\text{where}~~ S: \cH\to\cI, 
(Sf)(z') := \int k_z(z,z') f(x) dP_{zx},
\end{equation}
where $\cH,\cI$ are RKHSes, and $k_z$ is the reproducing kernel of $\cI$. This approach suffers from a similar saturation issue, in which their RKHS norm rate $\|\hat f_n - f_0\|_\cH$ cannot go beyond $\Omega(n^{-1/4})$. 
While such results are not directly comparable to ours due to the different norm, note that they cannot lead to better $L_2$ rates. 

Another important issue lies in hyperparameter selection. While we can usually consider find \emph{some} $\alpha_n$ that allows for consistent estimation, it is very unclear how optimal hyperparameters can be determined, especially when we do not have knowledge of $\beta$; in the simplified scenarios of $\hat E_n = E$, it can be readily shown that %
the choice of $\alpha_n$ that optimizes $\|E(\hat f_n-f_0)\|_2$ is different from the choice that optimizes $\|\hat f_n-f_0\|_2$. This is in stark contrast to our approach which, at least in simplified scenarios and with additional regularity assumptions on $f_0$, may identify order-optimal choices of $(\lambda_n,\alpha)$ using the statistic $\|E(\hat f_n-f_0)\|_2$ \citep[see e.g.,][]{szabo2015frequentist}.

\subsection{Inductive Bias of the Target Space \eqref{eq:ideal-rkhs}}\label{app:disc-gtRKHS}

In this subsection, we discuss the inductive bias of \eqref{eq:ideal-rkhs} and its implications. We note that the claims made in \cref{ex:main} and \cref{ex:main-torus} are proved as \eqref{eq:latent-fs}, \eqref{eq:product-E}, \eqref{eq:jjjjj}, \eqref{eq:top-level-space} and Claim~\ref{claim:pc}; see also the discussion around them.

\paragraph{Compositional structure} In compositional models as described by \cref{ex:main}, the space $\gtRKHS[\alpha]$ only depends on $\bx$ through the latent variables $\bar\bx$, for any $\alpha>0$. This can be proved following the argument in \citet[Section 3]{wang2022fast}:\footnote{
Note we do not distinguish between functions and their $L_2$ equivalence classes here, as noted in \cref{rmk:addi-conventions}. More rigorous claims can be established by applying the results in \citet{steinwart_mercers_2012}, just as is done in \citet[Appendix B]{wang2022fast}. We omit the discussion which is not particularly interesting. 
} since we have the eigendecomposition
$
(E^\top E)^{\alpha} = \sum_{i=1}^\alpha\bar s_i^{2\alpha}\gtsv{i}\gtsv{i}^\top, 
$
the space is equivalently defined as 
$$
\gtRKHS[\alpha] = \Big\{f = \sum_{i\in I} s_i^{\alpha} a_i \gtsv{i}: (a_i)\in\ell_2(\mb{N})
\Big\},
$$
where $I := \{i\in\mb{N}: \bar s_i\ne 0\}$. But all $\gtsv{i}$ (with $i\in I$) must depend on $x$ only through $\bar x = f_{enc,x}(x)$: this is because we have 
$$
\gtsv{i}(\bx) = \bar s_i^{-1} (E^\top \bar\varphi_i)(\bx) \overset{(i)}{=} \bar s_i^{-1} \EE(\bar\varphi_i(\bz)\mid\bx) \overset{(ii)}{=}
\bar s_i^{-1} \EE(\bar\varphi_i(\bz)\mid f_{enc,x}(\bx)) =: (\gtsv{i}^l \circ f_{enc,x})(\bx).
$$
In the above, (i) follows by Claim~\ref{claim:ET} below, (ii) by the assumed data generating process, and $\bar\varphi_i$ denotes the respective left singular vector of $E$; And we can verify $\|\gtsv{i}^l\|_{L_2(P_{\bar x})}=1$. 
Following this, %
any $f\in\gtRKHS[\alpha]$ will satisfy %
\begin{equation}\label{eq:latent-fs}
f = \bar f\circ f_{enc,x}, ~~\text{where}~~ 
\bar f \in \bar\cH_{\alpha}^{l} := \Big\{\bar f = \sum_{i\in I} s_i^{\alpha} a_i \gtsv{i}^l: (a_i)\in\ell_2(\mb{N})\Big\}  \subset L_2(P_{\bar x}).
\end{equation}

In the setting of \cref{ex:main}, for product functions of the form $f'(\bar\bx)f''(\bx_\perp)$, {\em s.t.~$\EE f''=0$},\footnote{
The condition $\EE f''=0$ was unfortunately omitted in the main text, although it does not affect any of the discussion below, as $f''$ with non-zero mean will be equivalent to a constant function. 
} we have
\begin{equation}\label{eq:product-E}
\EE(f'(\bar\bx)f''(\bx_\perp)\mid\bz) = \EE(f'(\bar\bx)\mid\bz)\EE(f''(\bx_\perp)\mid\bz) = \EE(f'(\bar\bx)\mid\bz)\EE(f''(\bx_\perp)).
\end{equation}
To understand the implication of the above observation,  
let us take some $\bx_\perp' = f_{\perp}'(\bx)$ s.t.~$\bx_\perp'\indep \bar\bx$, and there exists some function s.t.~$\bx = f_{dec,x}'(\bar\bx,\bx_\perp')$. (Note that such $\bx_\perp'$ always exist \citep[Prop~5.13]{kallenberg1997foundations}, %
 and \eqref{eq:product-E} holds with $\bx_\perp$ replaced by $\bx_\perp'$.) Then we have $\bx_\perp'\indep \bz$, and there exists a bijection between $\bx$ and $(\bx_\perp',\bar\bx)$. Therefore, for {\em any} $$
\{h_\perp^{(i)}: i\in \mb{N}\} \subset L_2(P_{\bx_\perp'}), 
\{h_\parallel^{(i)}: i\in \mb{N}\} \subset L_2(P_{\bar\bx})
$$
that constitute ONBs for their respective $L_2$ space, the product functions
$$
\{h^{(i,j)} := h_\perp^{(i)}(f_{\perp}'(\cdot))h_\parallel^{(j)}(f_{enc,x}(\cdot)), ~i,j\in\mb{N}\}
$$
form a complete basis for $L_2(P_x)$, which is orthonormal by independence.%
And any $f\in L_2(P_x)$ can be written as 
$$
f = \sum_{i,j\in\mb{N}} a_{ij} h^{(i,j)}, ~~\text{where}~~(a_{ij})\in\ell_2(\mb{N}\times\mb{N}). 
$$  
As the above holds for arbitrary choices of basis, let us set $
h_\perp^{(1)} = \mbf{1}, h_\parallel^{(1)} = \mbf{1}
$ to be constant functions. Then we have 
\begin{align*}
\EE(h^{(i)}_\perp(\bx_\perp')) &= \<h^{(i)}_\perp, \mbf{1}\>_2 = 0, ~\forall i>1, \numberthis\label{eq:iiiii}\\ 
Ef &= E\Big(\sum_{i,j\in\mb{N}} a_{ij}h^{(i,j)}\Big) \overset{(i)}{=} \sum_{i,j\in\mb{N}} a_{ij} E h^{(i,j)} \overset{\eqref{eq:product-E}}{=} \sum_{i,j\in\mb{N}} a_{ij} 
\EE(h^{(i)}_\perp(\bx_\perp'))\EE(h^{(j)}_\parallel(\bar\bx)\mid\bz) \\
&\overset{\eqref{eq:iiiii}}{=} \sum_{j\in\mb{N}} a_{1j} \EE(h^{(j)}_\parallel(\bar\bx)\mid\bz) \overset{(ii)}{=}  \EE\Big(\sum_{j\in\mb{N}} a_{1j} h^{(j)}_\parallel(\bar\bx)\,\Big\vert\, \bz\Big),
\numberthis\label{eq:jjjjj}
\end{align*}
where (i), (ii) both follow by continuity of the operator $E$. By \eqref{eq:jjjjj}, %
only functions of the form 
$
f = \bar f\circ f_{enc,x}
$
can be identified from the data. %

\paragraph{Regularity} We have established that $f_0\in\gtRKHS$ satisfies $f_0 = \bar f_0\circ f_{enc,x}$ for some $\bar f_0\in\gtRKHS^l$. For $f_0$ to be efficiently estimable, we still need $\bar f_0$, or the space $\gtRKHS^l$, to have good regularity.\footnote{Regularity assumptions on $f_{enc,x}$ will be introduced in \cref{thm:cmm}.} To establish this we just need to relate the functions $\gtsv{j}^l$ in \eqref{eq:latent-fs} to the basis of well-understood RKHSes. This is easy: as we establish in Claim~\ref{claim:l-sv}, $\gtsv{j}^l$ are simply the right singular vectors of a latent-space conditional expectation operator $\bar E$, and thus we have 
\begin{equation}\label{eq:top-level-space}
\gtRKHS[\alpha]^l = \Ran((\bar E^\top \bar E)^{\alpha/2}),
\end{equation}
and the inner product is defined using $(\bar E^\top \bar E)^{\alpha/2}$, similar to \eqref{eq:ideal-rkhs}. Regularity of such spaces is well-understood in literature; in particular, for our \cref{ex:main-torus} we know by \citet[Section 6]{chen_rate_2007} that $\gtRKHS[\alpha]^l$ is the periodic Sobolev space, which proves the claim made in the example.

\paragraph{Benefits, extensions, limitations} From the perspective of estimation theory, our choice to approximate $\gtRKHS[\alpha]$ can be motivated from the imposed source condition \eqref{eq:general-source-cond}: as discussed in \cref{app:disc-cmm-estimators}, similar assumptions have been essential for efficient estimation; thus, $\gtRKHS[\alpha]$ can be viewed as a natural choice to best utilize prior knowledge of the form \eqref{eq:general-source-cond}. 
From a modeling perspective, the adaptivity of $\gtRKHS[\alpha]$ to the aforementioned latent structure can also be appealing for proxy control problems, where $\bz$ and $\bx$ may contain high-dimensional components corresponding to the two measurements of the confounder, the information about which will be fully preserved in $(\bar\bz, \bar\bx)$ (Claim~\ref{claim:pc}).\footnote{While in the implementation of proximal causal inference, we define the approximation target using $\bx' = \bt$ and $\bz=(\bt,\bv)$, it is easy to establish similar results for that choice.
} However, there are also limitations. We have discussed the overlap variable issue in the text; here we discuss the following issues in more detail: 
\begin{enumerate}[leftmargin=*,label=(\roman*)]
    \item Direct use of $\gtRKHS[\alpha]$ will not benefit from information about $\by$, which may provide additional adaptivity. This is most relevant in high-dimensional IV regression \citep{fan_endogeneity_2014,lin_regularization_2015} where we want to do the variable selection for a large number of treatments, and the available instruments are informative for a large proportion of them. In such cases, $\gtRKHS[\alpha]$ will be information about all such treatments, and it will be desirable to further reduce its complexity based on information provided by $\by$. \\ 
    In the text we have outlined a procedure that swaps the order of these two steps: we start with a learned hypothesis space based on $\by$, and ``prune'' it to control the modulus of continuity; this ordering can be preferable as it avoids estimating the less relevant parts of $\gtRKHS[\alpha]$, and its  
    latter step may be justified following the reasoning in \cref{thm:bounded-ill-posedness} (which should apply to subspaces of $L_2(P_x)$). Here we note a possible alternative route, which is to conduct variable selection on $\bz$ first, based on the correlation structure with $\by$, and use the filtered instruments to define $\bar\bx$ and $\gtRKHS[\alpha]$. 
    \item The assumption \eqref{eq:general-source-cond} for inference or predictive uncertainty quantification. %
    We note that similar assumptions have been employed for these purposes: the assumptions %
    in, \eg, \citet{kato_quasi-bayesian_2013,chen_optimal_2015} are closely related, as discussed in \citet[p.~502]{chen_rate_2007}. Intuitively, \eqref{eq:general-source-cond} is more adequate when $\bz$ is believed to be informative. However, when $\bz$ is markedly non-informative and cannot explain important aspects of $\bx$, \eqref{eq:general-source-cond} could be violated, and the use of $\gtRKHS[\alpha]$ may lead to erroneously small confidence regions. At a high level, this is not dissimilar to the weak instrument problem in linear IV \citep{staiger_instrumental_1997}, which required a separate treatment. For nonparametric models, there are relatively fewer discussions on similar (nearly) nonidentified settings, and it is especially unclear what we should expect for general CMM problems in such settings when good hypothesis spaces are not known a priori.
\end{enumerate}

\paragraph{Deferred claims and proofs}
\begin{claim}\label{claim:ET}
We have $E^\top g = \EE(g(\bz)\mid\bx=\cdot)$ for all $g\in L_2(P_z)$. 
\end{claim}
\begin{proof}
We have, for all $h\in L_2(P_x)$,  
$$
\EE(h(\bx)(E^\top g)(\bx)) = 
\<h, E^\top g\>_2 = \<E h, g\>_2 = \EE(\EE(h(\bx)\mid\bz)g(\bz)) = \EE(h(\bx)g(\bz)),
$$
which fulfills the definition of conditional expectation \citep{kallenberg1997foundations}.
\end{proof}

\begin{claim}\label{claim:l-sv}
Let $\gtsv{j}^l$ be defined in \eqref{eq:latent-fs}, $\bar\bx,\bar\bz$ be defined in \cref{ex:main}. 
When $\{\bar s_j\}$ are strictly decreasing, $\gtsv{j}^l$ must be right singular vectors of the operator 
$$
\bar E: L_2(P_{\bar x})\to L_2(P_{\bar z}), \bar f\mapsto \EE(\bar f(\bar\bx)\mid\bar\bz=\cdot).
$$
\end{claim}
\begin{proof}
Let $\{\bar\psi'_j\}$ be the singular vectors of $\bar E$, so that they form an ONB for $L_2(P_{\bar x})$. It is thus clear that $\{\bar\psi'_j\circ f_{enc,x}\}$ form an ONS in $L_2(P_x)$. 
We claim that for all $J\in\mb{N}$, $\{\psi_j \gets \bar\psi'_j\circ f_{enc,x}\}$ maximizes the variational objective 
$
\cL(\{\psi_j\}_{j=1}^J) := \sum_{j=1}^J \|E \psi_j\|_2^2,
$
where $\{\psi_j\}$ ranges over all orthonormal systems. As the singular values are strictly decreasing, the optima of the above objective must define the same span as the true singular vectors, and the proof will complete by the arbitrariness of $J$.
But the optimality is clear given \eqref{eq:jjjjj},\footnote{and the fact that 
$\EE(f(\bx)\mid\bz)=\EE(f(\bx)\mid\bar\bz) a.s.$}
in which we can choose $\{\bar\psi'_j\}$ as 
$\{h^{(i)}_\parallel\}$. 
\end{proof}

\begin{claim}\label{claim:pc}
Let $\bz=(\bt,\bv), \bx=(\bt,\bw)$ be as in \cref{ex:pc}. Then we have $
\EE(f(\bu)\mid\bx)=\EE(f(\bu)\mid\bar\bx), 
\EE(f(\bu)\mid\bz)=\EE(f(\bu)\mid\bar\bz). 
$
\end{claim}
\begin{proof}
Any bounded linear operator is injective if and only if its adjoint has a dense image \citep[see e.g.][p.~500]{steinwart2008support}.
Therefore, the completeness assumption in Ex.~\ref{ex:pc} implies both $\Ran E_{x|ut} = \Ran E_{ut|x}^\top$ and $\Ran E_{z|ut} = \Ran E_{ut|z}^\top$ are dense in $L_2(P_{ut})$.\footnote{
The equalities follow by the same reasoning as Claim~\ref{claim:ET}; $E_{x|ut}$ refers to the conditional expectation operator from $L_2(P_x)$ to $L_2(P_{ut})$, and $E_{z|ut}$ is defined similarly.  
} Now, 
for any $f\in L_2(P_u)\subset L_2(P_{ut})$ in the second equality, we can take $h_n\in L_2(P_x)$ s.t.~$\|E_{x|ut} h_n - f\|_{L_2(P_{ut})}\to 0$. And we have 
\begin{align*}
E_{x|z} h_n &= \EE(h_n(\bt,\bw)\mid \bt,\bv) \overset{(i)}{=} \EE(\EE(h_n(\bt,\bw)\mid \bt,\cancel{\bv,}\bu)\mid \bt,\bv) =  %
E_{ut|z}(E_{x|ut} h_n). \\ 
\|E_{x|z} h_n - E_{u|z} f\|_2 &= 
\|E_{ut|z} (E_{x|ut} h_n) - E_{ut|z} f\|_2 \le \|E_{ut|z}\| \|E_{x|ut} h_n-f\|_2 \to 0.
\end{align*}
In the above, (i) follows by the assumed  conditional independence in Ex.~\ref{ex:pc}. Since we also have $E_{x|z} h_n = E_{x|\bar z} h_n$ by the condition in Ex.~\ref{ex:main}, %
and
$$
\|E_{x|\bar z} h_n - E_{x|\bar z} h_m\|_2 \le 
\|E_{x|z} h_n - E_{x|z} h_m\|_2 \le 
\|E_{ut|z}\| \|E_{x|ut} h_n - E_{x|ut} h_m\|_2\to 0, \quad(\text{as }\min\{n,m\}\to\infty)
$$
we conclude there exists some $g\in L_2(P_{\bar z})$, which we can take as the limit of $E_{x|\bar z}h_n\in L_2(P_{\bar z})$, s.t.~
$$
\EE(g(\bar\bz) - \EE(f(\bu)\mid\bz))^2 = 0.
$$
This proves the second equality in the claim. The first equality can be proved similarly.
\end{proof}

\subsection{Related Work: Representation Learning}\label{app:disc-repr-learning-lit}

\paragraph{Multi-view redundancy} 
It has been established in various settings that, given two views $(\bx,\bz)$ of a latent variable $\bv$, self-supervised learning procedures can learn representations that are suitable for predicting functions of $\bv$, or more generally, random variables that can be well-estimated using $\bx$ and $\bz$ \citep{foster_multi-view_2008,arora_theoretical_2019,lee_predicting_2020,tosh_contrastive_2021}. 
While the goal appears different from our work, it is connected through the observation in \citet{tosh_contrastive_2021}, that accurate estimation of the density ratio $p(x,z)/p(x)p(z)$ leads to an {\em infinite-dimensional} representation that enjoys a small approximation error. As remarked in their p.~13, 
\citet{tosh_contrastive_2021} did not make an attempt to construct compact finite-dimensional approximations, and in the general case only provided a $n$-dimensional representation, the estimability of which is also unclear; %
this is different from our $\cO(n^{1/2p})$-dimensional approximation (\cref{app:additional-results}), from which we were able to establish complete learning rates. 

As noted in the text, our results for representation learning are closest to \citet{haochen_provable_2021}, as we started with the same algorithm; we derived different results which are useful in different contexts. Other works in this literature also tend to have different assumptions, due to their different goal: \citet{arora_theoretical_2019,balestriero_contrastive_2022} considered homogenous sample space ($\cZ=\cX$) with discrete latents, which lead to a different eigenspectrum and is more suitable for classification problems; \citet{lee_predicting_2020} assumed %
it is easy to {\em reconstruct} one of the views using the other, which is suitable when its dimensionality is low, or there is a large number of unlabeled samples. %

We note that similar ideas have also appeared in recent works in reinforcement learning \citep{zhang2022making,qiu2022contrastive}, which studied contrastive estimation of a Bellman operator. 
However, existing works in this direction only use the Bellman operator to solve a {\em forward problem}, which is significantly less challenging than our inverse problem setting. (On the flip side, they require convergence in stronger norms.) Similar inverse problems arise in relation to confounded MDPs \citep{kallus2021causal,fu2022offline}; it remains future work to investigate kernel learning in such settings.

\paragraph{SVD, CCA and HSIC} 
\citet{michaeli_nonparametric_2016} formulated the SVD problem as a nonparametric analogy of canonical correlation analysis (CCA), thereby connecting to the parametric analysis in \citet{foster_multi-view_2008}. Here we connect this problem to a maximization of the Hilbert-Schmidt Independence Criterion (HSIC, \citealp{gretton2005kernel}), thereby connecting to the kernel learning algorithm \citet{li_self-supervised_2021}, which was designed for self-supervised learning:

Consider a compositional model similar to \cref{ex:main-hypo}, and suppose we know the structure of the latent-space distribution $P_{\bar\bx\bar\bz}$, and thus that of the ``latent-space RKHS''; that is, we know the function space $\bar\cH_{latent, \alpha}$ 
$$
\gtRKHS[\alpha] := \{\bar f\circ f_{enc,x}, \bar f\in \bar\cH_{latent, \alpha}\},
$$
as well as its (assumed) RKHS structure. Then we know the reproducing kernel of $\gtRKHS[\alpha]$ must have the form of 
$$
\bar k_{\alpha,x}(x,x') = \bar k_{\alpha, \mred{\bar x}}(f_{enc,x}(x), f_{enc,x}(x')), 
$$
where $\bar k_{latent, \alpha}$ {\em is known};\footnote{e.g., in \cref{ex:main-torus} we may know that $\bar k_{latent,\alpha}$ is a Mat\'ern kernel on torus with an known order.}
and it remains to learn $f_{enc,x}$ from data. Similarly, for $\bz$ we know that we need to learn the function $f_{enc,z}$ in 
$$
\bar k_{\alpha,z}(z,z') = \bar k_{\alpha, \mred{\bar z}}(f_{enc,z}(z), f_{enc,z}(z')). 
$$
Now, suppose in addition that the kernel $\bar k_{latent, \alpha}$ {\em is stationary}, and the marginal distributions $P_{\bar x}, P_{\bar z}$ are uniform, so that the top Mercer eigenfunction is the constant function $\mbf{1}$, and the kernel embedding, which can be identified using the scaled eigenfunctions as
$$
\begin{aligned}
\phi_x &= (
    \sqrt{\bar\lambda_{\alpha,x}^{(1)}}\bar \psi^{(1)}_x(x),
    \ldots,
    \sqrt{\bar\lambda_{\alpha,x}^{(m)}}\bar \psi^{(m)}_x(x),
    \ldots
) \\ &= (
    \sqrt{\bar\lambda_{\alpha,x}^{(1)}}\bar \psi_{\bar x}^{(1)}(f_{enc,x}(x)),
    \ldots,
    \sqrt{\bar\lambda_{\alpha,x}^{(m)}}\bar \psi_{\bar x}^{(m)}(f_{enc,x}(x)),
    \ldots
)
\in \ell_2(\mb{N}),
\end{aligned}
$$
satisfies 
$$
\EE_{\bx\sim P_{x}} \phi_\bx = \EE_{\bar\bx\sim P_{\bar x}} (
    \sqrt{\bar\lambda_{\alpha,x}^{(1)}}\bar\psi_{\bar x}^{(1)}(\bar\bx), \ldots
)
=  (1, 0, \ldots, 0,\ldots).
$$
where the last equality follows from the fact that
$$
\bar\psi_{\bar x}^{(1)} =\mbf{1}, ~~
\EE\bar\psi_{\bar x}^{(i)} =
\<\bar\psi_{\bar x}^{(i)},\bar\psi_{\bar x}^{(1)}\>_2
 = 0~\forall i>1.
$$
Thus, the HSIC using the product kernel reduces to 
$$
\mrm{HSIC}_{\bar k_{\alpha,x}\otimes \bar k_{\alpha,z}} = \|\EE(\phi_x\otimes \phi_z) - (\EE\phi_x\otimes\EE\phi_z)\|_{\mrm{HS}}^2 \overset{(i)}{=} 
\|\EE(\phi_x\otimes \phi_z)\|_{\mrm{HS}}^2 - 1 \overset{(ii)}{=}
\mrm{Tr}(\Lambda_z \bar\Psi_z^\top E \bar\Psi_x \Lambda_x \bar\Psi_x^\top E^\top \bar\Psi_z),
$$
In the above, 
(i) follows from the expression of kernel mean embedding derived above (and a similar result for $z$), and (ii) follows by basic linear algebra; and  
$$
\bar\Psi_x: \ell_2(\mb{N})\to L_2(P_x), (a_i)\mapsto \sum_{i=1}^\infty a_i \bar\psi_x(\cdot)
$$ can be viewed as an ``infinitely tall matrix'' obtained by ``stacking'' all eigenfunctions, and $\bar\Psi_z$ is defined similarly; $\Lambda_x: \ell_2(\mb{N}) \to \ell_2(\mb{N})$ is an ``infinitely large diagonal matrix'' consisting of Mercer eigenvalues. 
With some effort\footnote{
While we do not detail the proof here, it consists of plugging in the optimal choice of $U$ for \eqref{eq:hsic-var-form}, which must be eigenfunctions for the operator $E^\top V \Lambda_z V^\top E$; and applying a Mathematics StackExchange answer \href{https://math.stackexchange.com/questions/4321102/maximizing-weighted-sum-of-eigenvalues-of-a-matrix-lambda-1-u-top-lambda-2/}{linked here}. 
}
we can prove that, when the Mercer eigenvalues are fixed to {\em any decreasing sequences}, $(\bar\Psi_z,\bar\Psi_x)$ constitutes the unique maximizer for the objective 
\begin{equation}\label{eq:hsic-var-form}
(V,U)\mapsto 
\mrm{Tr}(\Lambda_z V^\top E U \Lambda_x U^\top E^\top V),
\end{equation}
where $(V,U)$ ranges over all unitary transforms.

While the above derivations provided nice intuitions, and connect to the practical self-supervised learning algorithm of \citet{li_self-supervised_2021}, it is unclear if similar algorithms can be applied to solve the SVD: in practice, we find it impossible to constrain the Mercer eigenvalues of the kernels, which not only depends on the assumed form of the latent-space kernels $(\bar k_{\alpha, \bar x}, \bar k_{\alpha, \bar z})$, but also on the {\em marginal distribution of the learned latents}. The latter can be altered in the learning process, which leads to altered values of $(\Lambda_x,\Lambda_z)$ and subsequently that of the variational objective. 

%% file: app-proof-sec3.tex
\section{Deferred Proofs: \cref{sec:estimability}}

The following facts will be used throughout the proofs: 
\begin{equation}\label{eq:basic-facts}
\|E\| = 1,~~ \|E f\|_2 = \<f, E^\top E f\>_2^{1/2} =: \<f, T f\>_2^{1/2} = \|T^{1/2} f\|_2.
\end{equation}
We also recall the conventions in \cref{rmk:addi-conventions}. 

\subsection{Proof for \cref{lem:approx-hs}}\label{app:proof-lem-approx-hs}

We prove the following generalization of the lemma. Note that its claim~(i) generalizes \cref{lem:approx-hs} because condition is implied by \eqref{eq:hs-norm}: the latter implies the weaker operator norm bound $\|E - \tilde E\|\le \epsilon_n$, which implies \eqref{eq:T-cond-relaxed} with $c=1$.

\begin{lemma}\label{lem:approx-HS-full}
Let $\beta\ge 1$, $\bar f$ satisfy \eqref{eq:general-source-cond}, and $\cH = \Ran \tilde T^{\alpha/2}$ as in \eqref{eq:H-defn}. Then 
\begin{enumerate}[leftmargin=*,label=(\roman*)]
    \item\label{it:lem-first-cond} If $\alpha=1$, and $\tilde T$ satisfies 
\begin{equation}\label{eq:T-cond-relaxed}\tag{\ref{eq:hs-norm}'}
c^{-1}(\|E f\|_2 - \epsilon_n)
\le 
\|\tilde T^{1/2} f\|_2 \le c(\|E f\|_2 + \epsilon_n), ~~~\forall \|f\|_2=1
\end{equation}
for some $c\ge 1,\epsilon_n>0$, there will exist $\tilde f\in\cH$ s.t.
$$
\|\tilde f\|_\cH \le c_{\alpha} \|f_0\|_{\gtRKHS[\beta]}, ~~
\|\tilde f - \bar f\|_2 \le c_{\alpha} \epsilon_n^{\alpha\wedge 1} \|\bar f\|_{\gtRKHS[\beta]}, ~~
\|E(\tilde f - \bar f)\|_2 \le c_{\alpha}\epsilon_n^{(\alpha+1)\wedge 2}\|\bar f\|_{\gtRKHS[\beta]}.
$$
\item\label{it:lem-generic-cond} For any $\alpha>0$, there will exist $\tilde f\in\cH$ s.t.
$
\|\tilde f\|_\cH \le c_{\alpha} \|f_0\|_{\gtRKHS[\beta]}, 
\|\tilde f - \bar f\|_2 \le \|T^{\alpha/2} - \tilde T^{\alpha/2}\| \|\bar f\|_{\gtRKHS[\beta]}.
$
\item\label{it:lem-second-cond} In particular, if $\tilde T$ satisfies \eqref{eq:hs-norm},  there will exist $\tilde f\in\cH$ s.t.
$$
\|\tilde f\|_\cH \le c_{\alpha} \|f_0\|_{\gtRKHS[\beta]}, ~~
\|\tilde f - \bar f\|_2 \le c_{\alpha} \epsilon_n^{\alpha\wedge 1} \|\bar f\|_{\gtRKHS[\beta]}.
$$
\item\label{it:lem-symm-result} In any case of (i)-(iii) above, we have, for all $f\in\cH$, the existence of $f'\in\gtRKHS[\alpha]$ s.t.~
\begin{equation}\label{eq:lem-approx-hs-symm-claim}\tag{\ref{eq:lem-approx-hs-claim}'}
\|f'\|_{\gtRKHS[\alpha]} \le c_\alpha \|f\|_\cH, ~~
\|f' - f\|_2 \le c_\alpha \epsilon_n^{\alpha\wedge 1} \|f\|_\cH.
\end{equation}
\end{enumerate}
\end{lemma}

\begin{proof}
In all cases, it suffices to prove the claim with $\beta$ replaced by $\alpha$, since \eqref{eq:general-source-cond} holds with smaller exponents, and we have %
\begin{equation}\label{eq:lem-gtrkhs-norm-bound}
\|\bar f\|_{\gtRKHS[\alpha]} = 
\|T^{-\alpha/2} \bar f\|_2 \le \|T^{(\beta-\alpha)/2}\| 
\|T^{-\beta/2} \bar f\|_2 = \|E\|^{\beta-\alpha} \|\bar f\|_{\gtRKHS[\beta]} 
= \|\bar f\|_{\gtRKHS[\beta]},
\end{equation}
where $T := E^\top E$. 
We also assume without loss of generality that $\epsilon_n < 1$. Now we prove the three cases in turn.

\paragraph{Proof for claim \ref{it:lem-first-cond}}
By \eqref{eq:general-source-cond}, 
there exists $g_0\in L_2(P(dx))$ s.t.~$\bar f = T^{\alpha/2} g_0$. 
Let $K$ be the largest integer s.t.~$\bar s_K > 2\epsilon_n$, and define 
$$
P_K := \sum_{i=1}^K \gtsv{i}\otimes \gtsv{i}, ~
g_{\parallel} := P_K g_0, %
~f_{\parallel} := T^{\alpha/2} g_{\parallel}. 
$$
Then
\begin{align}
\|f_{\parallel}\|_{\gtRKHS[\alpha]} &= \|g_{\parallel}\|_2 \le \|g\|_2 = \|\bar f\|_{\gtRKHS[\alpha]}, \label{eq:approx-1}\\ 
\|f_{\parallel} - \bar f\|_2 &= \|T^{\alpha/2} (g_{\parallel} - g_0)\|_2 %
\le \bar s_{K+1}^\alpha \|g_0\|_2 
< (2\epsilon_n)^\alpha \|\bar f\|_{\gtRKHS[\alpha]},  \label{eq:approx-2} \\ 
\|E(f_{\parallel} - \bar f)\|_2 &= \|T^{(\alpha+1)/2} (g_\parallel - g_0)\|_2 < (2\epsilon_n)^{\alpha+1} \|\bar f\|_{\gtRKHS[\alpha]},  \label{eq:approx-3}
\end{align}
showing that it suffices to construct an approximation for $f_{\parallel}$. For this purpose, first observe that for all $g\in \Ran P_K$ with $\|g\|_2=1$, we have $
\|E g\|_2 = \|T^{1/2} g\|_2 > 2\epsilon_n,
$ and thus \eqref{eq:T-cond-relaxed} implies 
$
\frac{1}{4c^2} g^\top T g < g^\top\tilde T g < \frac{9c^2}{4} g^\top T g,
$ 
and 
\begin{align*}
\frac{1}{4c^2} P_K T P_K \prec P_K \tilde T P_K &\prec \frac{9c^2}{4} P_K T P_K. 
\end{align*}
Thus, %
there exists $c_\alpha'>0$ s.t.\footnote{We have $c_\alpha' \le \left(2c\right)^\alpha$;
when \eqref{eq:hs-norm} holds, $c_\alpha'\le 2^\alpha$ only depends on $\alpha$.} 
\begin{align}\label{eq:approx-norm-proj-space}
\|\tilde T^{\frac{\alpha}{2}} P_K (P_K \tilde T^{\alpha} P_K)^{-1} f_\parallel\|_2 = \|(P_K \tilde T^\alpha P_K)^{-1/2} f_\parallel\|_2 
< c_\alpha' \|(P_K T^{\alpha} P_K)^{-1/2} f_\parallel\|_2 
= 
c_\alpha'
\|f\|_{\gtRKHS[\alpha]}. 
\end{align}
Consider 
$
\tilde f := \tilde T^{\alpha} P_K (P_K \tilde T^{\alpha} P_K)^{-1} f_{\parallel}. 
$
By \eqref{eq:approx-norm-proj-space}, $\|\tilde f\|_\cH = \|g'\|_2$ fulfills the claimed bound. 
And we have 
\begin{align*}
\|\tilde f - \bar f\|_2 &\le \|\tilde f - f_{\parallel}\|_2 + \|f_{\parallel} - \bar f\|_2  \\ 
&= \|(I-P_K) \tilde T^{\alpha} P_K (P_K \tilde T^\alpha P_K)^{-1} f_{\parallel}\|_2 +   \|f_{\parallel} - \bar f\|_2  \\ 
&\le \|(I-P_K) \tilde T^{\alpha/2}\| \|\tilde T^{\alpha/2} P_K (P_K \tilde T P_K)^{-\alpha} f_{\parallel}\|_2 +   \|f_{\parallel} - \bar f\|_2  \\ 
&\le \|(I-P_K) \tilde T^{\alpha/2}\| \cdot c_\alpha' \|f\|_{\gtRKHS[\alpha]} + (2\epsilon_n)^\alpha \|f\|_{\gtRKHS[\alpha]}, &&\text{\color{gray}(by \eqref{eq:approx-2}, \eqref{eq:approx-norm-proj-space})} \\ 
\|E(\tilde f - \bar f)\|_2 &\le \|E(\tilde f - f_{\parallel})\|_2 + \|E(f_{\parallel} - \bar f)\|_2  \\ 
&\le c_\alpha' \|E(I-P_K)\tilde T^{\alpha/2}\|\|f\|_{\gtRKHS[\alpha]} + (2\epsilon_n)^{\alpha+1} \|f\|_{\gtRKHS[\alpha]}&&\text{\color{gray}(by \eqref{eq:approx-3}, \eqref{eq:approx-norm-proj-space})} \\
&=c_\alpha' \|E(I-P_K)\cdot (I-P_K)\tilde T^{\alpha/2}\|\|f\|_{\gtRKHS[\alpha]} + (2\epsilon_n)^{\alpha+1} \|f\|_{\gtRKHS[\alpha]} \\ 
&\le c_\alpha' s_{K+1} \|(I-P_K)\tilde T^{\alpha/2}\|\|f\|_{\gtRKHS[\alpha]} + (2\epsilon_n)^{\alpha+1} \|f\|_{\gtRKHS[\alpha]}.
\end{align*}
Therefore, it remains to establish that 
\begin{equation}\label{eq:approx-goal}
\|(I-P_K)\tilde T^{\alpha/2}\|  \le c_\alpha'' \epsilon_n^{\min\{\alpha, 1\}},
\end{equation}
for some constant $c_\alpha''>0$. Observe that 
\begin{align*}
\|(I-P_K)\tilde T^{1/2}\| &= \|\tilde T^{1/2}(I-P_K)\| =
\sup_{\|g\|_2=1}    \|\tilde T^{1/2} (I-P_K) g\|_2 %
\overset{\eqref{eq:T-cond-relaxed}}{\le} \sup_{\|g\|_2=1} c(\|E (I-P_K) g\|_2 + \epsilon_n)  \\ &
= c(\bar s_{K+1} + \epsilon_n) < 3c\epsilon_n, %
\end{align*}
which proves 
\eqref{eq:approx-goal}, and thus the claim (i).

\paragraph{Proof for Claim \ref{it:lem-generic-cond}}
 Consider $\tilde f := \tilde T^{\alpha/2} T^{-\alpha/2} \bar f$. Then we have 
\begin{align}
\|\tilde f\|_{\cH} &= \|T^{-\alpha/2}\bar f\|_2 \overset{\eqref{eq:lem-gtrkhs-norm-bound}}{\le} \|\bar f\|_{\gtRKHS[\beta]}, \label{eq:lem-rkhs-norm-g}\\
\|\tilde f - \bar f\|_2 &\le \|\tilde T^{\alpha/2} - T^{\alpha/2}\| \|T^{-\alpha/2}\bar f\|_2 \le  \|\tilde T^{\alpha/2} - T^{\alpha/2}\|\|\bar f\|_{\gtRKHS[\beta]}. 
\label{eq:lem-L2-norm-g}
\end{align}

\paragraph{Proof for Claim \ref{it:lem-second-cond}}
It remains to bound $\|\tilde T^{\alpha/2} - T^{\alpha/2}\|$ using \eqref{eq:hs-norm}. 
First note that \eqref{eq:hs-norm} implies  
\begin{equation}\label{eq:op-norm-bound}
\| T^{1/2} - \tilde T^{1/2} \| \le \|(E^\top E)^{1/2} - (\tilde E^\top \tilde E)^{1/2}\|_{\mrm{HS}} \overset{(i)}{\le} \sqrt{2}\|E - \tilde E\|_{\mrm{HS}} 
\le \sqrt{2}\epsilon_n,
\end{equation}
where (i) can be found as \citet[Eq.~X.24]{bhatia2013matrix}. 
The result for $\alpha=1$ thus follows by \eqref{eq:op-norm-bound}. 
When $\alpha \in (0,1)$, we have 
$$
\|\tilde T^{\alpha/2} - T^{\alpha/2}\| \overset{(i)}{\le} \||\tilde T^{1/2} - T^{1/2}|\|^{\alpha} 
=\|\tilde T^{1/2} - T^{1/2}\|^{\alpha} 
 \le 2^{\alpha/2}\epsilon_n^\alpha,
$$
where (i) is \citet{birman1975estimates}, 
and $|\cdot|$ denotes the absolute value map. %
For $\alpha > 1$, first note that $\max\{\|\tilde T\|, \|T\|\} \le \|T\|+2\epsilon_n = \|E\|^2 + 2\epsilon_n \le 3$. Thus, we have 
$$
\|T^{\alpha/2} - \tilde T^{\alpha/2}\| = 
\|f_\alpha(T^{1/2}) - f_\alpha(\tilde T^{1/2})\| \overset{(i)}{=} 
c \|f_\alpha\|_{B^1_{\infty,1}} \|T^{1/2}-\tilde T^{1/2}\| \overset{(ii)}{\le} 
3^{\alpha+1} c\alpha \epsilon_n.
$$
In the above, $f_\alpha(\cdot)$ equals $x^\alpha$ on the interval $[0,3]$, and is extended to the real line to satisfy the Besov norm bound $\|f_\alpha\|_{B^1_{\infty,1}}\le \|f_\alpha\|_\infty + \|f_\alpha'\|_\infty \le 2\alpha\cdot 3^\alpha$;  (i) follows by \citet[Theorem 1.6.1]{aleksandrov2016operator}, and $c$ is the constant therein; (ii) follows by \eqref{eq:op-norm-bound} and our bound for $\|f_\alpha\|_{B^1_{\infty,1}}$. Combining, we have 
$$
\|\tilde T^{\alpha/2} - T^{\alpha/2}\| \le c_\alpha \epsilon_n^{\alpha\wedge 1}, 
$$
for any $\alpha > 0$. Plugging back to \eqref{eq:lem-rkhs-norm-g}-\eqref{eq:lem-L2-norm-g} completes the proof.

\paragraph{Proof for Claim~\ref{it:lem-symm-result}}
This is a symmetry version of \eqref{eq:lem-approx-hs-claim} and can be proved by repeating the original proof for each case, with all occurrences of $\tilde T$ and $T$ exchanged. The symmetry is obvious for case \ref{it:lem-generic-cond} and \ref{it:lem-second-cond}; for 
for case \ref{it:lem-first-cond}, its condition \eqref{eq:T-cond-relaxed} is equivalent to (recall \eqref{eq:basic-facts})
$
c^{-1}(\|T^{1/2} f\|_2 - \epsilon_n) \le \|\tilde T^{1/2} f\|_2 \le 
c(\|T^{1/2} f\|_2 + \epsilon_n),
$
which implies the symmetric condition of 
$$
c^{-1}(\|\tilde T^{1/2} f\|_2 - c\epsilon_n) \le \|T^{1/2} f\|_2 \le 
c(\|\tilde T^{1/2} f\|_2 + c\epsilon_n).
$$

This completes the proof.
\end{proof}

\begin{remark}
We note that exponentially growing constants w.r.t.~regularity parameters like $\alpha$ are common in the kernel literature. The saturation of the bound at $\alpha>1$ is likely due to our lack of further assumptions on $\tilde T$. 
\end{remark}

\newcommand{\powerSpace}{{\cH}}

\subsection{Proof for \cref{thm:bounded-ill-posedness}}\label{app:proof-thm}

We prove a more general version of the theorem, with \eqref{eq:hs-norm} replaced by the more general \eqref{eq:T-cond-relaxed}.

By \cref{lem:approx-HS-full}, we always have 
\begin{equation}\label{eq:thm-proof-lem-cond}
\|\tilde f\|_{\cH} \le C\|f_0\|_{\gtRKHS[\beta]}. %
\end{equation}
for some $C>0$ that only depends on $\alpha$ and the constant in \eqref{eq:T-cond-relaxed}. 
Let us restrict to $f$ s.t.
\begin{equation}\label{eq:thm-proof-restriction}
\|f-\tilde f\|_2 \ge \|\tilde f-f_0\|_2,
\end{equation}
since other $f$ cannot possibly violate the claimed bound for $\omega'_n$: we would have $\|f-f_0\|_2^2 \le 4\|\tilde f-f_0\|_2^2$. 

Recall the inequalities 
$$
2(\|a\|_2^2+\|b\|_2^2)\ge 
\|a + b\|^2 \ge \|a\|^2 + \|b\|^2 - 2\|a\|\|b\| \ge (1-\zeta^2) \|a\|^2 - (\zeta^{-2}-1)\|b\|^2,
$$
which hold for any $\zeta>0$ and Hilbert norm $\|\cdot\|$. Now, by the assumed condition of functions that appear in $\omega'_n$, \eqref{eq:thm-proof-lem-cond} and the above, 
we have, for all $f\in\cH$,  
\begin{align*}
\|E(f-\tilde f)\|_2^2 +\lambda_n \|f - \tilde f\|_\powerSpace^2 &\le 
2\|E(f-f_0)\|_2^2 + 2\lambda_n \|f\|_\powerSpace^2 + 
    2\|E(f_0-\tilde f)\|_2^2 + 2\lambda_n\|\tilde f\|_\powerSpace^2 \\ 
&\le 
2 \delta_n^2 + %
    2\|E(f_0-\tilde f)\|_2^2 + 2C^2 \lambda_n\|f_0\|_{\gtRKHS[\beta]}^2.
\end{align*}
And by \eqref{eq:thm-proof-restriction}, we have 
\begin{align*}
\|f-f_0\|_2^2 &\le 2\|f-\tilde f\|_2^2 + 2\|\tilde f-f_0\|_2^2  \le 4\|f-\tilde f\|_2^2.
\end{align*}
By the two displays above, it suffices to establish that for some constant $c_\alpha>0$, we have 
\begin{align*}
\sup_{f\in\powerSpace} \frac{\|f-\tilde f\|_2^2}{\|E(f-\tilde f)\|_2^2 + \lambda_n\|f-\tilde f\|_\powerSpace^2} \le 
c_\alpha(\max\{\lambda_n^{1/(1+\alpha)}, \lambda_n \epsilon_n^{-2\alpha}\})^{-1},
\end{align*}
or equivalently, 
\begin{align}\label{eq:thm-proof-goal}
\inf_{f'\in\powerSpace, \|f'\|_2=1}(\|Ef'\|_2^2 + \lambda_n\|f'\|_\powerSpace^2) \ge c_\alpha^{-1} \min\{\lambda_n^{1/(1+\alpha)}, \lambda_n \epsilon_n^{-2\alpha}\},
\end{align}
after which the proof is complete. 

Observe that for any $\|f'\|_2=1$, and any $\zeta\in (1/\sqrt{2}, 1)$, we have 
\begin{align*}
\|Ef'\|_2^2 + \lambda_n\|f'\|_\powerSpace^2 
&\overset{\eqref{eq:T-cond-relaxed}}{\ge} (c^{-1}\|\tilde T^{1/2} f'\|_2 - c\epsilon_n)^2
+ \lambda_n\|f'\|_\powerSpace^2 \\
&\ge (1-\zeta^2) \|c^{-1}\tilde T^{1/2} f'\|_2^2 + \lambda_n\|f'\|_\powerSpace^2 - (\zeta^{-2}-1)c^2\epsilon_n^2 \\ 
&\overset{\eqref{eq:H-defn}}{=}  \<f', ((1-\zeta^2)c^{-2}\tilde T + \lambda_n \tilde T^{-\alpha}) f'\>_2 - (\zeta^{-2}-1)c^2\epsilon_n^2 \\ 
&\ge 
\<f', ((1-\zeta^2)c^{-2}\tilde T + \lambda_n \tilde T^{-\alpha}) f'\>_2 - 2(1-\zeta^2)c^2\epsilon_n^2, \numberthis\label{eq:thm-proof-intermediate}
\end{align*}
Since $\|f'\|_2=1$, the first term above can be bounded by a smallest eigenvalue: %
denote by $\{e_i\}$ the eigenvalues of $\tilde T$, and define $u := 1-\zeta^2$, then we have 
\begin{align*}
&\phantom{=}\inf_{\|f'\|_2=1}
\<f', ((1-\zeta^2)c^{-2}\tilde T + \lambda_n \tilde T^{-\alpha}) f'\>_2 - 2(1-\zeta^2)c^2\epsilon_n^2\\
&= 
\inf_{i\in\mb{N}}(u c^{-2}e_i + \lambda_n e_i^{-\alpha}) - 2uc^2\epsilon_n^2 \\&
\ge 
\inf_{a>0}(u a + \lambda_n c^{-2\alpha}a^{-\alpha}) - 2uc^2\epsilon_n^2 \\ &
= 2u^{\frac{\alpha}{\alpha+1}}(c^{-2\alpha}\lambda_n)^{\frac{1}{\alpha+1}} - 2uc^2\epsilon_n^2, ~~\text{for all } u\in (0, 1/2).
\end{align*}
The choice of $$
u := \frac{1}{2}\min\left\{1, c^{-2\alpha}\lambda_n (c\epsilon_n)^{-2(\alpha+1)}
\right\}
$$
produces the bound 
$$
\|Ef'\|_2^2 + \lambda_n\|f'\|_\powerSpace^2 \ge (2^{\frac{1}{\alpha+1}} - 1)\min\{
    c^{-2\alpha/(\alpha+1)}
 \lambda_n^{1/(\alpha+1)}, c^{-4\alpha}\lambda_n \epsilon_n^{-2\alpha} 
\}.
$$
This completes the proof of \eqref{eq:thm-proof-goal}, and thus the theorem.\qed

Recall the constant $c=1$ when \eqref{eq:hs-norm} holds. And 
as a sanity check, we can verify the example constructed for maximizing \eqref{eq:psiJ} does not violate \eqref{eq:thm-proof-goal}: for $\tilde E = \sum_i s_i \psi_i\otimes \psi_i$ to satisfy the norm condition, we must have $s_i \le \|E\psi_i\|_2 + \epsilon_n$, which implies a lower bound for the RKHS norm.

%% file: app-proof.tex
\section{Deferred Proofs: Representation Learning}

\subsection{Proof for Lemma~\ref{lem:loss-equivalence}}\label{app:reproof-loss-equivalence}

For \eqref{eq:loss-equiv-1}, the first equivalence holds because
$$
\begin{aligned}
\divLB[h] + 1 &= \EE_{P_{zx}} h(\bz,\bx) - \EE_{P_z\otimes P_x} h^2(\bz,\bx) 
= \EE_{P_z\otimes P_x}\left(\frac{dP_{zx}}{d(P_z\otimes P_x)} h - h^2\right) = 
2\<h_0, h\>_{L_2(P_z\otimes P_x)} - \|h\|^2_{L_2(P_z\otimes P_x)} \\ &
= -\|h-h_0\|_{L_2(P_z\otimes P_x)}^2 + \|h_0\|_{L_2(P_z\otimes P_x)}^2,
\end{aligned}
$$
where the first equality is derived in the text. It remains to prove 
$
\|\tilde E - E\|_{HS} = \|h - h_0\|_2.
$
For this purpose we first claim that for all $f\in L_2(P_x)$, we have
\begin{equation}\label{eq:lem-le-claim}
(Ef)(\bz) = \int h_0(x, \bz) f(x) P_x(dx).~~a.s.
\end{equation}
To prove this, observe that for any $g\in L_2(P_z)$, we have 
\begin{align*}
\<E f, g\>_2 = \EE_{P_{xz}} f(\bx)g(\bz) = \EE_{P_z\otimes P_x} h_0(\bz,\bx) f(\bx) g(\bz)
= \int \left(\int h_0(x,z)f(x)P_x(dx)\right) g(z) dP_z.
\end{align*}
Thus, the right-hand side of \eqref{eq:lem-le-claim} fulfills the definition of conditional expectation. Now,  let $\{\phi_j\}, \{\psi_i\}$ be any orthonormal bases for $L_2(P_z)$, $L_2(P_x)$, respectively. Then
\begin{align*}
\|\tilde E - E\|_{HS}^2 &= \sum_{i=1}^\infty\sum_{j=1}^\infty \<\phi_j, (\tilde E - E) \psi_i\>_2^2 \\ &\overset{(i)}{=}
 \sum_{i=1}^\infty\sum_{j=1}^\infty \left(\int \phi_j(z)\left(\int (h-h_0)(x,z) \psi_i(x) P_x(dx)\right)P_z(dz)\right)^2 \\ &= 
 \sum_{i=1}^\infty\sum_{j=1}^\infty \<h-h_0, \phi_j\otimes\psi_i\>_{L_2(P_z\otimes P_x)}^2 \\ &= 
\|h-h_0\|_2^2,
\end{align*}
where (i) follows by \eqref{eq:lem-le-claim} and the definition of $\tilde E$. This completes the proof for \eqref{eq:loss-equiv-1}. 

Before we prove \eqref{eq:chisq-trace}, we first prove the following claim, which also appeared in Section 3: for any $h(z,x) = \Phi(z)^\top\Psi(x) = \sum_{i=1}^{J_n} \phi_i(z)\psi_i(x)$ satisfying \eqref{eq:loss-equiv-1}, there exists a linear transform of $\Psi$, $\tilde\psi_i := \sum_{j=1}^{J_n} a_{ij} \psi_j$, s.t.
\begin{equation}\label{eq:loss-equivalence}
\sum_{i=1}^{J_n}\|E\tilde\psi_i\|_2^2 \ge 
\sum_{i=1}^{J_n}\|E\bar\psi_i\|_2^2 -\epsilon_n^2.
\end{equation}
To establish this, 
let $\cD_{\mrm{col}}(\Psi) := \sup_{\Phi':\cZ\to\RR^k} \divLB[\Phi'^\top\Psi]$ be the collapsed population objective. 
Define the whitened feature maps $S := \EE_{P_x}[\Psi(\bx)\Psi^\top(\bx)], \tilde\Psi(\cdot) := (S^{-1/2})^\top\Psi(\cdot)$. Then we have $h=(S^{1/2}\Phi)^\top\tilde\Psi$, and 
\begin{align*}
\cD_{\mrm{col}}(\Psi) + 1 &= 
 \sup_{\tilde\Phi:\cZ\to\RR^k} 2\EE_{P_{zx}} \tilde\Phi(\bz)^\top\tilde\Psi(\bx) - \EE_{P_z\otimes P_x}(\tilde\Phi(\bz)^\top\tilde\Psi(\bx))^2 \\ & 
= \sup_{\tilde\Phi} 2\EE_{P_{zx}}\sum_{i=1}^k \tilde\phi_i(\bz)\tilde\psi_i(\bx) - 
\sum_{i=1}^{J_n}\sum_{i'=1}^{J_n} \EE_{P_z\otimes P_x} \tilde\phi_i(\bz)\tilde\psi_i(\bx)\tilde\phi_{i'}(\bz)\tilde\psi_{i'}(\bx) \\ &
= \sup_{\tilde\Phi} 2\EE_{P_{z}}\sum_{i=1}^k \tilde\phi_i(\bz)(E\tilde\psi_i)(\bz) - 
\sum_{i=1}^{J_n}\sum_{i'=1}^{J_n} \EE_{P_z}(\tilde\phi_i(\bz)\tilde\phi_{i'}(\bz))
\underbrace{\EE_{P_x}(\tilde\psi_i(\bx)\tilde\psi_{i'}(\bx))}_{=\delta_{ii'}} \\ & 
= \sum_{i=1}^{J_n}\sup_{\tilde\phi_i}\: \<2 E\tilde\psi_i-\tilde\phi_i, \tilde\phi_i\>_{L_2(P_z)}
= \sum_{i=1}^{J_n} \|E\tilde\psi_i\|_{L_2(P_z)}^2. \numberthis\label{eq:proof-le-1}
\end{align*}
The proof of \eqref{eq:loss-equivalence} completes by observing 
\begin{equation}\label{eq:proof-le-2}
\divLB[h] \le \cD_{\text{col}}(\Psi) \le \divLB[h_0] = \sum_{i=1}^\infty s_i^2(E) - 1.
\end{equation}

Finally, 
for \eqref{eq:chisq-trace}, observe that setting $\tilde\psi_i \gets \gtsv{i}, \tilde\phi_i\gets \gtsv{i}$ attains the supremum in \eqref{eq:proof-le-1}, at any fixed $J_n$. Setting $J_n\to\infty$ shows that $\cD_{\chi^2}(P_{zx}\Vert P_z\otimes P_x)\ge \sum_{i=1}^\infty s_i^2(E) - 1$. Combining with \eqref{eq:proof-le-2} completes the proof.
\qed

\subsection{Proof for \eqref{eq:kernel-used}}\label{app:popu-kernel-deriv}

\begin{lemma}[population kernel]\label{lem:popu-kernel}
Let $k_{popu}$ be defined as in \eqref{eq:kernel-used}. 
Then its integral operator equals $(\tilde E^\top \tilde E)^{\alpha}$, where $\tilde E$ is defined as in \cref{lem:loss-equivalence}. 
\end{lemma}

\begin{proof}
It suffices to show the integral operator defined by $k$ equals $(\tilde E^\top \tilde E)^\alpha$. 
First consider the $\alpha=1$ case. We have, for all $f'\in L_2(P_x)$ and $P_x$-almost all $x$:
\begin{align*}
 (\tilde E^\top \tilde E f)(x) &= 
\left(
    \tilde E^\top \int h(\cdot,x') f(x') dP_x(x')
\right)(x) \\ &= 
\int h(x,z) \left(
    \int h(z,x') f(x') dP_x(x')
\right) dP_z = 
\EE_{x',z\sim P_x\otimes P_z}\Psi(x)^\top \Phi(z) \Phi(z)^\top \Psi(x') f(x')
\\ &= 
\int (\Psi(x)^\top \Sigma_z \Psi(x')) f(x') P(dx)
= \int k(x,x') f(x') P(dx).
\end{align*}

When $\alpha\ne 1$, introduce the whitened feature maps $\Psi'(x) = \Sigma_x^{-1/2}\Psi(x), \Phi'(z) = \Sigma_x^{1/2}\Phi(z)$ which defines the same $h(z,x)$. 
Recall our convention of using $\Psi'$ to also refer to the operator $\Psi': L_2(P_x)\to\RR^J$ (\cref{rmk:addi-conventions}). 
Then the above display implies 
$$
\tilde E^\top \tilde E = \Psi'^\top \Sigma_z' \Psi' = \Psi'^\top \Sigma_x^{1/2} \Sigma_z\Sigma_x^{1/2} \Psi',
$$
and we have, since $\Psi'\Psi'^\top=I$, 
$$
(\tilde E^\top \tilde E)^\alpha = \Psi'^\top (\Sigma_x^{1/2} \Sigma_z\Sigma_x^{1/2})^\alpha \Psi'.
$$ 
The claim is proved by plugging in the definition of $\Psi'$ and computing the integral operator.
\end{proof}

\subsection{Proof for \cref{prop:dr-est}}\label{app:proof-est}

We prove a slightly more general version of the proposition, as stated below.

\renewcommand{\prodHypoSpace}{\cF_{h,n}}
\begin{proposition}[oracle inequality]\label{prop:est-general}
Let $B,C_1>0$ be constants, $h^*: \cZ\times \cX\to\RR$ be an arbitrary, square-integrable function.  
Suppose $\{\prodHypoSpace\}$ is a sequence of hypothesis spaces indexed by $n$, which are all star-shaped around $h^*$, and satisfy 
$$
\sup_{h\in\prodHypoSpace}\|h\|_\infty\le B, ~~log N(\prodHypoSpace, \|\cdot\|_\infty, \epsilon) \lesssim C_1 \epsilon^{-2}. 
$$
Let $\hat h_n, \cD_n$ be defined as in the text, and $\delta_{n,\prodHypoSpace} \ge n^{-1/2}$ solve the inequality 
$$
\frac{64}{\sqrt n} \int_{\delta^2}^\delta \sqrt{\log N(\prodHypoSpace, \|\cdot\|_\infty, t)} dt \le \delta^2.
$$
Then for some $C_2>0$, we have, for any $\zeta \in (0, 1/8)$, 
\begin{align*}
\PP\left(
\divLB[h_0] \:-\: \divLB[\hat h_n] \le 
C_2\left(\divLB[h_0] -\divLB[h^*] + B^4 \delta_{n,\prodHypoSpace}^2 + \frac{\log\zeta^{-1}}{n}
\right)\right) \ge 1-\zeta.
\end{align*}
\end{proposition}

\cref{prop:est-general} generalizes the claimed inequality in the main text. To see this, 
let $q\in (0, 1]$ be the smallest real number s.t.~$\log N(\prodHypoSpace, \|\cdot\|_\infty, \epsilon) \lesssim \epsilon^{-2q}$ may hold. Then we can choose $\delta_{n,\prodHypoSpace}^2 \asymp n^{-1/(1+q)}$, recovering the claim in the text. 

We now prove the proposition.

\newcommand{\prodLnorm}[1]{\|#1\|_{L_2(P_z\otimes P_x)}}

Denote by $P_{zx,n},P_{z,n},P_{x,n}$ the empirical measures determined by the training samples, and $(P_z\otimes P_x)_n$ the empirical measure defined by $\{(z_i,x_j): i\ne j\}$, 
so that we can write 
\begin{equation}\label{eq:emp-obj-expr}
\cD_n(h) = \EE_{P_{zx,n}} 2h - \EE_{(P_z\otimes P_x)_n} h^2. 
\end{equation}
We analyze the concentration of the two terms in turn.

For the first term above, by \citet[Theorem 14.20 (a)]{wainwright2019high}, there exists some universal $c_1>0$ s.t.~w.p.~$1-\zeta/8$, 
$$
|\EE_{P_{zx,n}}(h-h^*) - \EE_{P_{zx}} (h-h^*)| \le c_1 B \delta_n(\|h-h^*\|_{L_2(P_{zx})}+\delta_n),~~\forall h\in\prodHypoSpace,
$$
where $$
\delta_n := \delta_{n,\prodHypoSpace} + \frac{\log (c_1\zeta^{-1})}{n},
$$
and $\delta_{n,\prodHypoSpace}$ upper bounds the critical radius of the local Rademacher complexity of $\prodHypoSpace$.

Note that 
\begin{align*}
\|h-h^*\|_{L_2(P_{zx})}^2 &= \EE_{P_z\otimes P_x} (h_0-h^*+h^*) (h-h^*)^2 
\\ &\le 
\prodLnorm{h_0-h^*}\prodLnorm{h-h^*}\|h-h^*\|_\infty + \|h^*\|_\infty \prodLnorm{h-h^*}^2 \\ & 
\le 2B(\prodLnorm{h-h^*}^2 + \prodLnorm{h_0-h^*}^2), 
\end{align*}
where the last line follows from 
the inequality $ab\le (\epsilon^2 a^2 + \epsilon^{-2} b^2)/2$ (for all $\epsilon>0$). 
Combining the above bounds with another application of the same elementary inequality, we have, on the above high-probability event, 
\begin{equation}\label{eq:est-first-term}
|\EE_{P_{zx,n}} (h-h^*) - \EE_{P_{zx}} (h-h^*)| \le \frac{\prodLnorm{h-h^*}^2+\prodLnorm{h^*-h_0}^2}{8} + 
C_f B^3 \delta_n^2,
\end{equation}
for some $C_f>0$. 

Analysis of the second term in \eqref{eq:emp-obj-expr} follows \citet{clemencon2006,clemencon_ranking_2008}: 
Introduce the notations $$w:=(z,x),~ u_h(w, w') := \frac{1}{2}(h(z,x')^2+h(x,z')^2)$$ so that
$$
\EE_{(P_z\otimes P_x)_n} h^2 = \frac{1}{n(n-1)}\sum_{i\ne j} u_h(w_i, w_j), ~~
\EE_{P_z\otimes P_x} h^2 = \EE_{P_z\otimes P_x} h^2(\bz,\bx)
= \EE_{\bw,\bw'\sim P_w} u_h(\bw,\bw'),
$$
where $P_w = P_{zx}$. (The last equality above holds because $\bw,\bw'$ are i.i.d., and thus $(\bz,\bx')\sim P_z\otimes P_x$.) 
Recall the Hoeffding decomposition for $U$ statistic:
\begin{align*}
\EE_{(P_z\otimes P_x)_n} h^2 - \EE_{P_z\otimes P_x} h^2 &= 2T_n(h) + W_n(h), \\ 
T_n(h) &:= 
\frac{1}{n}\sum_{i=1}^n \EE_{\bw\sim P_w} u_h(w_i, \bw) - \EE_{\bw,\bw'\sim P_w} u_h(\bw,\bw'), \\
W_n(h) &:=  
\frac{1}{n(n-1)}\sum_{i\ne j} u_h(w_i,w_j) - \EE_{\bw,\bw'\sim P_w} u_h(\bw,\bw') - 2T_n(h).
\end{align*}
We first show that $W_n$ is insignificant. By \citet[Theorem 11]{clemencon_ranking_2008}, 
w.p.~$1-\zeta/8$, we have 
\begin{equation}\label{eq:est-wn}
\sup_{h\in\prodHypoSpace}|W_n(h)| \lesssim \frac{\EE Z_\epsilon + \EE U_\epsilon \sqrt{\log \zeta^{-1}} + \EE M\log \zeta^{-1} + n\log \zeta^{-1}
}{n^2},
\end{equation}
where
\begin{align*}
Z_\epsilon &= \sup_{h\in\prodHypoSpace}\Bigg|\sum_{i,j}\epsilon_i\epsilon_j u_h(w_i,w_j)\Bigg|, ~~
U_\epsilon = \sup_{h\in\prodHypoSpace}\sup_{\substack{\alpha\in\RR^n,\\\|\alpha\|_2\le 1}}\sum_{i,j}\epsilon_i\alpha_j u_h(w_i,w_j), ~~
M = \sup_{h\in\prodHypoSpace}\max_{1\le k\le n}\sum_{i=1}^n\epsilon_i u_h(w_i,w_k).
\end{align*}
and $\{\epsilon_i\}$ are i.i.d.~Rademacher random variables independent of $\{w_i\}$. We address the three terms in turn.
\begin{enumerate}[leftmargin=*,label=(\roman*)]
\item Since $\|u_h\|_\infty\le B^2$, we have $M\le n B^2$.
\item\label{it:Z-eps} For $Z_\epsilon$, define $X_h := \frac{1}{n}\sum_{i\ne j}\epsilon_i\epsilon_j u_h(w_i,w_j)$, so that 
$
Z_\epsilon \le %
n\sup_{h\in\prodHypoSpace}(B^2 + X_h). 
$
 by \citet[Proposition 2.2, 2.6]{arcones1993limit}, we have, for all $\delta'>0$ and $\cU := \{u_h: h\in\prodHypoSpace\}$, 
\begin{align*}
\EE_\epsilon\sup_{h\in\prodHypoSpace} X_h &\le 
\EE_{\epsilon} \sup_{h,h'\in\prodHypoSpace}(X_h - X_{h'}) \le 
\int_{\delta'}^D \log N(\cU, \rho, \delta) d\delta + 
    \EE_{\epsilon} \sup_{\substack{h,h'\in\prodHypoSpace,\\ \rho(h,h')\le\delta'}}  (X_h-X_{h'}), 
\end{align*}
where $
\rho(h,h') := (\EE_\epsilon (X_h-X_h')^2)^{\frac{1}{2}} 
\le \|u_h - u_{h'}\|_\infty\le 2B \|h-h'\|_\infty.
$ Plugging back, we have 
\begin{align*}
\EE_\epsilon\sup_{h\in\prodHypoSpace} X_h &\le 
\int_{\delta'}^D \log N(\prodHypoSpace,\|\cdot\|_\infty,\delta/2B)d\delta + n\delta' \lesssim  
B^{2q} \int_{\delta'}^D \delta^{-2q}d\delta + n\delta',
\end{align*}
where $D=\sup_{h,h'}\rho(u_h,u_{h'})\lesssim B^2$. 
Thus, when $2q\le 1$, we have $\EE_\epsilon\sup_{h\in\prodHypoSpace} X_h \lesssim B^2 + B \log D$; when $q\in (1/2, 1]$,  we have $
\EE_\epsilon \sup_{h\in\prodHypoSpace} X_h \lesssim Bn^{1-1/(2q)}  \lesssim B n\delta_{n,\prodHypoSpace}^2.
$
Combining, we have 
$$
\frac{1}{n^2}\EE_\epsilon Z_\epsilon \lesssim \frac{B^2}{n} + B \delta_{n,\prodHypoSpace}^2.
$$
\item 
Finally, for $U_\epsilon$, we have
$$
\EE_\epsilon U_\epsilon^2 \le \EE_\epsilon \sup_{h\in\prodHypoSpace} \sum_{j=1}^n \Bigg(\sum_{i=1}^n \epsilon_i u_h(w_i,w_j)\Bigg)^2 =
\EE_\epsilon \sup_{h\in\prodHypoSpace} \sum_{i,i'} \epsilon_i\epsilon_{i'} \underbrace{\sum_{j=1}^n u_h(w_i,w_j)u_h(w_{i'}, w_j)}_{=: \bar u_h(w_i, w_{i'})}.
$$
For any $\delta>0$ and $h,h'$ s.t.~$\|h - h'\|_\infty\le \delta$, we have (noting that $\|h-h'\|_\infty\le 2B$ always hold)
$$
\|\bar u_h - \bar u_{h'}\|_\infty \le n(4B^2\delta + \delta^2)\le 5nB^2\delta.
$$
Thus, we can follow \ref{it:Z-eps}, with $u_h$ replaced by $\bar u_h$ and the constants replaced appropriately, leading to  
\begin{align*}
\frac{1}{n} \EE_\epsilon U_\epsilon^2 &\le
    5 n B^3 %
    + \inf_{\delta'>0}\Big(
    \int_{\delta'}^{10 n B^3} \log N(\prodHypoSpace, \|\cdot\|_\infty, \delta / 5nB^2) d\delta + n\delta' \Big), %
\\ 
\EE_\epsilon U_\epsilon^2 &\lesssim n^2(B^3 + B^2 n\delta_{n,\prodHypoSpace}^2),
\\
n^{-2}\EE_\epsilon U_\epsilon &\le n^{-2}\sqrt{\EE_\epsilon U_\epsilon^2} \lesssim n^{-1}B^{3/2} + Bn^{-1/2}\delta_{n,\prodHypoSpace}.
\end{align*}
\end{enumerate}
Plugging the bounds for $U_\epsilon, Z_\epsilon$ and $M$ back to \eqref{eq:est-wn}, and recalling that $\delta_{n,\prodHypoSpace}\ge n^{-1/2}$, we can see that, for some $C_W>0$, 
\begin{equation}\label{eq:est-wn-fin}\tag{\ref{eq:est-wn}'}
\PP\Big(
\sup_{h\in\prodHypoSpace}|W_n(h)| \le C_W \big(B\delta_{n,\prodHypoSpace}^2 + B^2 n^{-1}\log\zeta^{-1}\big)
\Big) \ge 1 - \frac{\zeta}{8}.
\end{equation}

It remains to deal with $T_n$. Introduce $v_h(w) :=  \EE_{\bw\sim P_w} w_h(w,\bw)$, so that $T_n = \frac{1}{n}\sum_{i=1}^n v_h(w_i)  - \EE_\bw v_h$. 
For any $h^*\in L_2(P_{zx}), w=(z,x)$, we have 
\begin{align*}
v_h(w) &= v_{h-h^*}(w) + \EE_{(\tilde\bz,\tilde\bx)\sim P_{zx}}(h^*(\tilde\bz, x) h(\tilde\bz, x) + h^*(z, \tilde\bx) h(z, \tilde\bx)) + v_{h^*}(w),  \\ 
&=: v_{h-h^*}(w) + \tilde v_{h,h^*}(w) + v_{h^*}(w), 
\\
T_n(h) &= \frac{1}{n}\sum_{i=1}^n (v_{h-h^*}(w_i) -\EE_{\bw} v_{h-h^*}) + (\tilde v_{h,h^*}(w_i) - \EE_\bw \tilde v_{h,h^*}) + (v_{h^*}(w_i) - \EE_\bw v_{h^*}) \\ 
&=: T_{n,1}(h) + T_{n,2}(h) + T_{n,3}.
\end{align*}
Only the first two terms depend on $h$; we address them in turn.
\begin{enumerate}
\item For $T_{n,1}$, by Lemma~\ref{lem:Tn1} below we have, for some $c_1,C>0$,  
$$
\PP\Bigg(\forall h\in\prodHypoSpace,~ |T_{n,1}(h)| \le \frac{1}{4}\|h-h^*\|_{L_2(P_z\otimes P_x)}^2 + C B^2 \delta_{n,\prodHypoSpace}^2 + CB^4\frac{\log\zeta^{-1}}{n}\Bigg) \ge 1 - \frac{\zeta}{8}.
$$
Clearly, a similar bound holds for $|T_{n,1}(h) - T_{n,1}(h^*)|$. 
\item $T_{n,2}(h)$ is a sample average of 1-Lipschitz functions of $\tilde v_{h,h^*}$. Thus, by \citet[Theorem 14.20]{wainwright2019high}, there exists some $C'>0$ s.t.~w.p.~$\ge 1 - \exp(-C' n\delta_{n,\tilde\cV}^2)$, we have, for all $h\in\prodHypoSpace$, 
$$
|T_{n,2}(h) - T_{n,2}(h^*)| \le 10 \delta_{n,\tilde\cV}(\|\tilde v_{h,h^*} - \tilde v_{h^*,h^*}\|_{L_2(P_w)} + \delta_{n,\tilde\cV}), 
$$
where $\delta_{n,\tilde\cV}$ upper bounds the critical radius of $\tilde\cV := \{\tilde v_{h,h^*}(\cdot): h\in\prodHypoSpace\}$. By definition we have 
\begin{align*}
\|\tilde v_{h,h^*} - \tilde v_{h^*, h^*}\|_{L_2(P_w)} 
&= 
\big[\EE_{(\bz,\bx)\sim P_{zx}}(\EE_{(\tilde\bz,\tilde\bx)\sim P_{zx}}\{[h^*(h-h^*)](\tilde\bz,\bx) + [h^*(h-h^*)](\bz,\tilde\bx)\})^2\big]^{\frac{1}{2}} \\ 
&\le 2 \|h^*\|_\infty \|h-h^*\|_{L_2(P_z\otimes P_x)},
\\
\|\tilde v_{h,h^*} - \tilde v_{h', h^*}\|_\infty &\le 2\|h^*\|_\infty\|h-h'\|_\infty.
\end{align*}
The second inequality also implies we can choose 
$\delta_{n,\tilde\cV}\asymp \|h^*\|_\infty \delta_{n,\prodHypoSpace}$. Plugging back, we have, for some $C''>1$, and all $\delta_{n,\tilde\cV} \ge C''B \delta_{n,\prodHypoSpace}$, 
\begin{align*}
e^{-n\delta_{n,\prodHypoSpace}^2 / C''} &\ge 
\PP(\exists h\in\prodHypoSpace, ~|T_{n,2}(h) - T_{n,2}(h^*)| \ge C''(B^2 \delta_{n,\prodHypoSpace}\|h-h^*\|_{L_2(P_z\otimes P_x)} + B^2\delta_{n,\prodHypoSpace}^2))  \\ &
\ge 
\PP\Big(\exists h\in\prodHypoSpace, ~|T_{n,2}(h) - T_{n,2}(h^*)| \ge \frac{1}{4}\|h-h^*\|_{L_2(P_z\otimes P_x)}^2 + ((C''B)^4+B^2)\delta_{n,\prodHypoSpace}^2\Big).
\end{align*}
We use $\delta_{n,\tilde\cV} = C''B(\delta_{n,\prodHypoSpace} + \sqrt{\log \zeta^{-1}/n})$. 
\end{enumerate}
Combining the results above, we have, for some $C_T>0$,
\begin{equation}\label{eq:est-tn}
\PP\Big(\forall h\in\prodHypoSpace,~|T_n(h) - T_{n,3}| \le 
    \frac{1}{2}\|h-h^*\|_{L_2(P_z\otimes P_x)}^2 + C_T B^4 \Big(\delta_{n,\prodHypoSpace}^2 + \frac{\log\zeta^{-1}}{n}\Big)
\Big) \ge 1 - \frac{\zeta}{4}.
\end{equation}
Combining \eqref{eq:est-first-term}, \eqref{eq:est-wn-fin}, \eqref{eq:est-tn}, we can see that for some $C>0$, w.p.~$\ge 1-\zeta$ we have, for all $h\in\prodHypoSpace$, 
$$
|\cD_n(h) - \cD_n(h^*) - \EE_{P_{zx}}(\cD_n(h)-\cD_n(h^*))| \le
\frac{3}{4}\prodLnorm{h-h^*}^2 + \frac{1}{4}\prodLnorm{h^*-h_0}^2 + 
C B^4 \left(\delta_{n,\prodHypoSpace}^2 + \frac{\log\zeta^{-1}}{n}\right).
$$
For $\hat h_n = \arg\max_{h\in\prodHypoSpace} \cD_n(h)$, the LHS above is lower bounded by 
$
\EE_{P_{zx}}(\cD_n(h^*) -\cD_n(h)). 
$ 
Moreover, observe that 
\begin{align*}
\EE_{P_{zx}}\cD_n(h) &= \EE_{P_{zx}} 2h - \EE_{P_z\otimes P_x} h^2 
= \EE_{P_z\otimes P_x}(2h_0 h - h^2) = 
\prodLnorm{h_0}^2 + \prodLnorm{h_0-h}^2. 
\end{align*}
Thus, we have 
\begin{align*}
    &\phantom{=}
\prodLnorm{h_0 - \hat h}^2 - \prodLnorm{h_0 - h^*}^2 \\&\le
 \frac{3}{4}\prodLnorm{\hat h-h^*}^2 + \frac{1}{4}\prodLnorm{h_0-h^*}^2 + C B^4 
  \left(\delta_{n,\prodHypoSpace}^2 + \frac{\log\zeta^{-1}}{n}\right).
 \\ 
 &\le
 \frac{7}{8}\prodLnorm{\hat h-h_0}^2 + \frac{11}{2}\prodLnorm{h_0-h^*}^2 + C B^4  \left(\delta_{n,\prodHypoSpace}^2 + \frac{\log\zeta^{-1}}{n}\right),
\end{align*}
or
$$
\prodLnorm{h_0 - \hat h}^2 \le 52 \prodLnorm{h_0-h^*}^2 + 8C B^4  \left(\delta_{n,\prodHypoSpace}^2 + \frac{\log\zeta^{-1}}{n}\right).
$$
The proof completes by observing $\divLB[h_0] - \divLB[h] = \prodLnorm{h_0-h}^2.$ \qed

In the following lemma, note that the critical radius is defined w.r.t.~the different measure of $P_z\otimes P_x$; however, it can still be bounded using that of the sup norm.
\begin{lemma}\label{lem:Tn1}
In the setting of Proposition~\ref{prop:est-general}, there exists some $c_1>0$ s.t.~for any $t>\delta_{n}$, 
$$
\PP\Big(\forall h\in\prodHypoSpace, |T_{n,1}(h)| \ge \frac{1}{4}\|\Delta h\|_{L_2(P_z\otimes P_x)}^2 + \frac{B^2}{2}t^2\Big) \le 2e^{-c_1 n t^2/B^2},
$$
where $\delta_n$ upper bounds the critical radius of the local Rademacher complexity $\cR_n(B_2(r;\prodHypoSpace))$, and $B_2(r;\prodHypoSpace) := \{h-h^*: h\in\prodHypoSpace, \|h-h^*\|_{L_2(P_z\otimes P_x)}\le r\}$.
\end{lemma}
\begin{proof}
The proof modifies \citet[Section 14.1.3]{wainwright2019high} with two additions of Jensen's inequality. 
Let $$
Z_n(r) := \sup_{\Delta h\in B_2(r;\prodHypoSpace)} |T_{n,1}(h)| = 
\sup_{\Delta h\in B_2(r;\prodHypoSpace)} \Big|\frac{1}{n}\sum_{i=1}^n v_{\Delta h}(w_i) - \EE_{\bw} v_{\Delta h}
\Big|.
$$ Following the same scaling argument as in their Lemma 14.8,\footnote{Note we have $T_{n,1}(h^* + \alpha\Delta h) = \alpha^2 T_{n,1}(h^*+\Delta h)$, so the same argument applies.}, it suffices to establish that
\begin{equation}\label{eq:lem-Tn1-goal}
\PP\Big(Z_n(r) \ge \frac{B r\delta_n}{4}+\frac{B^2s^2}{4}\Big) \le 2 e^{-c_2 n \min\{s^4/r^2, s^2\}}, 
\end{equation}
for some $c_2>0$ and all $r,s > \delta_n$. 
We first bound the expectation. We have 
\begin{align*}
\EE_{\{w_i\}} Z_n(r) &\overset{(i)}{\le} 2\EE_{\{w_i,\epsilon_i\}} \sup_{\Delta h\in B_2(r;\prodHypoSpace)} \Big|\frac{1}{n}\sum_{i=1}^n \epsilon_i v_{\Delta h}(w_i)\Big| = 
2\EE_{\{w_i,\epsilon_i\}} 
\sup_{\Delta h\in B_2(r;\prodHypoSpace)} \Big|\frac{1}{n}\sum_{i=1}^n \epsilon_i \EE_{\tilde\bw} u_{\Delta h}(w_i,\tilde\bw)\Big| \\ 
&\overset{(ii)}{\le}
2\EE_{\{w_i,\epsilon_i\}} \EE_{\{\tilde w_i\}}
\sup_{\Delta h\in B_2(r;\prodHypoSpace)} \Big|\frac{1}{n}\sum_{i=1}^n \epsilon_i u_{\Delta h}(w_i,\tilde w_i)\Big| \\ 
&=
2\EE_{\{w_i,\tilde w_i,\epsilon_i\}} 
\sup_{\Delta h\in B_2(r;\prodHypoSpace)} \Big|\frac{1}{2n}\sum_{i=1}^n \epsilon_i (\Delta h^2(z_i,\tilde x_i) + \Delta h^2(\tilde z_i,x_i))\Big| \\
&\le 
2\EE_{\{z_i,\tilde x_i,\epsilon_i\}} 
\sup_{\Delta h\in B_2(r;\prodHypoSpace)} \Big|\frac{1}{n}\sum_{i=1}^n \epsilon_i \Delta h^2(z_i,\tilde x_i)\Big| %
\overset{(iii)}{\le}
4B\EE_{\{z_i,\tilde x_i,\epsilon_i\}} 
\sup_{\Delta h\in B_2(r;\prodHypoSpace)} \Big|\frac{1}{n}\sum_{i=1}^n \epsilon_i \Delta h(z_i,\tilde x_i)\Big| \\ 
&= 4B\cR_n(B_2(r;\prodHypoSpace)) \overset{(iv)}{\le} \frac{B r\delta_n}{4}.
\end{align*}
In the above, 
$\cR_n(B_2(r;\prodHypoSpace))$ is the Rademacher complexity of $B_2(r;\prodHypoSpace)$; (i), (iii), (iv) follow the same reasoning as \cite{wainwright2019high}: they apply symmetrization, Lipschitz contraction of $\cR_n$, and the monotonicity of $\cR_n(B_2(r;\prodHypoSpace))/r$, respectively. (ii) invokes Jensen's inequality and exchanges $\sup$ and $\EE$. %

Now we establish the tail bound above the expectation. Observe that 
\begin{align*}
\EE_{\bw}v_{\Delta_h}^2 &= 
\EE_{\bw} (\EE_{\tilde\bw} u_{\Delta h}(\bw,\tilde\bw))^2 \overset{(i)}{\le} 
\EE_{\bw,\tilde\bw} u_{\Delta h}(\bw,\tilde\bw)^2 \overset{(ii)}{\le} 
\|u_{\Delta h}\|_\infty \EE_{\bw,\tilde\bw} u_{\Delta h}(\bw,\tilde\bw) \le 
B^2 \EE_{\bz,\tilde\bx} \Delta h^2(\bz,\tilde\bx) \le B^2 r^2. 
\end{align*}
where (i) is Jensen's inequality, and (ii) follows by the non-negativity of $u$. Applying Talagrand's inequality as in \cite{wainwright2019high} establishes \eqref{eq:lem-Tn1-goal}. 
\end{proof}

%% file: app-proof-extra-repr.tex
\section{Additional Results: Representation Learning}

\subsection{Overview of Results}\label{app:additional-results}

In this section, \cref{prop:kern-est} bounds the error using the estimated kernel \eqref{eq:kernel-used}. 
\cref{prop:approx-general} establishes a generic approximation result for the spectral representation learning objective $\divLB$. \cref{prop:srl-running-example} and \cref{cor:srl-running-ex-alpha-2} illustrate it on our main running example. %

\begin{proposition}[estimation error of the population kernel]\label{prop:kern-est}
Let $k_x$ be defined by \eqref{eq:kernel-used} in the text. %
Let $T_x$ be the $P_x$-integral operator of $k_x$, and $\tilde T := T_x^{1/\alpha}$.  
Suppose 
$(\hat\Sigma_x,\hat\Sigma_z)$ are computed using $n$ i.i.d.~samples $\Dcov = \{(z_{c,i},x_{c,i})\}$ independent of $\hat h_n$. Then there exist $c',c''>0$ s.t.~
\begin{enumerate}[leftmargin=*,label=(\roman*)]
    \item when $\alpha=1$, with $\Dcov$-probability $\ge 1-\zeta$ w.r.t.~$(\hat\Sigma_x,\hat\Sigma_z)$, we have 
$\tilde T$ satisfies \eqref{eq:T-cond-relaxed}, with $c=1$, and $\epsilon_n^2$ bounded by 
$$
\epsilon_n^2 \le \epsilon_{n,1}^2 := \|\hat h_n - h_0\|_2^2 + c'\|\hat h_n\|_\infty^2 \frac{J_n + \log\zeta^{-1}}{n}.
$$
\item when $\alpha=2, J_n\le n$, and $(\hat h_n,\zeta)$ additionally satisfies 
\begin{equation}\label{eq:h-ev-constraint}
\lambda_{max}(\Sigma_x) \vee \lambda_{max}(\Sigma_z) \le 2, ~~
(\|\hat\Phi_n\|_\infty^2 + \|\hat\Psi_n\|_\infty^2) \sqrt{\frac{J_n+\log\zeta^{-1}}{n}}\le c'', ~~
\|\hat\Psi_n\|_\infty^2 \lambda_{min}^{-1}(\Sigma_x) \sqrt{\frac{J_n+\log\zeta^{-1}}{n}}\le c'',
\end{equation}
with $\Dcov$-probability greater than $1-4\zeta$, 
$\tilde T$ will satisfy \eqref{eq:T-cond-relaxed} with $$
c=2, ~~ \epsilon_n\le 2\epsilon_{n,1} + c'(
    \|\hat\Psi_n\|_\infty^2 + \|\hat\Phi_n\|_\infty^2
)\sqrt{\frac{J_n+\log\zeta^{-1}}{n}}.
$$ 
\end{enumerate}
Moreover, in both cases, \eqref{eq:lem-approx-hs-claim}, \eqref{eq:lem-approx-hs-symm-claim} and \eqref{eq:thm-bip-claim} will hold with the above $\epsilon_n$. 
\end{proposition}

\begin{remark}\label{rem:est-err}
Comparing with $k_{x,p}$ which satisfies \cref{lem:loss-equivalence} with $\epsilon_n^2 = \|\hat h_n - h_0\|_2^2$, we find {\em the sampling error in $k_x$ introduced an overhead of $\cO(J_n/n)$}. When $J_n$ is chosen appropriately, this term will have a negligible order,  
as exemplified in \cref{prop:srl-running-example} below. As we can see from the proof, the $J_n/n$ term may also be replaced by some intrinsic complexity conditions for $k_x$ (e.g., metric entropy bounds, which imply the same critical radius bounds, \citealp{wainwright2019high}); this is also the case for all occurrences of $J_n$ in the following, but we do not pursue this route for simplicity. 
\end{remark}

\begin{proposition}[generic approximation bound]\label{prop:approx-general}
Let the reproducing kernel $k_x^h, k_z^h$ be defined by the integral operator $(E^\top E)^{1/2}, (EE^\top)^{1/2}$. Suppose $k_x^h, k_z^h$ satisfy Assumption 4.1 in \citet{wang2022fast}. Define the truncated Gaussian process 
$
f_{x}^{(J_n)} := \sum_{j=1}^{\lceil J_n/2 \rceil}  \epsilon_j \sqrt{\bar s_j} \gtsv{j}, ~ \text{where } \epsilon_j \sim N(0, 1),
$ 
where we recall $E = \sum_{i=1}^\infty\bar s_i\bar\varphi_i\otimes\bar\psi_i$ denotes its SVD. 
Suppose the hypothesis space $\cF_{x,n}$ satisfy
$$
\EE_{f_{x}^{(J_n)}} \inf_{\tilde f\in\cF_{x,n}} \|f_x^{(J_n)}-\tilde f\|_2^2 \le \epsilon_{GP,n}^2 \le 1.
$$
Suppose hypothesis space $\cF_{z,n}\subset L_2(P_z)$ satisfy a similar bound, defined by the GP 
$
g_{z}^{(J_n)} := \sum_{j=1}^{\lceil J_n/2 \rceil}  \epsilon'_j \sqrt{\bar s_j} \bar\varphi_{j},
$ and
the same $\epsilon_{GP,n}$. 
Then there exists some universal $c_F>0$, s.t.~the hypothesis space 
\begin{align*}
\cF_{h,n} := \Bigg\{h(z,x) = \sum_{i=1}^{J_n}  \sum_{j=1}^{J_n} a_{ij} g_i(z) f_j(x):~& g_i\in\cF_{z,n},f_j\in\cF_{x,n}, A:=(a_{ij})=B^\top C,\\
& \text{ where } B, C \in \RR^{\lceil J_n/2\rceil\times J_n}, \|B\|+\|C\| \le C_F J_n^{-1/2}
    \Bigg\} \numberthis\label{eq:combined-space}
\end{align*}
satisfies the approximation bound 
$$
\inf_{h\in\cF_{h,n}} \|h - h_0\|_2^2 \lesssim \epsilon_{GP,n}^2 + \sum_{j\ge \lceil J_n/2\rceil} \bar s_j^2.
$$
\end{proposition}

\begin{remark}\label{rmk:Phi-h-defn}
For the above $\cF_{h,n}$, any $h\in\cF_{h,n}$ can be written in the form of $
h(z,x) = (B_h G_h(z))^\top (C_h F_h(x))$. We introduce the notations 
$
\Psi_h(x) := C_h F_h(x), \Phi_h(z) := B_h G_h(z), \hat\Psi_n := \Psi_{\hat h_n}, \hat\Phi_n := \Phi_{\hat h_n}.
$
\end{remark}

\begin{proposition}[spectral decomposition, compositional H\"older model]\label{prop:srl-running-example}
Let $\bx,\bz,f_{enc,x},f_{enc,z}$ be defined as in Example~\ref{ex:main}, \ref{ex:main-torus}. Suppose $\mrm{dim}\;\bx=\mrm{dim}\;\bz=d_{obs}$, and $f_{enc,x},f_{enc,z}$ are $\beta_d$ H\"older regular. Let $\cF_{x,n}, \cF_{z,n}$ be specified as DNN models in \citet{schmidt-hieber_nonparametric_2020}, with hyperparameters chosen appropriately (see proof for details). Let $\cF_{h,n}$ be defined as in \cref{prop:approx-general}. Then with probability $\ge 1-n^{-10}$, 
the estimate $\hat h_n$ in \cref{prop:dr-est} satisfies 
\begin{equation}\label{eq:srl-ex-claim-1}
\|\hat h_n\|_\infty^2 = \cO(\log^2 n), ~~
\|\hat h_n - h_0\|_2^2 = \tilde\cO\Big(
 J_n^{-(2p-1)} + \frac{J_n^2}{n} + 
J_n \big(
    n^{-\frac{p-2}{p-1}} + 
    n^{-\frac{2\beta_d}{2\beta_d+d_{obs}}} \big)
\Big).
\end{equation}
Consequently, when $\alpha=1, \beta_l \ge (p-2)d_{obs}/2, J_n \asymp n^{(p-2)/(2p+1)(p-1)}$, we have,  
with probability $\ge 1-n^{-11}$ w.r.t.~the samples $(\Dh, \Dcov)$, 
the RKHS corresponding to the kernel \eqref{eq:kernel-used} will satisfy \eqref{eq:T-cond-relaxed}, with $c=1$ and 
\begin{equation}\label{eq:eps-bound-main-ex}
\epsilon_n^2 = \tilde\cO\big(
n^{-\frac{(p-2)(2p-1)}{(p-1)(2p+1)}} 
\big).
\end{equation}
\end{proposition}

\begin{remark}
In the above, we have only assumed %
$f_{enc,x}, f_{enc,z}$ to have the same regularity for notational simplicity; the result can be trivially extended.  
Given \eqref{eq:eps-bound-main-ex}, 
\eqref{eq:lem-approx-hs-claim}, \eqref{eq:lem-approx-hs-symm-claim} and \eqref{eq:thm-bip-claim} follow immediately by \cref{lem:approx-HS-full} and \cref{thm:bounded-ill-posedness}.\footnote{Recall the proof for \cref{thm:bounded-ill-posedness} only requires \eqref{eq:T-cond-relaxed}.} 
\end{remark}

\begin{corollary}\label{cor:srl-running-ex-alpha-2}
Let $\bz,\bx,f_{enc,x},f_{enc,z},\beta_d,\cF_{h,n}$ be defined as in 
 \cref{prop:srl-running-example}. 
Let $A_n=\tilde\cO(1)$ be some slowly increasing sequence to be specified in the proof, and  
$\hat h_n(z,x) = \hat\Phi_n^\top(z)\hat\Psi_n(x)$ be the solution to the following constrained optimization problem:
\begin{equation}\label{eq:constrained-opt-problem}
\min_{h\in\cF_{h,n}} \cD_n(h), \quad \text{s.t.}~~ \lambda_{min}(\hat\Sigma_{x,h}) \ge A_n n^{-\frac{p(p-2)}{(p-1)(2p+1)}}, ~
\lambda_{max}(\hat\Sigma_{x,h}) \le \frac{6}{5}, ~
\lambda_{max}(\hat\Sigma_{z,h}) \le \frac{6}{5}.
\end{equation}
In the above, $\hat\Sigma_{x,h} = \frac{1}{n}\sum_{i=1}^n \Psi_n(\tilde x_i)\Psi_n^\top(\tilde x_i)$,
$\hat\Sigma_{z,h} = \frac{1}{n}\sum_{i=1}^n \Phi_n(\tilde z_i)\Phi_n^\top(\tilde z_i)$ are the empirical covariance matrices, and $(\hat\Sigma_{x,h}, \hat\Sigma_{z,h}, \cD_n)$ are all
using the samples $\Dh = \{(\tilde z_i, \tilde x_i)\}_{i=1}^n$. Then $\hat h_n$ continues to satisfy \eqref{eq:srl-ex-claim-1}. 
Moreover, suppose $\beta_l\ge (p-2)d_{obs}/2, \alpha=2, J_n\asymp A_n^{-1/p} n^{(p-2)/(2p+1)(p-1)}$. 
Then on a $(\Dh, \Dcov)$-measurable event w.p.~$\ge 1-3n^{-11}$, the RKHS $\cH$ corresponding to the kernel \eqref{eq:kernel-used} will satisfy \eqref{eq:lem-approx-hs-claim}, \eqref{eq:lem-approx-hs-symm-claim} and \eqref{eq:thm-bip-claim}, with 
$$ 
\epsilon_n^2 
= \tilde\cO\left(
    n^{-\frac{(p-2)(2p-1)}{(p-1)(2p+1)}}
\right).
$$
\end{corollary}

We now prove them in turn.

\subsection{Proof for \cref{prop:kern-est}}

We first establish \eqref{eq:T-cond-relaxed} for the two choices of $\alpha$. 

\underline{The $\alpha=1$ case}: define the integral operators $T_{x,p} f := \int f(x) k_{x,p}(x,\cdot) dP_x(x)$, $T_{x} f := \int f(x) k_{x}(x,\cdot) dP_x(x)$. As $\alpha=1$, we have, for all $f\in L_2(P_x)$, 
\begin{align*}
\<f, T_x f\>_2 &= \int f(x') \left(\int\hat k(x,x') f(x) dP_x(x)\right) dP_x(x') \\&
= \int f(x') \Psi(x')^\top \hat\Sigma_z\underbrace{\left(\int \Psi(x) f(x) dP_x(x)\right)}_{=: \mu_f} dP_x(x') \\ & 
=:  \mu_f^\top \hat\Sigma_z \mu_f
= \|\mu_f^\top \Psi(\cdot)\|_{n,2}^2, \\ 
\<f, T_{x,p} f\>_2 &= \|\mu_f^\top \Psi(\cdot)\|_{2}^2,
\end{align*}
where $\|g\|_{n,2} := \sqrt{\frac{1}{n}\sum_{i=1}^n g(z_i)^2}$ denotes the empirical $L_2$ norm defined w.r.t.~the $n$ samples used to estimate $\hat\Sigma_z$. 
Therefore, it suffices to show that
\begin{equation}\label{eq:emp-rkhs-goal}
\PP\Bigg(\sup_{\|f\|_2=1} \big|\|\mu_f^\top \Psi(\cdot)\|_{n,2} - \|\mu_f^\top \Psi(\cdot)\|_2\big|\ 
\le c \|\hat h_n\|_\infty \sqrt{\frac{J_n + \log \zeta^{-1}}{n}}
\Bigg) \ge 1 - \zeta,
\end{equation}
after which we will have, on the above event,  for all $\|f\|_2 = 1$, 
\begin{align*}
|\|\tilde T^{1/2} f\|_2 - \|T^{1/2} f\|_2| &\le 
|\|T_x^{1/2} f\|_2 - \|T_{x,p}^{1/2} f\|_2| + 
|\|T_{x,p}^{1/2} f\|_2 - \|T^{1/2} f\|_2|  \\ 
&\le c\|\hat h_n\|_\infty\sqrt{\frac{J_n + \log \zeta^{-1}}{n}} + 
|\|T_{x,p}^{1/2} f\|_2 - \|T^{1/2} f\|_2|  \\ 
&\overset{(i)}{\le} c\|\hat h_n\|_\infty\sqrt{\frac{J_n + \log \zeta^{-1}}{n}} + \|E - \tilde E\|_{\mathrm{HS}} \\ 
&\overset{\eqref{eq:loss-equiv-1}}{=} c\|\hat h_n\|_\infty\sqrt{\frac{J_n + \log \zeta^{-1}}{n}} + \|\hat h_n - h_0\|_2.
\end{align*}
In the above, (i) can be found in \cref{app:proof-lem-approx-hs}.  

\eqref{eq:emp-rkhs-goal} follows from a standard localization argument \citep[see e.g.,][Theorem 14.1]{wainwright2019high}: the function space $\cG_e := \{\mu_f^\top \Psi(\cdot): \|f\|_2=1\}$ is $d$-dimensional, and has sup norm bounded by 
$$
\sup_{g\in\cG_e} \|g\|_\infty = \sup_{\|f\|_2=1} \sup_{z\in\cZ} \int \Phi(z)^\top\Psi(x) f(x) dP_x(x)
\le \sup_{\|f\|_2=1} \sup_{z\in\cZ} \|h(z,\cdot)\|_{L_2(P_x)} \|f\|_{L_2(P_x)} \le \|h\|_\infty.
$$
Thus, by \citet[Corollary 14.5]{wainwright2019high}, the critical radius in their Theorem 14.1 can be chosen as $\delta_n^2 = c' \|h\|_\infty^2 \frac{J_n + \log \zeta^{-1}}{n}$, which provides the desired probability.

\underline{The $\alpha=2$ case:} In this case, we have 
\begin{align*}
T_x &= \Psi^\top 
    \hat\Sigma_x^{-1/2} (\hat\Sigma_x^{1/2}\hat\Sigma_z\hat\Sigma_x^{1/2})^\alpha\hat\Sigma_x^{-1/2}
    \Psi^\top \\ 
&= (\Sigma_x^{-1/2}\Psi)^\top 
   \Sigma_x^{1/2}\hat\Sigma_x^{-1/2} 
    (\underbrace{\hat\Sigma_x^{1/2}\hat\Sigma_z\hat\Sigma_x^{1/2}}_{=: S})^\alpha 
    \underbrace{\hat\Sigma_x^{-1/2}\Sigma_x^{1/2}}_{=: B}
    (\underbrace{\Sigma_x^{-1/2}\Psi}_{=: \Psi'}), \\ 
\tilde T = T_x^{1/\alpha} &= \Psi'^\top(B^\top S^\alpha  B)^{1/\alpha}\Psi',
\end{align*}
where the last equality follows as $\Psi' \Psi'^\top = I$. 
By the result for $\alpha=1$, 
we know that w.p.~$\ge 1-\zeta$, $\tilde T' = \Psi'^\top B^\top S B \Psi'$ will satisfy \eqref{eq:T-cond-relaxed} with $c=1,\epsilon_n=\epsilon_{n,1}$. %
Thus, to prove $\tilde T$ satisfy \eqref{eq:T-cond-relaxed} with the claimed constants, it suffices to show that on that event, we have, w.p.~$\ge 1-\zeta$,
\begin{equation}\label{eq:a2-goal}
\frac{1}{4}\tilde T' \prec \tilde T \prec 4\tilde T'.
\end{equation}
By the operator monotonicity of the function $(\cdot)^{\frac{1}{2}}$ \citep[Theorem V.1.9]{bhatia2013matrix}, \eqref{eq:a2-goal} is implied by 
$
\frac{1}{16} \tilde T'^2 \prec \tilde T^2 \prec 16\tilde T'^2;
$
since $\Psi'\Psi'^\top = I$, the latter condition is equivalent to 
$
\frac{1}{16} B^\top SBB^\top SB  \prec B^\top S^2 B \prec 16 B^\top SBB^\top SB,
$
which in turn is implied by %
\begin{equation}\tag{\ref{eq:a2-goal}'}\label{eq:a2-goal-t}
\frac{1}{16} I \prec \Sigma_x^{-1/2}\hat\Sigma_x  \Sigma_x^{-1/2} \prec 16I.
\end{equation}
It remains to establish \eqref{eq:a2-goal-t}. 
Observe that for $\bx\sim P_x$, 
The vector $\Sigma_x^{-1/2}\Psi_n(\bx)$ is $\sigma_x^2$-subgaussian with
$
\sigma_x^2 \lesssim \sup_{x\in\cX}\|\hat\Psi_n(x)\|_2^2 \lambda_{min}^{-1}(\Sigma_x).
$
Thus, by concentration of sub-gaussian matrices \citep[Theorem 6.5]{wainwright2019high}, there exists some universal $c>1$, s.t.~\eqref{eq:a2-goal-t} will hold w.p.~$\ge 1-\zeta$ when %
$$
(\sup_{x\in\cX}\|\hat\Psi_n(x)\|_2)^2 s_{min}^{-1}(\Sigma_x) \sqrt{\frac{J_n+\log\zeta^{-1}}{n}} \le c.
$$
This proves \eqref{eq:T-cond-relaxed}. 

\underline{It remains to establish \eqref{eq:lem-approx-hs-claim}, \eqref{eq:lem-approx-hs-symm-claim} and \eqref{eq:thm-bip-claim}.} 
As noted in its proof, \eqref{eq:thm-bip-claim} always holds under the relaxed condition \eqref{eq:T-cond-relaxed}. 
For \eqref{eq:lem-approx-hs-claim}, the $\alpha=1$ case follows by \cref{lem:approx-HS-full}~\ref{it:lem-second-cond}. When $\alpha=2$, by \citet[Theorem 6.5]{wainwright2019high} we have, w.p.~$\ge 1-2\zeta$, 
$$
\|\hat\Sigma_z - \Sigma_z\| + \|\hat\Sigma_x - \Sigma_x\| \le 
c'''(\sup_{x\in\cX} \|\hat\Psi_n(x)\|_2^2 + \sup_{z\in\cZ} \|\hat\Phi_n(z)\|_2^2)\sqrt{\frac{J_n+\log\zeta^{-1}}{n}} \overset{\eqref{eq:h-ev-constraint}}\le c'''c'',
$$
for some universal $c'''>0$. 
Thus, \eqref{eq:lem-approx-hs-claim} is proved by plugging into \cref{lem:approx-HS-full}~\ref{it:lem-generic-cond} the high-probability bound 
\begin{align*}
\|T^{\frac{\alpha}{2}} - \tilde T^{\frac{\alpha}{2}}\| &= 
    \|\Psi^\top(\Sigma_z\Sigma_x\Sigma_z - \hat\Sigma_z \hat\Sigma_x\hat\Sigma_z)\Psi\| \\ 
&\lesssim \|\Psi\|^2(\|\hat\Sigma_z-\Sigma_z\|(\|\Sigma_x\|\|\Sigma_z\|+1) + (\|\Sigma_z\|^2+1)\|\hat\Sigma_x-\Sigma_x\|) \\ 
&\overset{\eqref{eq:h-ev-constraint}}{\lesssim} \|\hat\Sigma_z-\Sigma_z\| + \|\hat\Sigma_x - \Sigma_x\| \\ 
&\lesssim (\sup_{x\in\cX} \|\hat\Psi_n(x)\|_2^2 + \sup_{z\in\cZ} \|\hat\Phi_n(z)\|_2^2)\sqrt{\frac{J_n+\log\zeta^{-1}}{n}}.
\end{align*} 
(All hidden constants above are universal.) \eqref{eq:lem-approx-hs-symm-claim} now follows by \cref{lem:approx-HS-full} \ref{it:lem-symm-result}, which holds for both $\alpha$. 
This completes the proof.
\qed

\subsection{Proof for \cref{prop:approx-general}}\label{app:proof-prop-approx-general}
Define $J_h := \lceil J_n/2 \rceil$, 
$$
\bar\Phi: L_2(P_z)\to \RR^{J_h}, g\mapsto \begin{bmatrix}
    \<g, \sqrt{\bar s_1}\bar\varphi_1\>_2\\
    \ldots \\
    \<g, \sqrt{\bar s_{J_h}}\bar\varphi_{J_h}\>_2
\end{bmatrix}.
$$
As noted in \cref{rmk:addi-conventions}, we can view $\bar\Phi = (\sqrt{\bar s_1}\bar\varphi_1; \ldots; \sqrt{\bar s_{J_h}}\bar\varphi_{J_h})$ as an ``infinitely wide matrix''. 
Define $\bar\Psi: L_2(P_x)\to\RR^{J_h}$ similarly, using $(\sqrt{\bar s_1}\bar\psi_1, \ldots, \sqrt{\bar s_{J_h}}\bar\psi_{J_h})$. 
Recall that by \cref{app:reproof-loss-equivalence}, we have 
$
h_0(z, x) = \sum_{i=1}^\infty \bar s_i \bar\varphi_i(z) \bar\psi_i(x), 
$ and thus 
\begin{equation}\label{eq:generic-approx-1}
\|\bar\Phi^\top \bar\Psi - h_0\|_2^2 %
= \sum_{j>{J_h}} \bar s_j^2.
\end{equation}
We now construct a probabilistic approximation to $\bar\Phi^\top\bar\Psi$. Let 
$\{g^{(j)}: j\in [J_n]\}$ be $J_n$ i.i.d.~copies of $g^{(J_n)}_z$, so that 
we have 
\begin{align*}
\bar G := \begin{bmatrix} g^{(1)} \\ \ldots \\ g^{(J_n)}\end{bmatrix} = \begin{bmatrix}
    \epsilon_{z,11} & \ldots & \epsilon_{z,1,J_h} \\ 
    \ldots & \ldots & \ldots \\ 
    \epsilon_{z,J_n,1} & \ldots & \epsilon_{z,J_n,J_h} 
\end{bmatrix} \begin{bmatrix}
    \sqrt{\bar s_1} \gtsv{1} \\ 
    \ldots \\ 
    \sqrt{\bar s_{J_h}} \gtsv{J_h} 
\end{bmatrix}
=: \Xi \bar\Phi,
\end{align*}
where $\{\epsilon_{z,i,j}: i\in [J_n], j\in [J_h]\}$ are i.i.d.~standard normal variables. 
Define 
$$
\tilde G(z) := (\tilde g^{(1)}(z), \ldots, \tilde g^{(J_n)}(z)),
$$
where $\tilde g^{(j)}\in\cF_{z,n}$ is the best approximation to $g^{(j)}$. Then by the assumed condition on $\cF_{z,n}$ we have 
$$
\|\tilde G - \bar G\|_{2,2}^2 = \EE \sum_{j=1}^{J_n} \|\tilde g^{(j)} - g^{(j)}\|_2^2 
\le J \epsilon_{GP,n}^2.
$$
and by Markov's inequality,
\begin{equation}\label{eq:markov-approx}
\PP_{\Xi}(\|\tilde G - \bar G\|_{2,2}^2 \le 9 J\epsilon_{GP,n}^2) \ge 1 - \frac{1}{9}.
\end{equation}
We also have \citep[Eq.~17]{wang2022fast}, with high probability, 
\begin{equation}\label{eq:random-mat}
\|(\Xi^\top \Xi)^{-1}\Xi^\top\| \le c J_n^{-1/2},
\end{equation}
where $c>0$ is a universal constant. Let us assume $J_n \ge c'$, where the constant $c'$ is chosen so that the above event has probability $\ge 8/9$. 
Define 
$
\hat\Phi := (\Xi^\top \Xi)^{-1} \Xi^\top \tilde G.
$ Then, 
on the intersection of the above two events, we have 
\begin{align*}
\|\hat\Phi - \Phi\|_{2,2} &= \|(\Xi^\top \Xi)^{-1} \Xi^\top (\tilde G - \bar G)\|_{2,2} 
\le  \|(\Xi^\top \Xi)^{-1} \Xi^\top\|\|\tilde G - \bar G\|_{2,2} 
\le 3c \epsilon_{GP,n}.
\end{align*}
As a similar condition holds for $\cF_{x,n}$, following the same reasoning we find the existence of some 
$
\Psi := (\Xi'^\top \Xi')^{-1} \Xi'^\top \tilde F,
$
where $\tilde F = (\tilde f^{(1)},\ldots, \tilde f^{(J_n)})$ and $\tilde f^{(j)} \in \cF_{x,n}$, 
such that 
$$
\PP_{\Xi'}(\|\hat\Psi - \Psi\|_{2,2} \le 3c \epsilon_{GP,n}) \ge \frac{7}{9}.
$$
Combining, we have 
$$
\PP_{\Xi,\Xi'}(
\max\{\|\hat\Phi - \Phi\|_{2,2}, 
    \|\hat\Psi - \Psi\|_{2,2}\} \le 3c \epsilon_{GP,n}) \ge \frac{5}{9}.
$$
Thus, there {\em exists} some $(\hat\Phi,\hat\Psi)$ so that the above inequalities hold true. And we have,
\begin{align*}
\|\hat\Phi^\top \hat\Psi - \Phi^\top\Psi\|_2^2 &\le
    2\int \int (\hat\Phi(z)^\top (\hat\Psi-\Psi)(x))^2 dP_z dP_x + 
    2\int \int ((\hat\Phi-\Phi)(z)^\top \Psi(x))^2 dP_z dP_x \\ 
    &= 2(\|\hat\Phi\|_{2,2}^2 \|\hat\Psi-\Psi\|_{2,2}^2 + \|\hat\Phi-\Phi\|_{2,2}^2\|\Psi\|_{2,2}^2) \\ 
    &\le 2((\|\Phi\|_{2,2}+\|\Phi-\hat\Phi\|_{2,2})^2 \|\hat\Psi-\Psi\|_{2,2}^2 + \|\hat\Phi-\Phi\|_{2,2}^2\|\Psi\|_{2,2}^2) \\ 
    &\lesssim \epsilon_{GP,n}^2.
\end{align*}
Combining the above with \eqref{eq:generic-approx-1}, we have 
$$
\|\hat\Phi^\top\hat\Psi - h_0\|_2^2 \lesssim 
\epsilon_{GP,n}^2 + \sum_{j\ge J_h} \bar s_j^2.
$$
It remains to verify that 
\begin{equation}\label{eq:h-approx-defn}
 \hat\Phi^\top\hat\Psi =
\tilde G^\top (\underbrace{((\Xi^\top \Xi)^{-1}\Xi^\top)}_{=:B})^\top \underbrace{(\Xi'^\top \Xi')^{-1}\Xi'^\top}_{=: C} \tilde F 
\in\cF_{h,n}.
\end{equation}
This follows from \eqref{eq:random-mat} and the similar condition on $\Xi'$, and because 
we always have 
$
\tilde g^{(i)}\in\cF_{z,n}, \tilde f^{(i)}\in \cF_{x,n}.
$
This completes the proof.\qed

\begin{remark}
We note that for the matrices $B,C$ defined above, $A=B^\top C$ satisfies 
$$
\|A\|_F^2 = Tr(A^\top A) \le J_n \|A^\top A\| \le J_n \|B\|^2\|C\|^2 \le c^2 J_n^{-1},
$$
where the constant $c>0$ is universal. Thus, we have 
\begin{equation}\label{eq:slightly-larger-space}
\cF_{h,n} \subset \cF_{h,n}' := \{h(z,x)  = \sum_{ij} a_{ij} g_i(z) f_j(x): g_i\in\cF_{z,n}, f_j\in\cF_{x,n}, A:=(a_{ij})\text{ satisfies }\|A\|_F \le c^2 J_n^{-1}\}.
\end{equation}
\end{remark}

\subsection{Proof for \cref{prop:srl-running-example}}\label{app:proof-srl-running-example}

We shall construct approximate models for the truncated Gaussian processes in \cref{prop:approx-general}, and invoke \cref{prop:est-general}, also using the covering number bound of \cref{lem:covering}. 

\newcommand{\tGPz}{g^{(J_h)}_z}
\newcommand{\tGPl}{g^{(J_h)}_l}

First, consider the truncated Gaussian process 
$
\tGPz = \sum_{i=1}^{J_h} \sqrt{\epsilon_i \bar s_i} \bar\varphi_i.
$ %
By Claim~\ref{claim:l-sv} and \ref{claim:ET}, we have $\bar\varphi_i = \bar\varphi_i^l\circ f_{enc,z}$, where $\{\bar\varphi_i^l\}$ are the left singular vectors of the operator 
$$
\bar E: L_2(P(d\bar x)) \to L_2(P(d\bar z)), f_l \mapsto \EE(f_l(\bar\bx)\mid\bar\bz).
$$
Thus, we have 
$
\tGPz = \tGPl \circ f_{enc,z},
$
where $\tGPl(\bar z) = \sum_{i=1}^{J_h} \sqrt{\epsilon_i \bar s_i} \bar\varphi_{i}^l(\bar z)$ defines a ``latent space GP''. 
\newcommand{\latentI}[1]{\bar\cI_l^{#1}}
We repeat the argument in Ex.~\ref{ex:main-torus} to characterize its regularity. 
Introduce the latent function spaces $$
\latentI{\gamma} := \Ran((\bar E\bar E^\top)^{\gamma/2}),~ \|g\|_{\latentI{\gamma}} := \|(\bar E\bar E^\top)^{-\gamma/2} g\|_2,
$$
where $\gamma>0$ is any real number. Recall $\bar s_i\asymp i^{-p}$ (Ex.~\ref{ex:main-torus}). Thus, the eigenvalues of the operator $(\bar E\bar E^\top)^{\gamma}$ decays at $e_i(\bar E\bar E^\top)^{\gamma} \asymp i^{-2p\cdot \gamma}$. As noted in \citet{chen_rate_2007}, for $(\bar\bx,\bar\bz)$ as defined in Ex.~\ref{ex:main-torus}, both the left and right singular vectors of $\bar E$ are the Fourier basis functions, and 
$\latentI{\gamma}$ is norm-equivalent to the periodic Sobolev space of order $\gamma p d$. 
Thus, {\em for any} $\epsilon>0$, the natural inclusion operator from $\latentI{1/2}$ to $\latentI{(p-1-\epsilon)/2p}$ is Hilbert-Schmidt, and by \citet{steinwart_convergence_2019}, there exists a version of $g_l^{(\infty)} := \sum_{i=1}^\infty \sqrt{\epsilon_i\bar s_i}\bar\varphi_i$ taking value in $\latentI{(p-1-\epsilon)/2p} = W_{per}^{\frac{(p-1-\epsilon)d}{2}}$, and we have
$$
\EE_{g_l^{(J_h)}} \|g_l^{(J_h)}\|_{\latentI{(p-1-\epsilon)/2p}}^2 \le 
\EE_{g_l^{(\infty)}} \|g_l^{(\infty)}\|_{\latentI{(p-1-\epsilon)/2p}}^2 < \infty,
$$
where the first equality holds pointwise, by definition of the RKHS norm. 
Now, by the Sobolev embedding theorem, we have
$$
\EE_{g_l^{(J_h)}} \|g_l^{(J_h)}\|_{\cC^{\frac{(p-2-\epsilon)d}{2}}}^2 \le 
\EE_{g_l^{(\infty)}} \|g_l^{(\infty)}\|_{\latentI{(p-1-\epsilon)/2p}}^2 < \infty,
$$
where for any $r>0$, $\cC^{r}$ denotes the H\"older space of order $r$ \citep{gine2021mathematical}. Note the above bound 
holds uniformly across all $J_h\in\mb{N}$. 

Now we invoke the approximation result in \citet{schmidt-hieber_nonparametric_2020}: by invoking their Lemma 5 as in their Theorem 1, we find that, for any constant $C>0$, there exists a ReLU DNN model with 
\begin{equation}\label{eq:entropy-naive}
\log N(\cF_{z,n}, \|\cdot\|_\infty, \delta) \lesssim 
\big(n^{\frac{1}{p-1-\epsilon}} + n^{\frac{d_{obs}}{2\beta_d+d_{obs}}}\big)(\log^2 n+\log \delta^{-1}),
\end{equation}
and some constant $C'>0$, 
s.t.~for all $g_l\in \cC^{(p-2-\epsilon)d}$ with 
$\|g_l\|_{\cC^{(p-2-\epsilon)d}} \le C\log n$, 
we have 
$$
\inf_{\tilde g\in\cF_{z,n}} \|\tilde g - g_l\circ f_{enc,z}\|_2^2 \le C'\log^3 n\Big(
    n^{-\frac{p-2-\epsilon}{p-1-\epsilon}} + 
    n^{-\frac{2\beta_d}{2\beta_d+d_{obs}}}
\Big) \|g\|_{\cC^{(p-2-\epsilon)d}}.
$$
By their construction, $\cF_{z,n}$ also satisfies the sup norm bound 
\begin{equation}\label{eq:Bn-bound}
    B_n := \sup_{g\in\cF_{z,n}}\|g\|_\infty \lesssim \log n. 
\end{equation}
Invoking the Borell-TIS inequality as in \citet[Appendix C.5.1]{wang2022fast}, we find 
$$
\EE_{\tGPz} \inf_{\tilde g \in\cF_{z,n}}\|\tilde g - \tGPz\|_2^2 \le C'\log^4 n\Big(
    n^{-\frac{p-2-\epsilon}{p-1-\epsilon}} + 
    n^{-\frac{2\beta_d}{2\beta_d+d_{obs}}}
\Big).
$$
Following the same reasoning, we can establish the same approximation and covering bounds for $(\cF_{x,n},f_x^{(J_n)})$. Thus, we conclude that 
\begin{equation}\label{eq:eps-gp-bound-inst}
\epsilon_{GP,n}^2 \le 
C'\log^4 n\Big(
    n^{-\frac{p-2-\epsilon}{p-1-\epsilon}} + 
    n^{-\frac{2\beta_d}{2\beta_d+d_{obs}}} \Big),
\end{equation}
and, by \cref{prop:approx-general}, the existence of some $h^*\in\cF_{h,n}$ s.t.
\begin{equation}\label{eq:approx-bound-inst}
\|h^* - h_0\|_2^2 \lesssim 
\log^4 n\Big(
    n^{-\frac{p-2-\epsilon}{p-1-\epsilon}}  + 
    n^{-\frac{2\beta_d}{2\beta_d+d_{obs}}} \Big) + J_n^{-(2p-1)}.
\end{equation}

Plugging \eqref{eq:entropy-naive} (and a similar result for $\cF_{x,n}$) into \cref{lem:covering}, and plugging the result into \cref{prop:est-general}, we find the following holds with probability $\ge 1-\zeta$:
\begin{align*}
\divLB[h_0] - \divLB[\hat h_n] &\lesssim 
    \divLB[h_0] - \divLB[h_*] + B^4\Big(J_n \log^4 n\Big(
    n^{-\frac{p-2-\epsilon}{p-1-\epsilon}}  + 
    n^{-\frac{2\beta_d}{2\beta_d+d_{obs}}} \Big) + \frac{J_n^2}{n}  + \frac{\log\zeta^{-1}}{n}\Big) 
\numberthis\label{eq:oracle-ineq-instantiated-0}.
\end{align*}
To bound $B = \sup_{h\in\cF_{h,n}} \|h\|_\infty$ above, observe that for any $h\in\cF_{h,n}$, we have (recall the notations in \cref{rmk:Phi-h-defn})
\begin{align*}
    \|\Phi_h\|_\infty = \sup_{z\in\cZ} \|B_h G_h(x)\|_2 &\le \|B_h\| \cdot \sqrt{J_n} B_n \overset{\eqref{eq:combined-space}}{\le} C_F B_n, ~~~~
    \|\Psi_h\|_\infty \le C_F B_n, \numberthis\label{eq:Psi-Phi-norm-bound} \\
    \|h\|_\infty &\le \|\Psi_h\|_\infty \|\Phi_h\|_\infty \lesssim B_n^2 
 \overset{\eqref{eq:Bn-bound}}{\lesssim} \log^2 n.
\end{align*}
Plugging \eqref{eq:approx-bound-inst} and the above into \eqref{eq:oracle-ineq-instantiated-0}, we find 
\begin{align*}
\divLB[h_0] - \divLB[\hat h_n] &\lesssim 
\tilde\cO\Big(
 J_n^{-(2p-1)} + \frac{J_n^2}{n} + 
J_n \big(
    n^{-\frac{p-2}{p-1}} + 
    n^{-\frac{2\beta_d}{2\beta_d+d_{obs}}} \big) + \frac{\log\zeta^{-1}}{n}
\Big). 
\numberthis\label{eq:oracle-ineq-instantiated-1} 
\end{align*}
This proves \eqref{eq:srl-ex-claim-1}. 

When $\beta_l \ge \frac{p-2}{2} d_{obs}$, the term $n^{-\frac{2\beta_d}{2\beta_d+d_{obs}}}$ does not dominate. Thus, for $J_n \asymp n^{\frac{(p-2)}{(p-1)(2p+1)}}$, \eqref{eq:oracle-ineq-instantiated-1} reduces to 
$$
\divLB[h_0] - \divLB[\hat h_n] = \tilde\cO\Big(n^{-\frac{(p-2)(2p-1)}{(p-1)(2p+1)}} + \frac{\log\zeta^{-1}}{n}\Big).
$$
\eqref{eq:eps-bound-main-ex} is proved by invoking \cref{prop:kern-est}~(i), and observing the added terms do not dominate. 
\qed

\subsection{Proof for \cref{cor:srl-running-ex-alpha-2}}

We first show that with high probability \eqref{eq:srl-ex-claim-1} continues to hold for the new $\hat h_n$. 
Denote by $\hat\cF_{h,n}\subset\cF_{h,n}$ the set of functions that satisfy the eigenvalue constraints. 
Then $\hat\cF_{h,n}$ satisfies the same the entropy bound as $\cF_{h,n}$, and thus the oracle inequality \eqref{eq:oracle-ineq-instantiated-0} continues to hold. 
To bound the approximation error in \eqref{eq:oracle-ineq-instantiated-0}, 
let $h^*\in\cF_{h,n}$ be defined as in the proof of \cref{prop:srl-running-example}; 
observe that we can choose $A_n = \tilde\cO(1)$ and $J_n \asymp A_n^{-1/p} n^{(p-2)/(2p+1)(p-1)}$ so that $\hat \cF_{h,n} = 
\hat\cF_{h,n,\frac{2}{3}\bar s_{J_n},\frac{6}{5}}$, where the right hand side is defined in \cref{lem:constrained-opt-err} below. 
Let $E_n$ denote the $\Dh$-measurable event defined in the lemma, with probability $\ge 1-2n^{-11}$.
By its claim~(ii), we have $h^*\in\hat\cF_{h,n}$ on $E_n$. Thus, 
\eqref{eq:oracle-ineq-instantiated-1} and subsequently, \eqref{eq:srl-ex-claim-1}, also continue to hold. 

On the event $E_n$, 
by \cref{lem:constrained-opt-err}~(i)~(b), (iii), \eqref{eq:Psi-Phi-norm-bound} and our choice of $J_n$, 
the optima $\hat h_n$ for \eqref{eq:constrained-opt-problem} will satisfy \eqref{eq:h-ev-constraint} for $\zeta = n^{-11}$. Thus, by \cref{prop:kern-est}~(ii), on the intersection of $E_n$ and a $\Dcov$-measurable event w.p.~$\ge 1-n^{-11}$, 
\eqref{eq:lem-approx-hs-claim}, \eqref{eq:lem-approx-hs-symm-claim} and \eqref{eq:thm-bip-claim} hold with 
$$
\epsilon_n^2 = \tilde\cO\left(J_n^{-(2p-1)} + \frac{J_n^2}{n} + J_n n^{-\frac{p-2}{p-1}} + \cancel{\frac{J_n + \log n}{n}}
\right)
= \tilde\cO\left(
    n^{-\frac{(p-2)(2p-1)}{(p-1)(2p+1)}}
\right).
$$
This completes the proof.\qed

\subsection{Auxiliary Results}

\begin{lemma}[covering number bound]\label{lem:covering}
Let $\cF_{x,n}, \cF_{z,n}$ be function spaces with sup norm bounded by $B_n$. Let $\cF_{h,n}$ be defined as in \cref{prop:approx-general}. Then there exists some universal $c>0$ s.t.~
$$
\log N(\cF_{h,n}, \|\cdot\|_\infty, \delta) \le c J_n(\log (\cF_{x,n}, \|\cdot\|_\infty, B_n^{-1}\delta) + \log (\cF_{z,n}, \|\cdot\|_\infty, B_n^{-1}\delta)) + 
 J_n^2 \log (1+B_n^2\delta^{-1}),
$$
for all $\delta>0$.
\end{lemma}

\begin{proof}
We construct a covering for the slightly larger space $\cF_{h,n}'$ defined in \eqref{eq:slightly-larger-space}. 
Let $\{f^{(i)}_c: i\in [N_{x,\delta}]\}$, $\{g^{(i)}_c: i\in [N_{z,\delta}]\}$ denote a pair of $\delta$-coverings of $\cF_{x,n}, \cF_{z,n}$ in sup norm. Let $\{A^{(i)}_c: i\in [N_{A,\delta}]\}$ denote a $B_n^{-1}J_n^{-1}\delta$-covering of $\{A: \|A\|_F \lesssim J_n^{-1}\}$ in Frobenious norm. %
For any $h =  G^\top A  F\in \cF_{h,n}$, let $\tilde G, \tilde F, \tilde A$ denote the closest element of $ G, F,A$ in the respective covering set. Then we have 
\begin{align*}
\|h - \tilde G^\top\tilde A\tilde F\|_\infty  \le 
\|( G-\tilde G)^\top\tilde A\tilde F\|_\infty + 
\| G^\top(\tilde A-A)\tilde F\|_\infty + 
\| G^\top A(\tilde F -  F)\|_\infty.
\end{align*}
To bound the terms above, we first observe that, for any $F' = (f'_1;\ldots;f'_{J_n}), G'=(g'_1;\ldots;g'_{J_n}), A'\in\RR^{J_n\times J_n}$, we have 
\begin{align*}
\|G'^\top A' F'\|_\infty &= \sup_{x\in\cX,z\in\cZ} G'(z)^\top A' F'(x)
\le \sup_{x\in\cX,z\in\cZ} \|G'(z)\|_2 \|A'\| \|F'(x)\|_2 \\ 
&\le \sqrt{J_n} (\max_{i\in [J_n]} \|g'_i\|_\infty) \cdot \|A'\| \cdot 
\sqrt{J_n} (\max_{i\in [J_n]} \|f'_i\|_\infty). \numberthis\label{eq:sup-norm-bound}
\end{align*}
Plugging back, we have 
\begin{align*}
\|( G-\tilde G)^\top\tilde A\tilde F\|_\infty &\lesssim \sqrt{J_n}\delta \cdot J_n^{-1} \cdot \sqrt{J_n} B_n = B_n\delta. \\ 
\|G^\top A(\tilde F-F)\|_\infty &\lesssim B_n\delta. \\ 
\|G^\top (\tilde A - A) \tilde F\|_\infty &\lesssim \sqrt{J_n}B_n \cdot B_n^{-1}J_n^{-1}\delta \cdot \sqrt{J_n}B_n = B_n\delta.
\end{align*}
Thus, the set $
\{h := \sum_{kl} g_k A_{kl} f_l\}$
where $(g_k, A, f_l)$ are in the respective covering sets defined above, constitute a $\cO(B_n\delta)$ covering for $\cF_{h,n}'$ (the hidden constant is universal). The logarithm of its cardinality is 
$$
\log N_{x,\delta}^{J_n} N_{z,\delta}^{J_n} N_{A,\delta} \le J_n(\log N_{x,\delta} + \log N_{z,\delta})  + J_n^2 \log(1 + 2B_n\delta^{-1}).
$$
The proof completes by a scaling argument.
\end{proof}

\begin{lemma}\label{lem:constrained-opt-err}
Let $\cD^{(n)} = \{(z_i,x_i)\}$ be $n$ i.i.d.~samples from $P_{zx}$. 
For any $h\in\cF_{h,n}$ %
let $(\Psi_h,\Phi_h)$ be defined as in \cref{rmk:Phi-h-defn}, 
$\Sigma_{x,h} = \EE_{\bx\sim P_x} \Psi_h(\bx)\Psi_h^\top(\bx)$, and $\hat\Sigma_{x,h}$ denotes its empirical estimate using $\cD^{(n)}$. Define $(\hat\Sigma_{z,h}, \Sigma_{z,h})$ similarly using $\Phi_h$. 
For any $0<s<t$, define the spaces
\begin{align*}
\cF_{h,n,s,t} &:= \left\{h\in \cF_{h,n}: s \le \lambda_{min}(\Sigma_{x,h}) \le \lambda_{max}(\Sigma_{x,h})\le t, \lambda_{max}(\Sigma_{z,h})\le t\right\}, \\
\hat\cF_{h,n,s,t} &:= \left\{h\in \cF_{h,n}: s \le \lambda_{min}(\hat\Sigma_{x,h}) \le \lambda_{max}(\hat\Sigma_{x,h})\le t, \lambda_{max}(\hat\Sigma_{z,h})\le t\right\}.
\end{align*}
Suppose $p\ge 2,\beta_l\ge \frac{p-2}{2}d_{obs}$. 
Then there exists $C_0,C_1,C_2>0$, and a slowly growing sequence $M_n=\tilde\cO(1)$ s.t.~on a $\cD^{(n)}$-measurable event w.p.~$\ge 1-2 n^{-11}$, 
the following holds for any $n\ge C_0, J_n \le M_n^{-1} n^{\frac{p-2}{(2p+1)(p-1)}}$: 
\begin{enumerate}[leftmargin=*,label=(\roman*)]
\item %
For any $(s,t)$: %
\begin{enumerate}
    \item For all $h\in\cF_{h,n,s,t}$, we have $h\in \hat\cF_{h,n,s-\Delta s_n,t+0.1}$, for 
$
\Delta s_n =  C_1\log^3 n \sqrt{J_n} n^{-\frac{p-2}{p-1-\epsilon}}$.
\item For any $h\in\hat\cF_{h,n,s,t}$, we have $h\in \cF_{h,n,s-\Delta s_n,t+0.1}$.
\end{enumerate}
\item %
For $s= \frac{2}{3}\bar s_{J_n}, t = \frac{11}{10}$, we have %
$h^*\in \cF_{h,n,s,t}\cap \hat\cF_{h,n,s/2,t+0.1}$.
\item For the above $(s,t)$, any $h\in \cF_{h,n,s/2,t+0.1}$ 
will satisfy \eqref{eq:h-ev-constraint}.
\end{enumerate}
\end{lemma}

\begin{proof}We prove the three claims in turn.
\begin{enumerate}[leftmargin=*,label=(\roman*)]
\item
For all $h\in\cF_{h,n}$, recall by \eqref{eq:Psi-Phi-norm-bound} that $\|\Psi_h\|_\infty \vee \|\Phi_h\|_\infty \le C_F \log n$. Thus, by \citet[Theorem 6.5]{wainwright2019high}, 
there exists some $C>0$, s.t.~for all 
$h\in\cF_{h,n}, \delta\in (0,1)$, we have
\begin{equation}\label{eq:prop-E3-intermediate}
\PP_{\cD^{(n)}}\left(\|\Sigma_{x,h} -\hat\Sigma_{x,h}\| \ge C\log^2 n\Big(\sqrt{\frac{J_n}{n}} + \delta\Big)\right) \le e^{-C n\delta^2}.
\end{equation}
We now build a covering for 
$
\cS_{x,n} := \{\hat\Sigma_{x,h}: h\in\cF_{h,n}\}.
$
Using an argument similar to \cref{lem:covering} below, and plugging in the symmetric version of \eqref{eq:entropy-naive} for $\cF_{x,n}$ (see the proof for details), 
we find, for all $\delta'>0$,  
\begin{align*}
\log N(\cS_{x,n}, \|\cdot\|, \delta') &\le 
    c(J_n \log N(\cF_{x,n}, \|\cdot\|_\infty, B_n \delta') 
    + J_n^2\log (1+B_n^2\delta'^{-1}))  \\
&\lesssim J_n (n^{1/(p-1-\epsilon)} + J_n)(\log^2 n + \log \delta'^{-1}).
\end{align*}
Denote by $\mathrm{F}_{h,n,\delta'}\subset\cF_{h,n}$ the function set that induced the $\delta'$-covering of $\cS_{x,n}$, so that its cardinality can be bounded as above. 
Recall that $J_n \le n^{1/(p-1-\epsilon)}$, and 
consider $\delta'\gets n^{-1}, \delta \gets C' n^{-(p-2)/2(p-1-\epsilon)}\sqrt{J_n}\log n$, where $C'>0$ is sufficiently large.
Then a union bound over \eqref{eq:prop-E3-intermediate} shows that
\begin{equation}\label{eq:ev-event}
\PP_{\cD^{(n)}}\Bigl(
    \sup_{\tilde h\in \mathrm{F}_{h,n,\delta'}} \|\Sigma_{x,\tilde h} -\hat\Sigma_{x,\tilde h}\| \le 
    C'\log^3 n \Bigl(
        \cancel{\sqrt{\frac{J_n}{n}}} + \sqrt{J_n} n^{-\frac{p-2}{2(p-1-\epsilon)}}
    \Bigr)
\Bigr) %
\ge 1-e^{-(C'/2)n^{1/(p-1-\epsilon)}J_n\log^2 n} \ge 1-n^{-11}.
\end{equation}
For any $(s,t)$ and $h\in\cF_{h,n,s,t}$, let $\tilde h\in\mathrm{F}_{h,n,\delta'}\cap\cF_{h,n,s,t}$ minimize %
$\|\hat\Sigma_{x,\tilde h} - \hat\Sigma_{x,h}\|$. Then, on \eqref{eq:ev-event}, we have 
\begin{align*}
\lambda_{min}(\hat\Sigma_{x,h}) &= \lambda_{min}(\Sigma_{x,\tilde h} - \Sigma_{x,\tilde h} + \hat\Sigma_{x,\tilde h} - \hat\Sigma_{x,\tilde h} + \hat\Sigma_{x,h})
\ge  
    \lambda_{min}(\Sigma_{x,\tilde h}) - (\|\Sigma_{x,\tilde h} - \hat\Sigma_{x,\tilde h}\| + 
    \|\hat\Sigma_{x,\tilde h} - \hat\Sigma_{x,h}\|)  \\
&\ge s - C'\log^3 n\sqrt{J_n} n^{-\frac{p-2}{2(p-1-\epsilon)}} \cancel{-\frac{1}{n}}. \\ 
\lambda_{max}(\hat\Sigma_{x,h}) &\le 
\lambda_{max}(\Sigma_{x,\tilde h}) + (\|\hat\Sigma_{x,\tilde h} - \Sigma_{x,\tilde h}\| + \|\hat\Sigma_{x,\tilde h} - \hat\Sigma_{x,h}\|) \le t + \frac{1}{10}.
\end{align*}
.
In the above, the eigenvalue inequalities can be found as \citet[Exercise III.2.5]{bhatia2013matrix}, and the second display holds for $n\ge C_{02}$ sufficiently large. 
Following the same argument we find $\lambda_{max}(\hat\Sigma_{z,h})\le t + \frac{1}{10}$. This proves (a). \\
The proof for (b) follows the same argument, with all occurrences of $\hat\Sigma$ and $\Sigma$ exchanged. We omit it for brevity. %
\item In App.~\ref{app:proof-prop-approx-general} we have established that $h^*(z,x) = \hat\Phi^\top(z)\hat\Psi(x)$, where (recall our convention in \cref{rmk:addi-conventions}) $$
\|\hat\Psi-\Psi\| \le 
\|\hat\Psi-\Psi\|_{\text{HS}} =
\EE\|\hat\Psi(\bx) - \Psi(\bx)\|_2^2 \le 3c\epsilon_{GP,n}, ~~ \text{and}~
s_j(\Psi) = \sqrt{\bar s_j}~~\forall j\in [J_n/2].
$$
Thus we have, for some $C_{01}>0$ and any $n\ge C_{01}$, 
\begin{align*}
\lambda_{max}(\Sigma_{x,h^*}) &\le (\sqrt{\bar s_1} + 3c \epsilon_{GP,n})^2 
= (1 + 3c \epsilon_{GP,n})^2 
\le \frac{11}{10}, \\ 
\lambda_{min}(\Sigma_{x,h^*}) &= s_{min}^2(\hat\Psi^\top) \ge (s_{min}(\Psi) - 3c \epsilon_{GP,n})^2 \ge
\frac{3}{4}\bar s_{J_n} - 27\epsilon_{GP,n}^2 \ge \frac{2}{3} \bar s_{J_n},
\end{align*}
where the last inequality follows by plugging \eqref{eq:eps-gp-bound-inst} and our condition on $\beta_l$. 
As the same bound was established for $\EE \|\hat\Phi(\bz)-\Phi(\bz)\|_2^2$, we can similarly show that 
$\lambda_{max}(\Sigma_{z,h^*}) \le \frac{11}{10}$, and 
$$
h^* \in \cF_{h,n,\frac{2}{3}\bar s_{J_n},\frac{11}{10}}.
$$
By claim (i)~(b) and the fact that $\bar s_{J_n} \gg \Delta s_n$ for our choice of $J_n$, 
we have $h^*\hat\cF_{h,n,\frac{1}{3}\bar s_{J_n}, \frac{6}{5}}$. 
\item 
By \eqref{eq:Psi-Phi-norm-bound}, there exists some $C\in (0,1)$ s.t.~\eqref{eq:h-ev-constraint} holds for such $h$ as long as 
$$
\zeta\ge e^{- C J_n}, ~ \bar s_{J_n}^{-1} \sqrt{J_n/n} \log^2 n  \le C, ~\bar s_{J_n} \ge C^{-1} \epsilon_{GP,n}^2.
$$
By \eqref{eq:eps-gp-bound-inst} and our condition on $\beta_l$, the latter two inequalities will hold when $
J_n \le n^{(p-2)/(p-1)2p}.
$
\end{enumerate}
\end{proof}

%% file: app-cmm-results.tex
\section{Results for Conditional Moment Models}

\subsection{Proof for \cref{thm:cmm}}\label{app:proof-cmm-thm}

We prove the following more general result, which covers 
\cref{thm:cmm} in its claim \ref{it:thm-concrete}.

\begin{theorem}\label{thm:generic-rate-restated}
Suppose there exists some $f_0$ that satisfies \eqref{eq:general-source-cond} with $\beta\ge 1$.  
Let 
\begin{itemize}%
    \item 
$\hat h_n(z,x) = \hat\Phi_n^\top(z)\hat\Psi_n(x)$ be defined using $n$ samples $\Dh$, and suppose $\hat\Phi_n(z)\in\RR^{J_n}$. 
\item $\Dcov$ be an independent set of $n$ i.i.d.~samples, 
$(k_x,\cH)$ be defined by \eqref{eq:kernel-used} using $\hat h_n,\Dcov$, and some $\alpha\in\mb{Z}_+$, and 
$(k_z,\cI)$ be defined by \eqref{eq:kz-used} using $\hat h_n, \Dcov$ and $\alpha'=1$.
\item $\Dkiv$ be a third independent set of $n$ i.i.d.~samples from $P_{zxy}$, and 
$\hat f_n$ be defined by \eqref{eq:minimax-estimator}, using $k_z,k_x,\Dkiv$ and $\nu_n,\lambda_n$ as in \eqref{eq:iv-hps-setting} below. 
\end{itemize}

Then for some constant $C_N>0$ and any $n\ge C_N$: 
\begin{enumerate}[label=(\roman*),leftmargin=*]
\item\label{it:thm-generic}
Suppose $\{\epsilon_n\}, \{\zeta_n\}$ be s.t.~with probability $\ge 1-\zeta_n$, \eqref{eq:lem-approx-hs-claim}, \eqref{eq:lem-approx-hs-symm-claim} and \eqref{eq:thm-bip-claim} hold, and that $\|\hat h_n - h_0\|_2\le \epsilon_n$.
Then we have, w.p.~$\ge 1-\zeta_n-2n^{-11}$, 
\begin{equation}\label{eq:cmm-generic-claim}
\|\hat f_n - f_0\|_2 = \cO\Big(
\Big(\epsilon_n^{\frac{\alpha}{\alpha+1}} + 
    \Big(\frac{
        \|\hat h_n\|_\infty^2 J_n^2 + \log n
    }{n}\Big)^{\frac{\alpha}{2(\alpha+1)}}
\Big)(1+\|f_0\|_{\gtRKHS[\beta]}^{\frac{2\alpha}{\alpha+1}})
\Big). 
\end{equation}
\item\label{it:thm-concrete}
When $\bx,\bz,f_{enc,x},f_{enc,z},p,\beta_d$ are defined as \cref{thm:cmm}, 
$J_n \asymp n^{\frac{(p-2)}{(2p+1)(p-1)}}$, and 
either \emph{(a)} $\alpha=1$, $\hat h_n$ maximizes \eqref{eq:emp-obj-expr}, or \emph{(b)} $\alpha=2$, $\hat h_n$ solves \eqref{eq:constrained-opt-problem}, then we have, w.p.~$\ge 1-n^{-10}$, 
$$
\|\hat f_n - f_0\|_2 = \tilde\cO\big(n^{-\frac{\alpha(p-2)}{(\alpha+1)(4p+1)}}(1+\|f_0\|_{\gtRKHS[\beta]}^{\frac{2\alpha}{\alpha+1}})\big),
$$
where $\tilde\cO$ hides all sub-polynomial factors. 
\end{enumerate}
\end{theorem}

\begin{proof} We first establish \ref{it:thm-generic} in there steps:
\begin{enumerate}[leftmargin=*]
    \item 
We first derive approximation and estimation bounds for $\cH$. 
By \eqref{eq:general-source-cond}, 
on the event w.p.~$\ge 1-\zeta_n$, \eqref{eq:lem-approx-hs-claim} holds for $\bar f = f_0$, and there exists some
\begin{equation}\label{eq:H-approx}
\tilde f_n\in\cH, ~\|\tilde f_n\|_\cH\lesssim \|f_0\|_{\gtRKHS[\alpha]}, ~
\|\tilde f_n - f_0\|_2^2\lesssim \epsilon_n^{2(\alpha\wedge 1)}\|f_0\|_{\gtRKHS[\alpha]}^2. 
\end{equation}
{\em We restrict to the above event throughout the proof.} 
For estimation, we first claim that 
\begin{equation}\label{eq:kx-sup-norm-bound}
    \sup_{x\in\cX} k_x(x,x) \le \|\hat h_n\|_\infty^{2\alpha},~~\forall\alpha\in\mb{Z}_+.
\end{equation}
To prove this, let $\{(z^s_i,x^s_i)\} := \Dcov$. Then 
for $
S_x = \frac{1}{\sqrt{n}}(\hat\Psi_n(x_1^s), \ldots, \hat\Psi_n(x_n^s)),
S_z = \frac{1}{\sqrt{n}}(\hat\Phi_n(z_1^s), \ldots, \hat\Phi_n(z_n^s)),
$ we have $\hat\Sigma_z = S_z S_z^\top, \hat\Sigma_x = S_x S_x^\top$, and
\begin{align*}
\|S_z^\top S_x\| &\le \|S_z^\top S_x\|_F = 
\sqrt{
    n^{-2} \sum_{i=1}^n\sum_{j=1}^n\hat\Phi_n(z_i^s) \hat\Psi_n(x_j^s) 
} \le \|\hat h_n\|_\infty,  \\ 
\|S_z^\top\hat\Psi_n(x)\|_2 &\le \sqrt{n} \|S_z^\top\hat\Psi_n(x)\|_\infty 
= \max_{i\in[n]}\hat\Phi_n(z_i^s)\hat\Psi_n(x) \le \|\hat h_n\|_\infty, \\
k_x(x,x) &= \hat\Psi_n(x)^\top \hat\Sigma_z (\hat\Sigma_x\hat\Sigma_z)^{\alpha-1}\hat\Psi_n(x) 
= \hat\Psi_n^\top(x) S_z (S_z^\top S_x S_x^\top S_z)^{\alpha-1} S_z^\top \hat\Psi_n(x) 
\le \|\hat h_n\|_\infty^{2\alpha}.
\end{align*}
Given \eqref{eq:kx-sup-norm-bound}, the critical radius \citep[Eq.~14.4]{wainwright2019high} of local Rademacher complexity of the unit norm ball $\cH_1$ can now be bounded as \citep[Corollary 14.5]{wainwright2019high}
\begin{equation}\label{eq:H-crit-rad}
\delta_{n,\cH_1}^2 \lesssim \frac{\|\hat h_n\|_\infty^{2\alpha} J_n}{n}. %
\end{equation}
\item\label{it:proof-step2} 
We now derive approximation and estimation bounds for $\cI$. We first remark on the full symmetry of our previous results: as proved in Claim~\ref{claim:ET}, 
$E^\top$ is the symmetric conditional expectation operator from $L_2(P_z)$ to $L_2(P_x)$. 
Thus, all results in \cref{sec:estimability}  and \cref{sec:srl}, as well as \cref{prop:kern-est}, apply to $\cI$, with all occurrences of $z,x$ exchanged, $\gtRKHS[\alpha]$ replaced by the following (series of) RKHS
\begin{equation}\label{eq:gt-I}
\bar\cI_{\alpha} := \Ran(EE^\top)^{\alpha/2}, ~\|g\|_{\bar\cI_\alpha} := \|(EE^\top)^{-\alpha/2} g\|_2,
\end{equation}
and $\hat h_n, h_0$ replaced by 
$$
\hat h_n^*(x,z) := \hat h_n(z,x), ~~h_0^*(x,z) := h_0(z,x) = \frac{dP_{xz}}{d(P_x\otimes P_z)}(z,x).
$$
As $\|\hat h_n^* - h_0^*\|_{L_2(P_x\otimes P_z)} = \|\hat h_n - h_0\|_{L_2(P_z\otimes P_x)} \le \epsilon_n^2$, we can invoke the symmetric version of \cref{prop:kern-est} (with  $\alpha\gets \alpha'=1$), 
showing that 
on an event with $\Dcov$-probability $\ge 1-n^{-11}$,
the kernel $k_z$ has integral operator $T_z =: \tilde T_z^{\alpha'}$, where $\tilde T_z$ satisfy 
\begin{equation}\label{eq:I-ET-approx}
\|E^\top g\|_2 - \epsilon'_n \le 
\|\tilde T_z^{1/2} g\|_2 \le \|E^\top g\|_2 + \epsilon'_n~~~~\text{for all }\|g\|_2=1,~~
\text{where}~\epsilon'_n := \epsilon_n + \cO\Big(\|\hat h_n\|_\infty^2\frac{J_n+\log n}{n}\Big).
\end{equation}
{\em We also condition on the above event in the remainder of the proof.} As \eqref{eq:I-ET-approx} fulfills \eqref{eq:T-cond-relaxed}, 
the symmetric version of \cref{lem:approx-HS-full}~\ref{it:lem-first-cond} implies that 
\begin{equation}\label{eq:quick-equiv-1}
\forall g\in \bar\cI_{1},~~ \exists \tilde g \in \cI ~~s.t.~~
\|\tilde g\|_\cI \lesssim \|g\|_{\bar\cI_{1}}, ~\|\tilde g - g\|_2 \lesssim \epsilon_n'\|g\|_{\bar\cI_{1}}.
\end{equation}
By \eqref{eq:lem-approx-hs-symm-claim}, 
for all $f\in\cH$, there exists %
$$
\bar f\in\gtRKHS[\alpha],  ~
\|\bar f\|_{\gtRKHS[\alpha]} \lesssim \|f\|_{\cH}, ~
\|E(\bar f-f)\|_2 \overset{\eqref{eq:basic-facts}}{\le} 
\|\bar f - f\|_2 \lesssim %
 \epsilon_n'\|f\|_\cH.
$$
By definitions, we have $$
\|E \bar f\|_{\bar\cI_{1}} \le \|E \bar f\|_{\bar\cI_{2}} = \|\bar f\|_{\gtRKHS[1]} \le \|\bar f\|_{\gtRKHS[\alpha]}.
$$
Instantiating 
\eqref{eq:quick-equiv-1} with $g\gets E \bar f \in \bar\cI_{\alpha}$ and combining with the above displays establish the existence of $g_f\in\cI$ s.t.
$$ 
\|g_f\|_\cI \lesssim \|E\bar f\|_{\bar\cI_{1}}
\le %
\|\bar f\|_{\gtRKHS[\alpha]} \lesssim \|f\|_\cH, ~~
\|E f - g_f\|_2 \le \|E(f-\bar f)\|_2 + \|g - g_f\|_2 
\lesssim \epsilon_n'\|f\|_\cH.
$$
Applying the above display to $f\gets f-\tilde f_n$, we find, for all $f\in\cH$, the existence of $g_f'\in\cI$ s.t.
\begin{equation}\label{eq:I-approx}
\|g_f'\|_\cI \lesssim \|f - \tilde f_n\|_\cH, \|g_f' - E(f-f_0)\|_2 \le 
 \|g_f' - E(f-\tilde f_n)\|_2 + \|\cancel{E(}\tilde f_n-f_0\cancel{)}\|_2 \lesssim \epsilon_n'(\|f-\tilde f_n\|_\cH+\|f_0\|_{\gtRKHS[\alpha]}),
\end{equation}
where the last inequality also uses \eqref{eq:H-approx}. This establishes the approximation conditions. For estimation, observe 
\begin{equation}\label{eq:kz-sup-norm}
\sup_{z\in\cZ} k_z(z,z) = \sup_{z\in\cZ} \hat\Phi_n(z)^\top \hat\Sigma_x\hat\Phi_n(z)
= \sup_{z\in\cZ} \frac{1}{n}\sum_{i=1}^n \hat\Phi_n(z)^\top\hat\Psi_n(x_i^s) \hat\Psi_n(x_i^s)^\top\hat\Phi_n(z) 
\le  \|\hat h_n\|_\infty^2.
\end{equation}
Since $k_z$ also has rank $J_n$, the product function space $\cG$ defined in \citet[Eq.~2, see also App.~E.2]{dikkala_minimax_2020} is an RKHS with dimensionality $J_n^2$, and has its critical radius trivially bounded as 
\begin{equation}\label{eq:G-crit-rad}
\delta_{n,\cG}^2 \lesssim \frac{\|\hat h_n\|_\infty^{\cancel{2\vee}2\alpha} J_n^2}{n}. %
\end{equation}
\item\label{it:proof-step3} 
Combining \eqref{eq:H-approx}, \eqref{eq:H-crit-rad}, \eqref{eq:I-approx}, \eqref{eq:G-crit-rad} fullfills the condition of \citet[Theorem 1]{dikkala_minimax_2020}, which now states that for %
\begin{equation}\label{eq:iv-hps-setting}
\nu_n \asymp \lambda_n \asymp \frac{\|\hat h_n\|_\infty^{2\alpha} J_n^2}{n} + \epsilon_n'^2,
\end{equation}
we have, with probability $\ge 1-n^{-11}$ w.r.t.~the samples used to estimate $\hat f_n$, 
$$
\|E(\hat f_n - f_0)\|_2^2 \lesssim 
\Big(
\frac{\|\hat h_n\|_\infty^{2\alpha} J_n^2}{n} + \epsilon_n'^2\Big)(1+\|f_0\|_{\gtRKHS[\beta]}^4) + \frac{\log n}{n} \lesssim 
\Big(\frac{\|\hat h_n\|_\infty^{2\alpha} J_n^2}{n} + \epsilon_n^2\Big)(1+\|f_0\|_{\gtRKHS[\beta]}^4) + \frac{\log n}{n}
=: \tilde\epsilon_n^2.
$$
Now, \cref{thm:bounded-ill-posedness} implies that 
$$
\|\hat f_n - f_0\|_2^2 = \cO\big(
    \tilde\epsilon_n^2 + \tilde\epsilon_n^{2\alpha} + \tilde\epsilon_n^{2\alpha/(\alpha+1)}
\big) = \cO\Big(\Big(\epsilon_n + \Big(\frac{\|\hat h_n\|_\infty^{2\alpha} J_n^2 + \log n}{n}\Big)^{\frac{1}{2}}\Big)(1+\|f_0\|_{\gtRKHS}^2)\Big)^{\frac{2\alpha}{\alpha+1}}.
$$
This proves \eqref{eq:cmm-generic-claim}, which holds with probability $\ge 1 - \zeta_n - 2n^{-11}$.  
\end{enumerate}

It remains to prove the claim \ref{it:thm-concrete}. 
By \cref{prop:srl-running-example} (for $\alpha=1$) or \cref{cor:srl-running-ex-alpha-2} (for $\alpha=2$), on a $(\Dh, \Dcov)$-measurable event w.~p.~$\ge 1-3n^{-11}$, we have 
\eqref{eq:lem-approx-hs-claim}, \eqref{eq:lem-approx-hs-symm-claim} and \eqref{eq:thm-bip-claim} hold, and $$
\epsilon_n^2 = \tilde\cO\bigl(
    n^{-\frac{(p-2)(2p-1)}{(p-1)(2p+1)}}
\bigr), ~~\|\hat h_n\|_\infty = \tilde\cO(1).
$$
The proof completes by plugging the above to \eqref{eq:cmm-generic-claim}, and observing that $(2p-1)(p-2)/(2p+1)(p-1) \le 4(p-2)/(4p+1)$ when $p>2$. %
\end{proof}

\subsection{Additional Discussion}\label{app:disc-cmm-results}

\paragraph{Adaptivity} 
\cref{thm:cmm} quantified the $L_2$ error for conditional moment models under assumptions on the source condition ($\beta \ge 1$), and the singular value spectrum $\{\bar s_i\}$, under the compositional data generating process \cref{ex:main}. While 
we can cast any data generating process into this setup by setting $\bar\bz\gets\bz,\bar\bx\gets\bx$, the situation becomes more interesting when the observed $(\bx,\bz)$ contains additional information, i.e., when we can recover high-dimensional $(\bx_\perp,\bz_\perp)$ using $(\bx,\bz)$. 
As we noted in \cref{app:disc-gtRKHS}, 
in such settings, the singular value spectrum of $E$ coincides with that of a 
latent-space conditional expectation operator $\bar E$, defined by $(\bar\bz,\bar\bx)$; and when $f_0$ satisfies \eqref{eq:general-source-cond}, the best possible value for $\beta$ is also determined solely by the regularity of a latent-space function $\bar f_0$, which defines $f_0$ through 
$$
f_0 = \bar f_0 \circ f_{enc,x}.
$$
In aggregate, {\em the quantified rate is adaptive to such latent variable structures}, and is independent of the possibly high-dimensional variables $(\bar\bx_\perp,\bar\bz_\perp)$ which represents redundant information. Such results cannot be established for estimators based on fixed-form kernels; for example, under the setup of \cref{thm:cmm}, $\bar f_0$ will satisfy 
$\bar f_0\in \mrm{Ran}(\bar E^\top \bar E)^{\frac{\alpha}{2}} = W_{per}^{\alpha pd_l,2}\subset W^{\alpha p d_l, 2}$ (see Ex.~\ref{ex:main-torus}, and derivation in App.~\ref{app:disc-gtRKHS}). 
Across all possible $f_{enc,x}\in\cC^{\beta_d}$ as assumed in the theorem, 
we can only have $f_{enc,x}\in W^{\beta_d+\frac{d_{obs}}{2},2}$, due to the Sobolev embedding theorem, and thus 
$f_0 = \bar f_0\circ f_{enc,x}\in W^{r, 2}$, with\footnote{
Recall our convention (\cref{rmk:addi-conventions}) of using $E$ to also refer to a map between functions, as opposed to mere $L_2$ equivalence classes. 
}
$$
r = \min\Big\{\alpha pd_l, \beta_d + \frac{d_{obs}}{2}\Big\} = \min\Big\{p d_l, \beta_d + \frac{d_{obs}}{2}\Big\}, 
$$
where $\beta_d\ge (p-2)d_{obs}$ as assumed in the theorem. Consequently, for a Mat\'ern-$\alpha'$ RKHS, which is equivalent to $W^{\frac{2\alpha'+d_{obs}}{2},2}$ \citep[see \eg,][]{kanagawa_2018_gaussian}, 
to contain $f_0$, we must have $$
\alpha' + \frac{d_{obs}}{2}\le r \le p d_l.
$$
This can become {\em impossible} when $d_{obs} \ge 2p d_l-1$, a realistic scenario when $d_{obs}\gg d_l$. In contrast, our established result \eqref{eq:cmm-concrete-claim} is unaffected by this condition. 
More generally, there will always exist regimes of $(p,d_l,d_{obs})$, in which nonparametric models that can only utilize differentiability conditions of $f_0$ will suffer from the curse of dimensionality, whereas the learned kernel enjoys dimension-independent guarantees.\footnote{
Formal separability results can also be established by comparing the established rate for $\|E(\hat f_n-f_0)\|_2$ with a nonparametric regression rate.
} 
The improvement comes from the established adaptivity of DNN models \citep{schmidt-hieber_nonparametric_2020}. 

\paragraph{Choices of $\cI$, technical restrictions} The theorem uses the kernel \eqref{eq:kz-used} with $\alpha'\gets \alpha$. As noted in the proof, this approximates a kernel with integral operator $$
(EE^\top)^{\alpha'} = E(E^\top E)^{\alpha'-1} E^\top.
$$ 
In general, kernels with integral operator having the form $ET_x E^\top$ are ``optimal'' first-stage models for $\cH$ defined with $T_x$ \citep{wang2022fast}; this is a generalization of the concept of optimal linear instrument \citep{chamberlain_asymptotic_1987}. 
Thus, as noted in the text, a choice of $\alpha'\gets \alpha+1$ would have approximated the optimal kernel for $\cH$. 
Our restrictions of $\alpha'=\alpha$ and $\alpha=1$ are both due to \cref{prop:kern-est}, which bounds the extra error incurred by using plug-in estimates for the kernels (e.g., \eqref{eq:kernel-used} in place of \eqref{eq:popu-kx}), but 
only applies to $\alpha'=\alpha=1$. It is unclear if this is merely a technical artifact,\footnote{
We note that the proof of the theorem %
can be modified to establish a similar rate using the {\em population counterpart} of \eqref{eq:kz-used}, for $\alpha'=\alpha+1$. We may also replace $k_z$ with the construction of \citet{wang2022fast}, but this introduces additional assumptions.
} 
but in any case, 
our results are best viewed as a proof-of-concept that demonstrates qualitatively desirable properties (\eg, adaptivity). 
The experiments will evaluate other choices of $(\alpha,\alpha')$, and test other deviations from the conditions in the theorem.

%% file: app-exp.tex
\section{Experiment Setup and Additional Results}

\subsection{Synthetic Experiment}\label{app:exp-synth}

We first evaluate our method in a more controlled setting which is consistent with our running example. Concretely, we consider a nonparametric IV (NPIV) problem, where the observations $(\bz,\bx,\by)$ are generated using the process 
\begin{align*}
\bar\bz&\sim\cN(0, 1), ~ \bar\bu\sim\cN(0, 1), ~\bar\bx := \rho\bar\bz + \sqrt{1-\rho^2} \bar\bu, ~(\bz_\perp;\bx_\perp)\sim\cN(0, I_{2(d-1)\times 2(d-1)}), \\ 
\bx &:= f_{dec,x}(\bar\bx,\bx_\perp), ~
\bz := f_{dec,z}(\bar\bz,\bz_\perp), ~
\bar\by \sim \cN(f_0(\bar\bx) + 2\bu, 0.01), ~
\by := s_y^{-1}(\bar\by - m_y).
\end{align*}
In the above, $f_{dec,x}, f_{dec,z}$ are chosen as randomly initialized MLPs with two hidden layers and output dimensionality $2d$, $f_0$ is chosen as the absolute value function, and the constants $s_y, m_y$ are chosen to standardize $\by$. This setup generalizes the low-dimensional simulations used in a thread of NPIV works starting from \citet{lewis2018adversarial}. 

As the latent variables $(\bar\bx,\bar\bz)$ follow a multivariate normal distribution, 
it is well known that the singular vectors of the ``latent space conditional expectation operator'', $\bar E: L_2(P_{\bar x})\to L_2(P_{\bar z})$, are the scaled Hermite polynomials, 
and its $j$-th singular value equals $\rho^j$; as we discussed in \cref{app:disc-gtRKHS}, $E$ and $\gtRKHS[\alpha]$ are also defined accordingly, provided that there exist the functions $f_{enc,x}: x\mapsto\bar x, f_{enc,z}: z\mapsto \bar z$.\footnote{
While we do not explicitly ensure their existence in the design, empirically we find it possible to reconstruct $\bar\bx,\bar\bz$ using the observed $\bx,\bz$. 
}
And the $\chi^2$ divergence equals $\frac{\rho^2}{1-\rho^2}$. 

We are interested in the convergence of the spectral decomposition error $\epsilon_n^2 = \cD(h_0) - \cD(\hat h_n)$, and the NPIV estimation error $\|\hat f_n -f_0\|_2^2$ using the learned kernel. For the former, we compare $\cD(\hat h_n)$ with the $\chi^2$ divergence which it should converge to (\cref{lem:loss-equivalence}), and an estimate of the divergence
\begin{equation}\label{eq:chisq-est}
\hat\cD_{\chi^2, n'} := \frac{1}{n'} \sum_{i=1}^{n'} (h_0(z_i', x_i') - 1)^2.
\end{equation}
We compare with \eqref{eq:chisq-est} since due to the light tail of the Gaussian distribution, finite-sample estimates of the $\chi^2$ divergence may converge slowly, especially when $\rho$ is large. 
For the NPIV error, we compare it with the kernelized IV estimator \citep{dikkala_minimax_2020} instantiated with a fixed-form RBF kernel. This is only provided for the reader's convenience, and more baselines are included in the more realistic semi-synthetic experiment in the following. 

\begin{figure}[htbp]
\centering
\subfigure[$\rho=0.5,d=1$]{\includegraphics[width=0.31\linewidth]{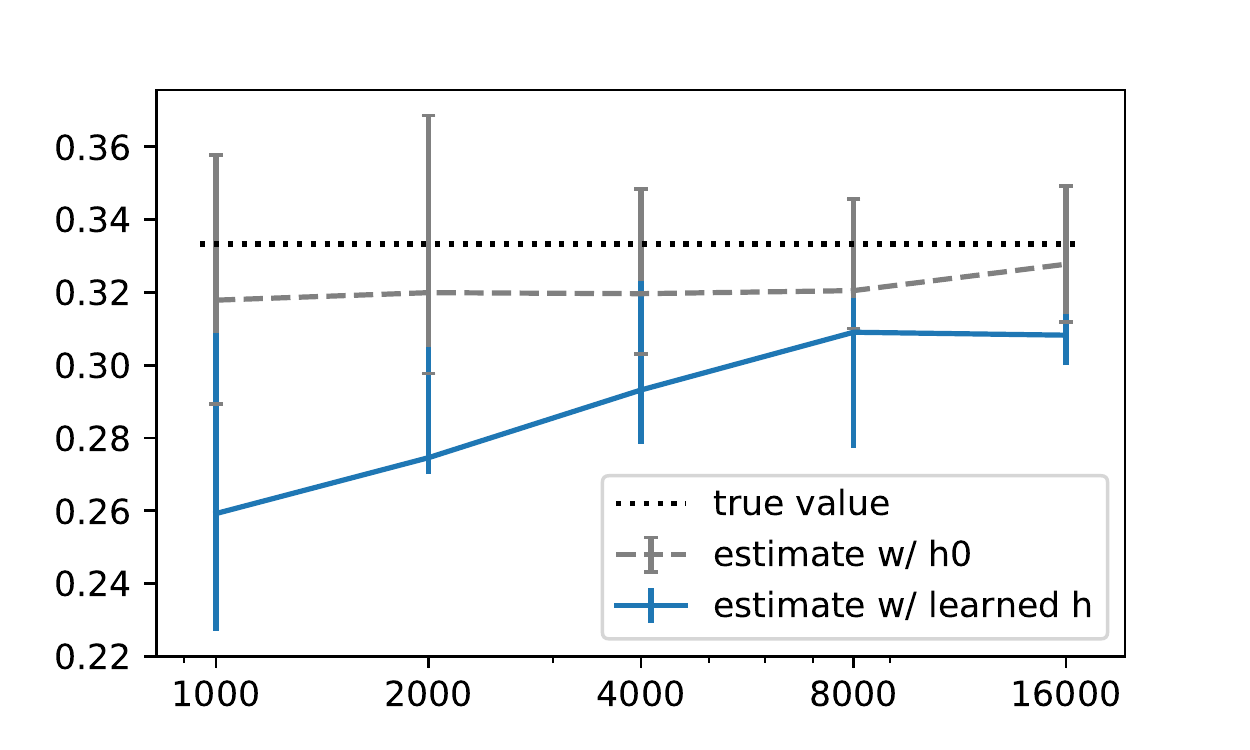}}
\subfigure[$\rho=0.7,d=1$]{\includegraphics[width=0.31\linewidth]{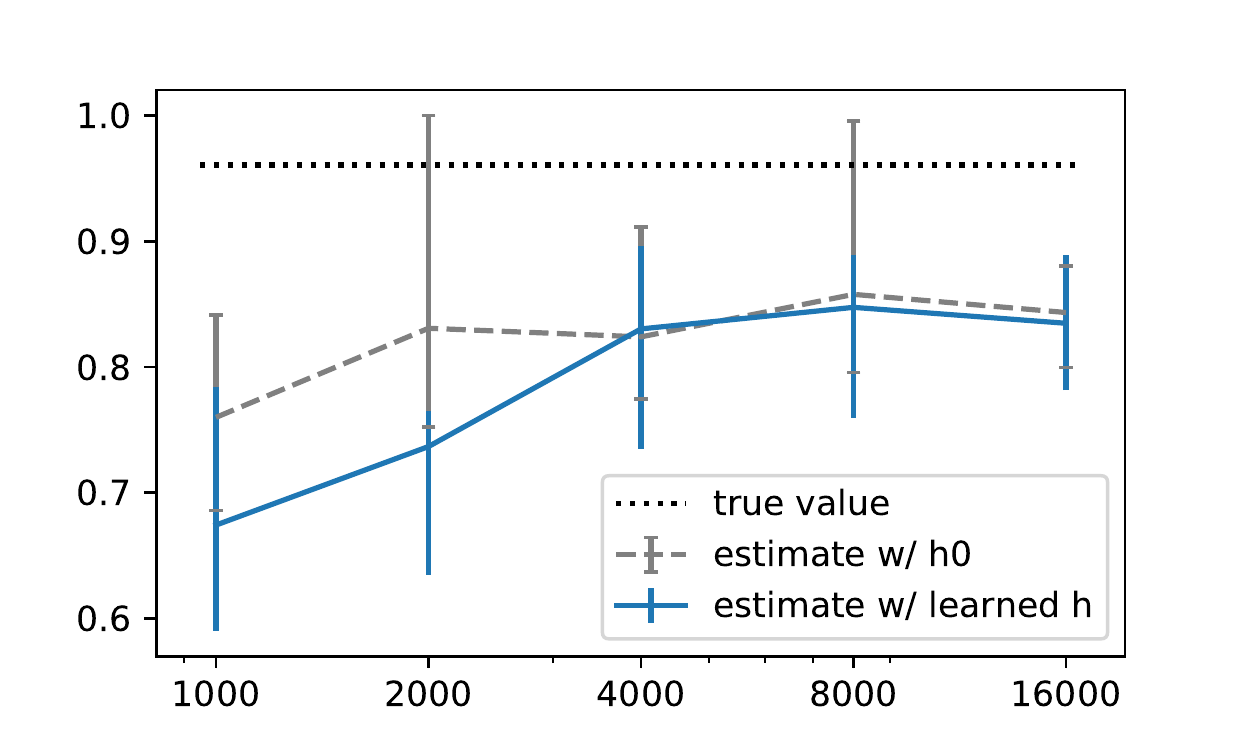}}
\subfigure[$\rho=0.9,d=1$]{\includegraphics[width=0.31\linewidth]{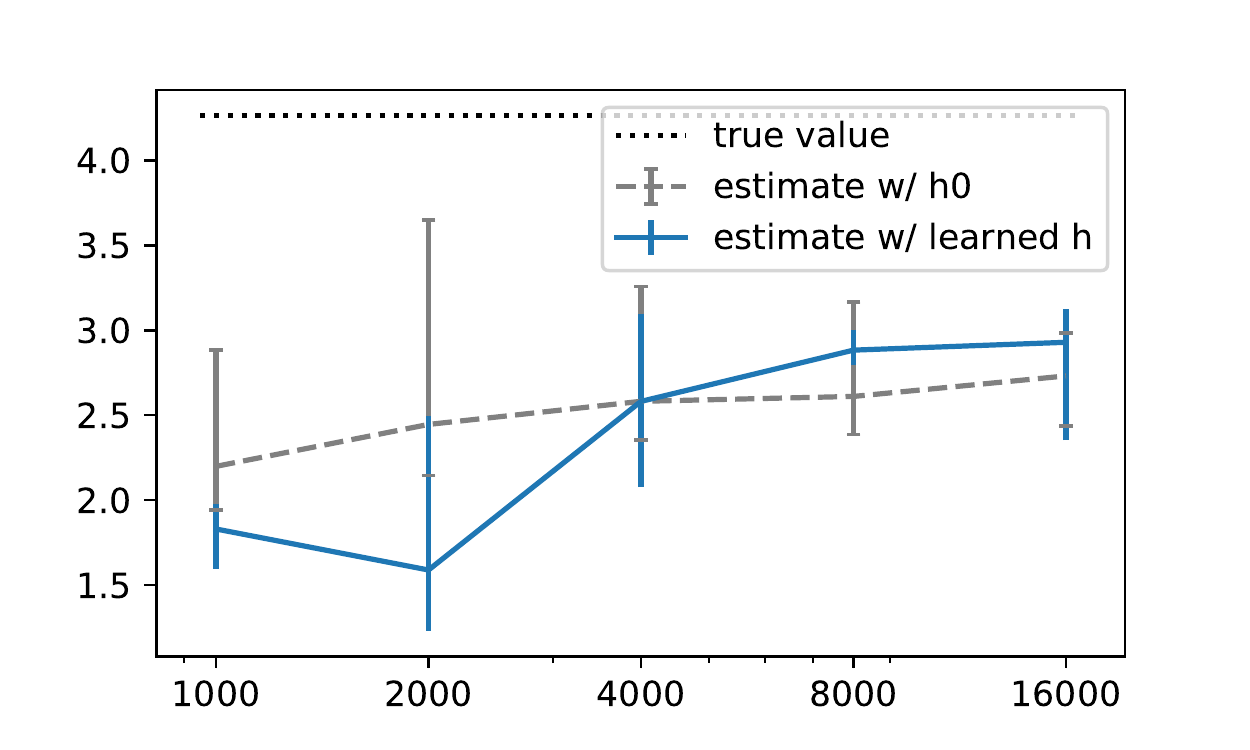}}
\subfigure[$\rho=0.7,d=1$]{\includegraphics[width=0.31\linewidth]{figs/synth_0.7_1.h.pdf}}
\subfigure[$\rho=0.7,d=2$]{\includegraphics[width=0.31\linewidth]{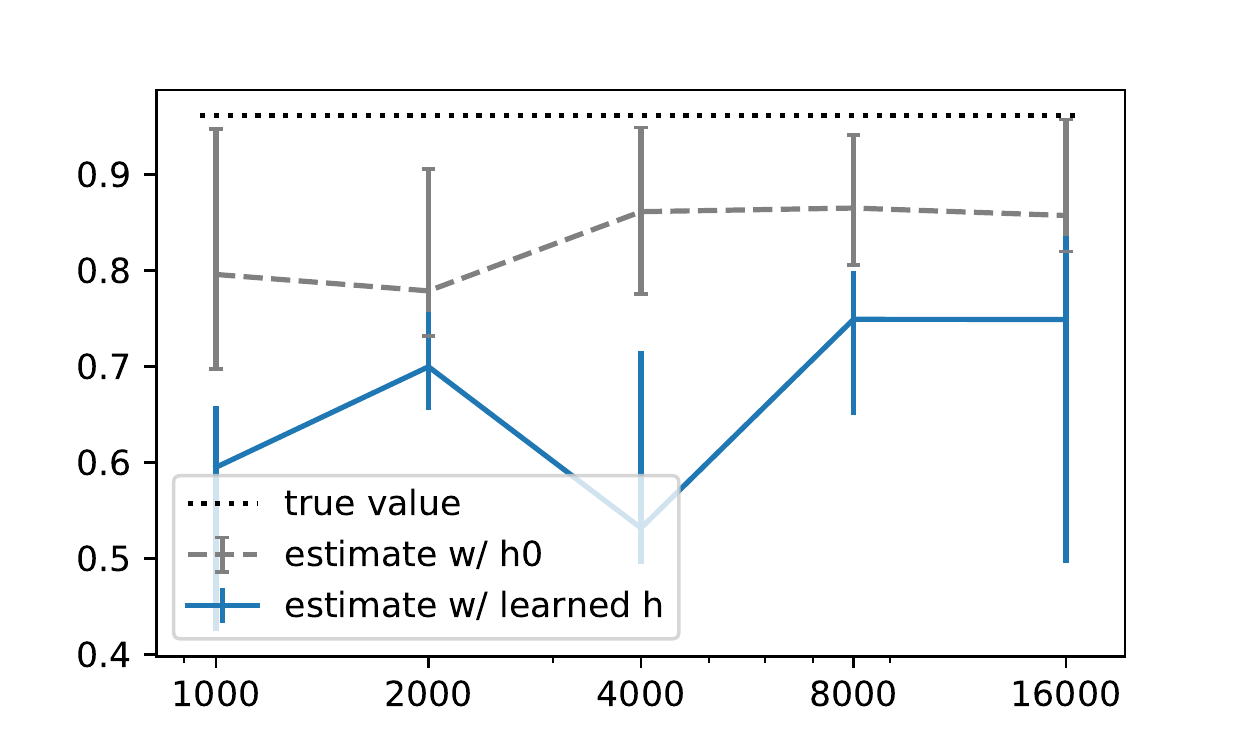}}
\subfigure[$\rho=0.7,d=10$]{\includegraphics[width=0.31\linewidth]{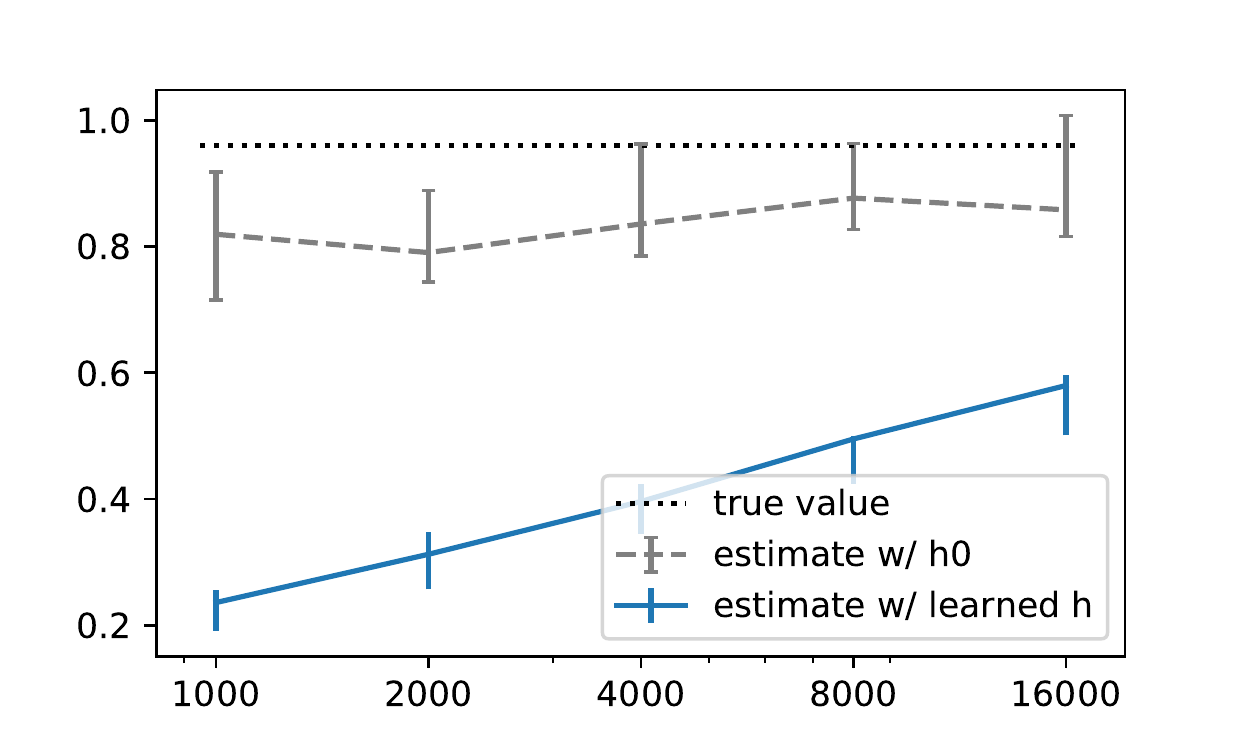}}
\caption{Synthetic experiment: estimated lower bound of $\cD_{\chi^2}[P_{zx}\Vert P_z\otimes P_x]$ using the learned $\hat h_n$. Error bar denotes the $20\%$ and $80\%$ percentile across $10$ independent runs.}\label{fig:synth-div}
\end{figure}

We vary $d\in\{1, 2, 10\}, \rho\in\{0.5, 0.7, 0.9\}$ and $n\in\{1,2,4,8,16\}\times 10^3$. 
For the estimation of $\hat h_n$, we parameterize $h(z,x) = \Phi(z)^\top\Psi(x)$ using $4$-layer MLPs for both $\Phi$ and $\Psi$, with Swish activation and width of $50$, and apply dropout regularization with a rate of $0.2$. The NN model is trained with AdamW, with a learning rate of $10^{-3}$ and batch size of $512$. We have experimented with a variety of NN architecture and training hyperparameters and selected this configuration by evaluating the training objective on held-out samples. We conduct early stopping, and determine the value of $J_n\in\{10, 20, 30\}$ by the same validation procedure. 
For IV estimation, we use plug-in estimates for the kernels \eqref{eq:kernel-used} and \eqref{eq:kz-used}, and select $\alpha\in\{0.5, 1, 2, 3\}, \lambda_n\in [0.5, 2], \nu_n\in [0.5, 2]$ using the validation protocol of \citet{singh_kernel_2020,muandet_dual_2020}. For the RBF baseline, we also select the bandwidth of $k_x$ in $\{0.5, 1, 1.5\} \times \mrm{med}_x$, and $k_z$ in $\{1, 2, 3\} \times \mrm{med}_z$ using the same protocol. Note that RBF kernels with varying bandwidths form a similar Hilbert scale \citep[e.g.,][]{steinwart2008support}. Both methods observe a total of $2n$ samples, and are tested on a separate set of $n'=5n$ samples. 

\begin{figure}[htbp]
\centering
\subfigure[$\rho=0.5,d=1$]{\includegraphics[width=0.31\linewidth]{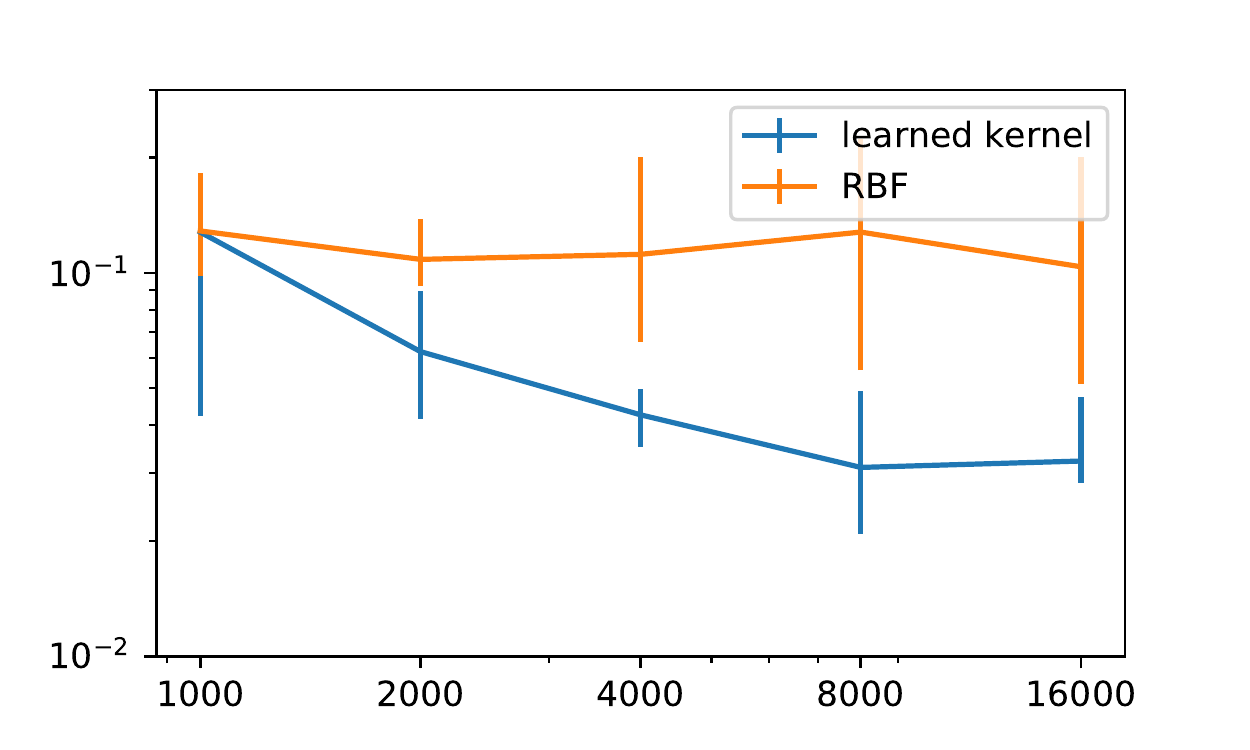}}
\subfigure[$\rho=0.7,d=1$]{\includegraphics[width=0.31\linewidth]{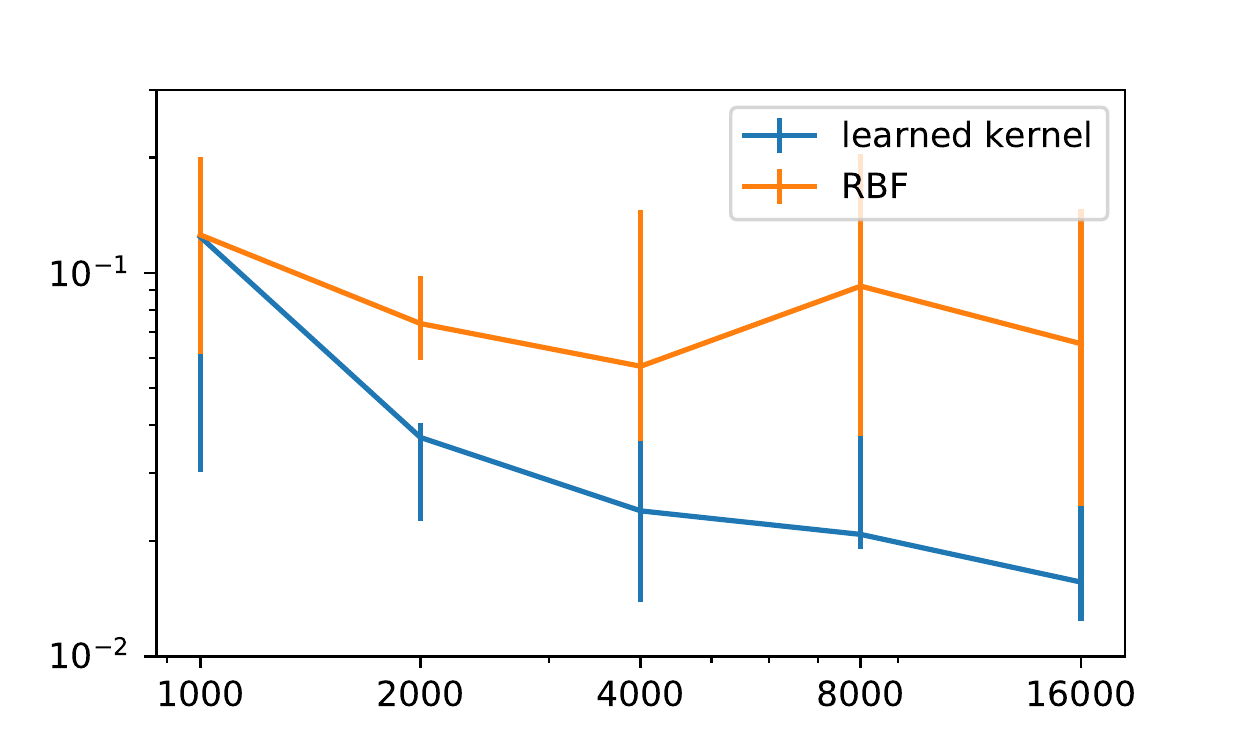}}
\subfigure[$\rho=0.9,d=1$]{\includegraphics[width=0.31\linewidth]{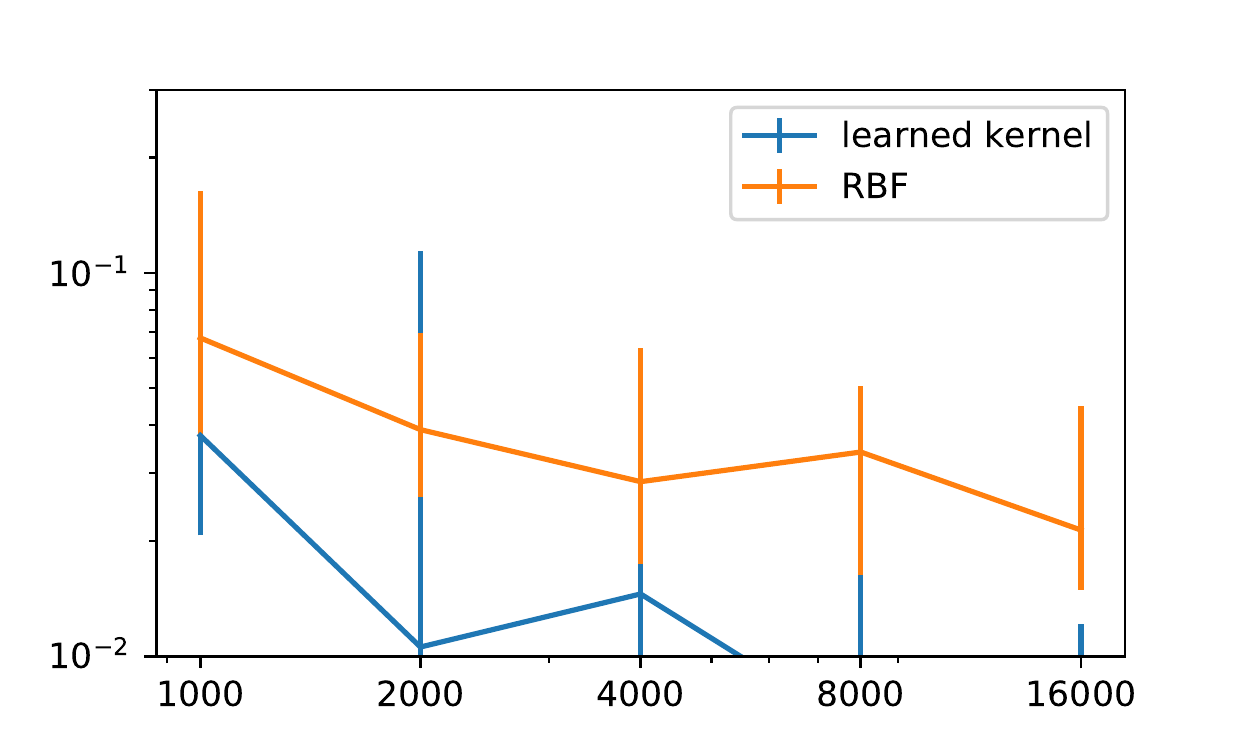}}
\subfigure[$\rho=0.7,d=1$]{\includegraphics[width=0.31\linewidth]{figs/synth_0.7_1.iv.pdf}}
\subfigure[$\rho=0.7,d=2$]{\includegraphics[width=0.31\linewidth]{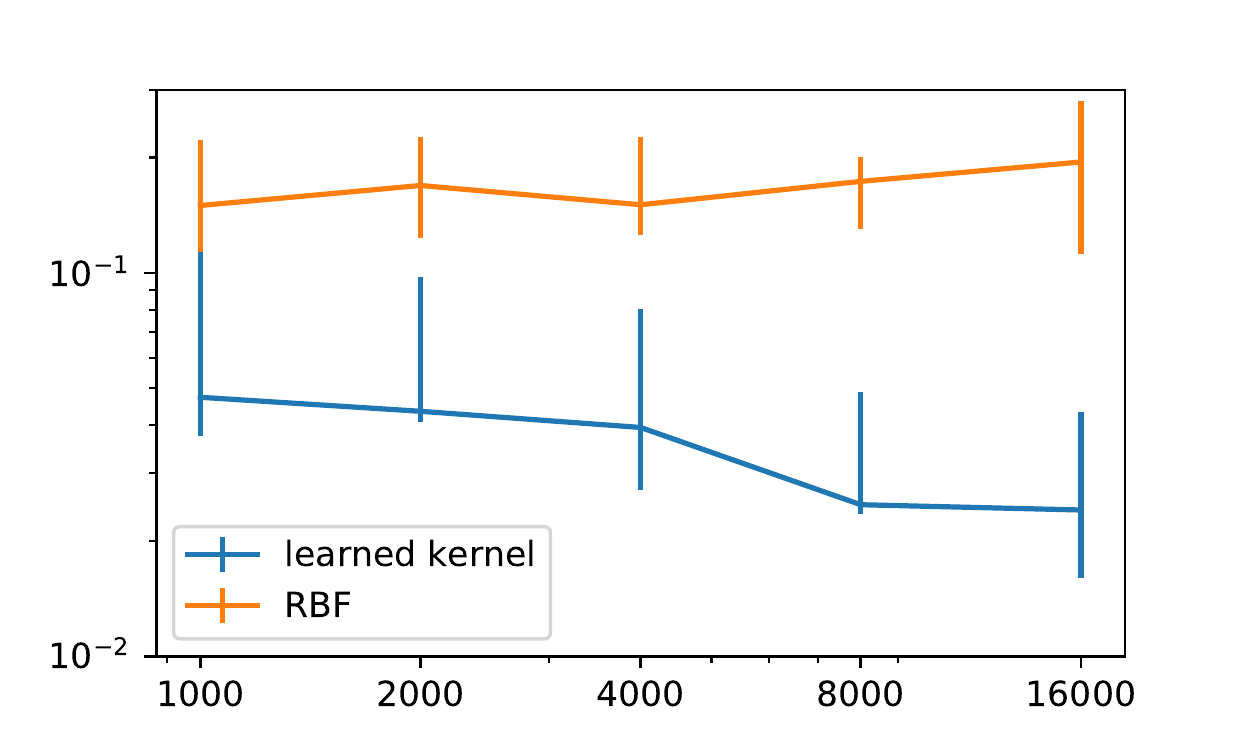}}
\subfigure[$\rho=0.7,d=10$]{\includegraphics[width=0.31\linewidth]{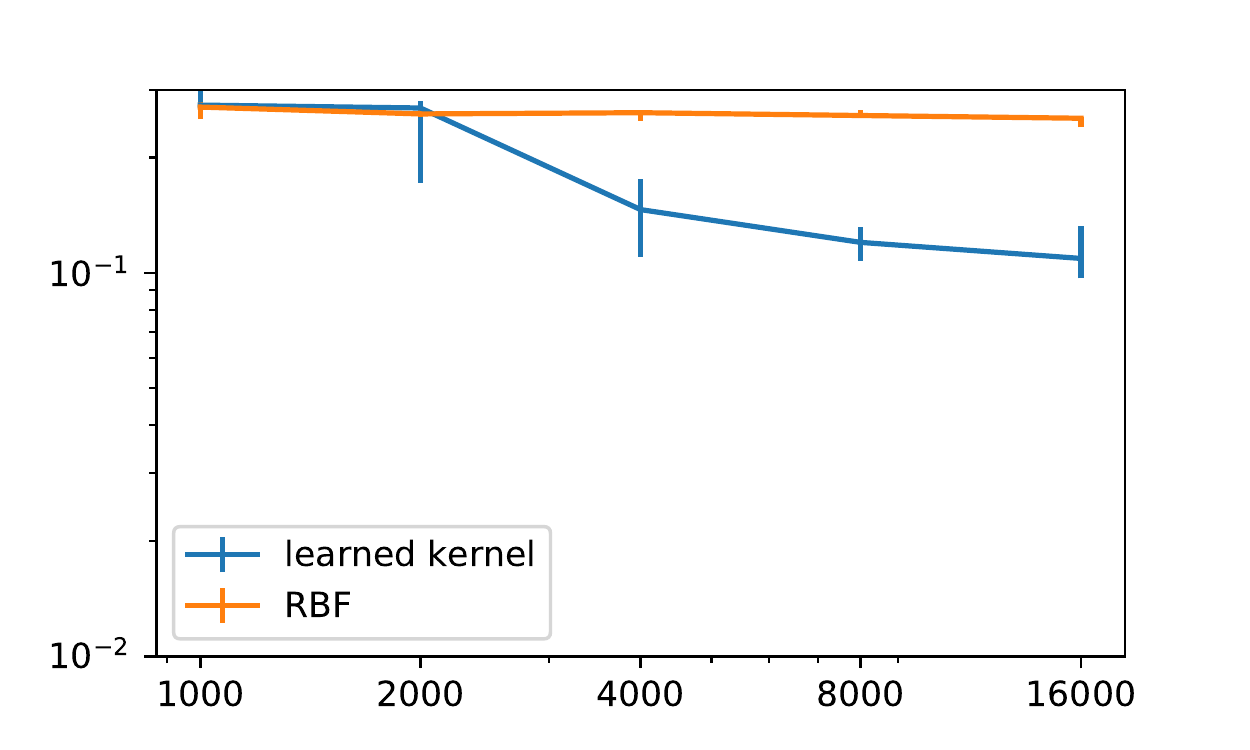}}
\caption{Synthetic experiment: IV estimation error $\|\hat f_n - f_0\|_2^2$ using learned and fixed-form kernels. Error bar denotes the $20\%$ and $80\%$ percentile across $10$ independent runs.}\label{fig:synth-iv}
\end{figure}

The results are shown in Figure~\ref{fig:synth-div}-\ref{fig:synth-iv}. As we can see, the estimated density ratio $\hat h_n$ approaches the ground truth as sample size increases, and leads to accurate estimates for the $\chi^2$ divergence when $d$ and $\rho$ have moderate values. Obviously, estimation will become more difficult as $d$ increases; this is also the case for increasing $\rho$, in which case the trailing eigenfunctions become more relevant in the $\chi^2$ divergence. As for IV estimation, the learned kernel leads to better estimates when $d$ is large, as we would expect given the discussion in \cref{app:disc-cmm-results}. It is interesting that it also outperforms the fixed-form kernel when $d=1$, i.e., when no redundant latent features are present, which suggests the method may also be useful in more classical, low-dimensional settings. 

\subsection{Proxy Control Experiment}\label{app:exp-pc}

We now turn to the proxy control experiment. Our setup is adapted from \citet{mastouri_proximal_2021}, based on which we modify the proxy variables as follows: denote by $\bar\bv, \bar\bw$ the low-dimensional proxy variables in \citet{mastouri_proximal_2021}; our observed proxies are generated as 
$$
\bv := f_{dec,v}(\bar\bv, \bv_\perp), ~
\bw := f_{dec,w}(\bar\bw, \bw_\perp), ~~~\text{where}~
\bv_\perp \sim\cN(0, I)\in\RR^{D_{ex}}, ~
\bw_\perp \sim\cN(0, I)\in\RR^{D_{ex}}
$$
are independent of all other variables, and $f_{dec,v}, f_{dec,w}$ are randomly initialized neural networks with two hidden layers, and output dimensionality $4$ times the input dimensionality. The other data variables are kept unchanged. We vary $D_{ex}\in\{0, 4, 32, 256\}, N\in \{1000, 2000, 4000, 8000\}$. 

For our method, we employ a tensor product kernel described in \cref{sec:algo-impl}, where $k_w$ is an RBF kernel with bandwidth determined by the median trick; the other hyperparameters are determined as in \cref{app:exp-synth}. 
For the \texttt{RBF} baseline, we use RBF kernels, with bandwidth determined from the same grid as in \cref{app:exp-synth}, using the same protocol. For the \texttt{AGMM-NN} baseline, we vary the learning rates in $\{5,10,50\}\times 10^{-4}$ and $L^2$ regularization in $\{5,10\}\times 10^{-5}$, and report the configuration that attains the best median MSE. 
For all configurations, we repeat the experiment for $10$ times and report the $25\%, 50\%$, and $75\%$ percentile of the test MAE. All methods observe a total of $2N$ samples. 

The results are shown in \cref{tab:proxy_experiment}, \cref{fig:proxy-varying-N} in the text, and \cref{fig:proxy-varying-N-lowdim} below. As we can see, the proposed method outperforms other baselines in most settings and becomes especially competitive as the dimensionality of the observed proxy variables increases. 

\begin{figure}[htbp]
 \centering 
 \includegraphics[width=0.85\linewidth]{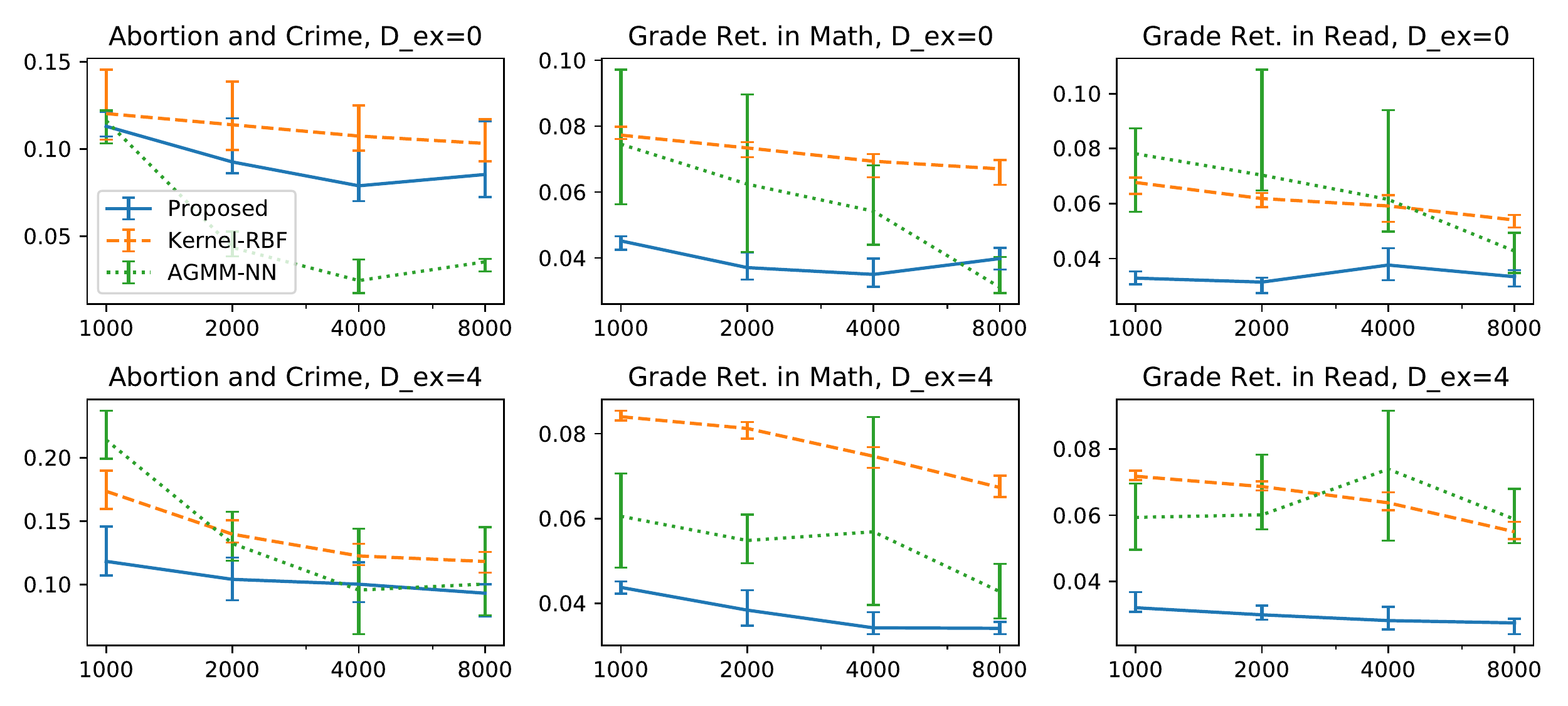}   
 \caption{Proxy experiment: test MAE for varying $n$, and $D_{ex}\in\{0, 4\}$.}\label{fig:proxy-varying-N-lowdim}
\end{figure}